\pdfoutput=1

\documentclass[11pt]{article}

\usepackage[]{ACL2023}

\usepackage{times}
\usepackage{latexsym}

\usepackage[T1]{fontenc}

\usepackage[utf8]{inputenc}

\usepackage{microtype}

\usepackage{inconsolata}

\usepackage{amsmath}
\usepackage{booktabs}
\usepackage{longtable}
\usepackage{enumerate,enumitem}

\usepackage{tcolorbox}
\tcbuselibrary{breakable, skins}

\newtcolorbox{promptbox}[2][]{
  breakable,
  colback=blue!5,
  colframe=blue!40,
  fonttitle=\bfseries\small,
  title=#2,
  left=6pt, right=6pt, top=4pt, bottom=4pt,
  #1
}

\usepackage{subcaption}
\usepackage{float}

\usepackage{twemojis}

\usepackage{makecell}
\usepackage{amssymb}

\newcounter{notecounter}
\newcommand{\enotesoff}{\long\gdef\enote##1##2{}}

\enotesoff

\title{Do We Still Need Humans in the Loop? Human vs. LLM Annotation in Active Learning for TikTok Hate Speech Detection}

\author{%
  Ahmad Dawar Hakimi$^{1,2,3}$ \hspace{0.05cm}
  Lea Hirlimann$^{1,3}$ \hspace{0.05cm}
  Isabelle Augenstein$^{2}$ \hspace{0.05cm}
  Hinrich Schütze$^{1,3}$ \\[0.5em]
  $^1$Center for Information and Language Processing, LMU Munich, Germany \\
  $^2$Department of Computer Science, University of Copenhagen, Denmark \\
  $^3$Munich Center for Machine Learning, Germany \\
  \texttt{adhakimi@cis.lmu.de}
}

\begin{document}
\maketitle

\begin{abstract}
Annotating data remains a costly bottleneck for supervised NLP.
Active learning (AL) reduces the number of human labels needed by
selecting only the most informative instances, while instruction-tuned
LLMs attack the same bottleneck from the other side, making labels
cheap enough to annotate entire corpora. This raises two questions:
can LLM labels replace human labels within the AL loop, and does AL
remain necessary when entire corpora can be cheaply labeled? We
investigate both by training supervised hate speech classifiers on a
new dataset of 278K German political TikTok comments, comparing human
and LLM annotation under matched conditions. LLM annotation at scale
outperforms human-supervised classifiers at roughly one-tenth the cost,
for both a closed-source (GPT-5.2) and an open-weight (Qwen3.5-122B-A10B)
LLM, and the advantage is robust under soft-label evaluation. It hinges
on the annotation interface: only a two-question decomposition mirroring
the human annotation task unlocks it. AL provides no reliable advantage
over random sampling in our prefiltered pool. Error structure
depends on the LLM: only GPT-5.2 matches the human FP/FN balance,
while other variants over-flag border-control and economic-competition
discourse. Humans remain essential as evaluators; for training labels,
the question shifts to which LLM, which interface, and what shape of
pool. We release the dataset and
code.\footnote{\url{https://github.com/adhakimi/do-we-still-need-humans-in-the-loop}}

\textcolor{red}{\textbf{Content warning.}} This paper contains examples
of hostile and anti-immigrant language from social media, included for
illustration.
\end{abstract}

\enote{hs}{it would be good if you could use consistent
names of the different conditions. e.g., you use both
\textsc{Full-Human}  and Full-Human (without sc)}

\section{Introduction}
\label{sec:intro}

\begin{figure}[t]
    \centering
    \includegraphics[width=\columnwidth]{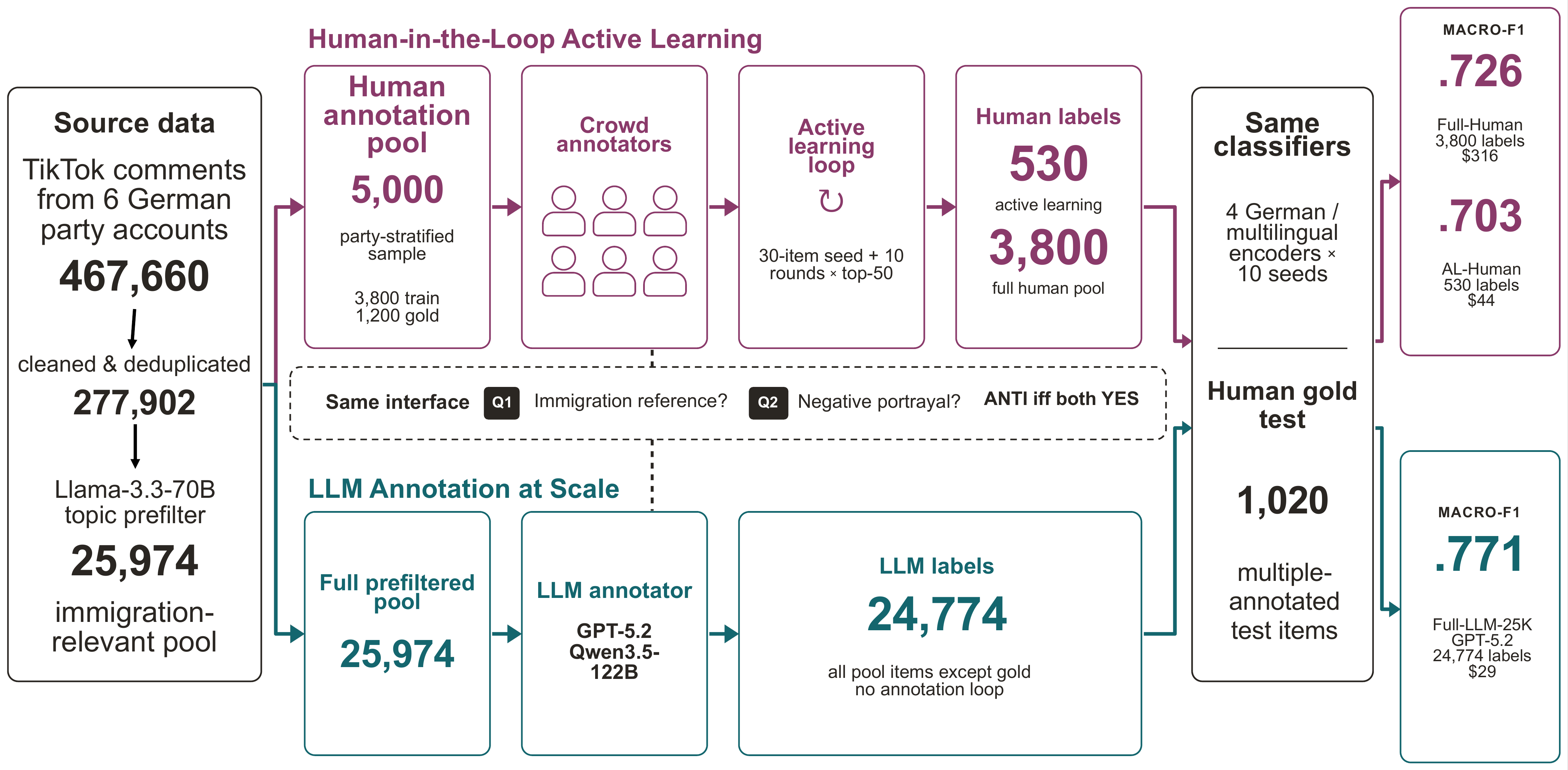}
    \caption{\textbf{Two annotation paradigms compared} on German political TikTok comments. Human-in-the-loop AL over a crowd annotated subset (top) vs.\ LLM annotation of the full prefiltered pool under the decomposed interface (bottom). Both train the same encoder classifiers and are evaluated on the same human gold test set.}
    \label{fig:overview}
\vspace{-5mm} 
\end{figure}

Annotating data is expensive, a persistent bottleneck for supervised NLP. A recent survey finds that 94\% of practitioners consider annotation a limiting factor, and that active learning (AL) remains widely used~\citep{romberg2025have}. In AL, a classifier iteratively selects the most informative unlabeled instances, queries a human for labels, and retrains, aiming to reach target performance with fewer labels than random sampling~\citep{settles2009active}.

Recent work places LLMs inside the AL loop as
annotators \citep{zhang2023llmaaa, kholodna2024llms}, but
these studies do not compare human and LLM labels under
matched conditions, nor do they examine how the annotation
source affects classifier behavior. When an
LLM can cheaply label an entire corpus,
the practical question shifts: from how to allocate a human annotation budget (AL versus random sampling) to whether to spend one at all, with AL (human labels) now competing against full LLM annotation at scale.
\enote{hs}{this sound like ``the question is no longer
chocolate-or-vanilla, but 
chocolate-or-vanilla-on-hot-days.  
The question chocolate-or-vanilla-on-hot-days is still a
type
of chocolate-or-vanilla question, so the question still is 
chocolate-or-vanilla. ideally, you would put a qualifier for
what is no longer a question. do you mean individual
annotatinos? annotation of small datasets?}
Apart from cost, LLM annotation
does not expose workers to
psychological harm \citep{steiger2021wellbeing},
may help overcome the scarcity of qualified workers  in many languages, and
enables rapid iteration over annotation guidelines. We
use cost as the operational lens because it is the most
directly measurable, and test whether AL retains a
performance advantage that justifies its cost premium.

We investigate this question on detecting anti-immigrant hostility, a form of hate speech targeting immigrant groups, in German TikTok comments collected from the official accounts of six major political parties. The task is challenging to annotate: it requires differentiating hostility directed at immigrant groups from criticism of immigration as a policy issue, a culturally specific distinction where annotation source matters~\citep{sap2022annotators}. It also represents a common deployment scenario: training a lightweight supervised classifier from limited labels for inference at scale.
Figure~\ref{fig:overview} contrasts the two paradigms we compare; the full design spans seven labeling conditions (Table~\ref{tab:conditions}) that vary annotation source, sampling strategy (AL vs.\ random vs.\ full pool), and label volume (530 to 25,974 instances) across four encoders and 10 random seeds. The LLM conditions use GPT-5.2 and Qwen3.5-122B-A10B (henceforth Qwen3.5) with a two-question decomposition matching the human annotation interface; an interface ablation (GPT-5.2 with a holistic single-question prompt) and a reasoning-mode ablation (Qwen3.5 without reasoning) are reported in App.~\ref{app:interface-ablation} and App.~\ref{app:reasoning-ablation}. The labeling pool comprises 25,974 immigration-relevant comments constructed via a two-stage LLM pipeline, of which 5,000 are independently annotated by six crowdworkers, yielding 1,200 gold evaluation items and 3,800 training labels. Throughout, we describe classifiers as \emph{human-supervised} or \emph{LLM-supervised} according to the source of their training labels; in the AL conditions, the same source provides the labels for queried instances.
We organize the paper around four research questions:
\begin{enumerate}[
    leftmargin=*,
    align=left,
    labelsep=0.3em,
    label=\textbf{RQ\arabic*:},
    topsep=2pt,
    itemsep=1pt,
    parsep=0pt,
    partopsep=0pt
]
    \item Can LLM annotation at scale match or exceed human-supervised
    classifiers, and how does this depend on the annotation interface and
    the choice of LLM?
    \item Does any AL acquisition strategy reliably outperform random sampling in an LLM-prefiltered pool, under either human or LLM labels?
    \item Is the annotation at scale advantage robust to soft-label evaluation
    that accounts for human annotator disagreement?
    \item How do error structure and topical disagreement vary across LLM
    annotators?
\end{enumerate}
\enote{hs}{above: i don't think putting rq1-rq4 to the left
    of the column margin is compatible with the acl template}

Our central finding is that under a two-question interface that mirrors the human annotation task, LLM annotation at scale considerably outperforms human-supervised classifiers at roughly one-tenth the cost using a closed-source LLM (GPT-5.2), while an open-weight alternative (Qwen3.5) run on local hardware matches human supervision.

\section{Related Work}
 
\subsection{Active Learning and LLM Annotators}

AL reduces annotation cost by selecting the most informative instances for human labeling \citep{settles2009active}, but practical adoption remains limited by unreliable performance gains, hyperparameter sensitivity, and difficulty obtaining reliable uncertainty estimates for acquisition functions \citep{lowell2019practical, romberg2025have, schroeder2020survey}. AL strategies frequently fail to outperform random selection on text and multimodal tasks, with pool composition often the underlying cause \citep{karamcheti2021mind, snijders2023investigating}. Instruction-tuned LLMs offer a complementary cost reduction: zero-shot ChatGPT achieves higher accuracy than crowd workers relative to trained-annotator labels across several text annotation tasks at roughly one-thirtieth the cost \citep{gilardi2023chatgpt}. These findings extend to open-source models and other domains \citep{alizadeh2025opensource, tornberg2024llms}, including German migration discourse \citep{kostikova2024fine}. Recent work places LLMs inside the AL loop as annotators \citep{zhang2023llmaaa, kholodna2024llms, he2024annollm, xia2025selection}, though annotation quality is task- and prompt-sensitive and requires validation against human gold labels \citep{pangakis2023automated}, with hate-speech zero-shot performance depending on the definition prompt \citep{melis2025modular}.
We address these gaps with a matched-condition comparison of human and LLM annotations that evaluates downstream classifier behavior (not just annotation agreement) and contrasts AL with human labels against LLM annotations at scale.

\enote{hs}{this may not be possible because of space issues,
but what's missing here for me is a clear statement at the
end that repeats why your work is novel and clearly
differeniated from prior work. (the information is not
missing from what's been said so far, but still it's good to
repeat it here)}

\subsection{Subjective Annotation and Disagreement}
 
Hate speech detection has been widely surveyed \citep{schmidt2017survey, fortuna2018survey}; annotation difficulty increases when hostility toward a \emph{group} must be distinguished from criticism of a \emph{policy} \citep{waseem2016hateful} or when bias is implied rather than explicit \citep{sap2020social}, the central challenge in our task. Related work creates dehumanization annotation resources \citep{engelmann2024annotating} and treats ambiguity as signal \citep{assenmacher2025bilingual}. LLM hate-speech annotation is sensitive to extralinguistic context: party cues for political text \citep{vallejo2025llms} and geographical, persona, and anchoring cues \citep{masud2024hate}. The most related German dataset \citep{fillies2025german} covers far-right ecosystems rather than party accounts. Hate-speech annotation faces further challenges in distinguishing hate from offensive language \citep{davidson2017automated}, cross-dataset and cross-cultural transfer \citep{antypas2023robust, lee2024exploring}, and dataset construction \citep{piot2024metahate, vidgen2021learning}. Moreover, dataset construction methods may inadvertently bias hate speech classifiers toward identity-related cues \cite{sen-etal-2022-counterfactually}.
Annotator disagreement on subjective tasks is increasingly understood as signal rather than noise \citep{bjerva-etal-2020-subjqa,plank2022problem, rottger2022two, davani2022dealing}, with low inter-annotator agreement common even under detailed guidelines for German hate speech \citep{ross2017measuring}. Recent work proposes aggregating soft labels to capture this signal in both training and evaluation \citep{wright2025aggregating}, raising the question of whether LLMs can capture human disagreement patterns \citep{ni2025llm}.

\section{Methodology}

\subsection{Dataset Collection and Preprocessing}

We collected public comments from the official TikTok
 accounts of six major German political parties (CDU/CSU,
 Bündnis~90/Die~Grünen, SPD, FDP, AfD,  Die~Linke) via the
 TikTok Research API, covering January 2024 to September
 2025, yielding 467{,}660 comments. Anti-immigrant hostility
 is a culturally embedded subjective task where annotation
 source plausibly matters, and our corpus is newly collected
 rather than drawn from an existing benchmark, reducing
 pretraining exposure for LLM annotators. We applied a
 four-stage preprocessing pipeline: (1)~removal of duplicate
 comment IDs; (2)~content filtering to remove
 non-informative comments (emoji-only, empty, or
 single-token sequences);
 (3)~exact text deduplication within and across party
 accounts; and (4)~language identification via
 GlotLID~\citep{kargaran2023glotlid} to
remove non-German.
This yielded a
corpus of \textbf{277{,}902} German comments (median 14 words); per-step/per-account counts are in App.~\ref{app:data}. This corpus forms the input to the LLM-based prefiltering stage (\S\ref{sec:llm-pool}).

\subsection{Label Space Definition}
\label{sec:label_space}

Following prior work on anti-immigrant hate
speech~\citep{basile2019semeval, sanguinetti2018italian}, we
define a binary classification task.
A comment is
labeled \textsc{anti-immigrant} if it expresses hostility
towards immigrants (see next paragraph for detailed description).
All other comments are
labeled \textsc{not anti-immigrant}.

\enote{hs}{ since this is central, it's probably ok to
repeat it, but not if we're short of space. OLD VERSION
A comment is
labeled \textsc{anti-immigrant} if it expresses hostility,
fear, or opposition toward immigrants, refugees, or asylum
seekers as a group, including xenophobic language,
stereotypes, deportation rhetoric, or attribution of social
problems to these groups. All other comments are
labeled \textsc{not anti-immigrant}.
}

\paragraph{Two-Question Annotation Framework.}
We operationalize this label through the two sequential yes/no questions Q1/Q2 rather than a direct binary judgment (same decomposition for both human and LLM annotation), paralleling HatEval's definition~\citep{basile2019semeval}, which likewise requires both reference to the target group and hateful content.
\begin{itemize}[noitemsep, topsep=0pt, partopsep=0pt]
    \item \textbf{Q1 (reference):} Does this comment refer to immigrants, refugees, asylum seekers, foreigners, or immigration policy?
    \item \textbf{Q2 (valence):} Are these groups portrayed negatively through hostility, stereotypes, deportation rhetoric, or attribution of social or economic problems?
\end{itemize}

The final label is computed as:
$\textsc{anti-immigrant} \Leftrightarrow \text{Q1\,=\,YES} \wedge \text{Q2\,=\,YES}$;
otherwise \textsc{not anti-immigrant}. The decomposition
serves two purposes. For human annotators, it addresses low
agreement in an initial pilot using a single holistic prompt
(Fleiss $\kappa$$=$$.175$; App.~\ref{app:holistic-pilot}),
which required integrating topic, target, and valence in a
single step. Embedding each criterion in the interface
forces explicit evaluation and enables diagnostic tracing of
disagreements to either reference (Q1) or  valence
(Q2). The same decomposition
is used for the LLM interface,
supporting a direct comparison without interface asymmetry confounds. We refer to this as the \emph{decomposed} (D) interface; a \emph{holistic} (H) ablation is reported in App.~\ref{app:interface-ablation}. Annotated example comments for each case are in App.~\ref{app:example-comments}.

\subsection{Pool Construction and Annotation}
\label{sec:llm-pool}
 
We construct the labeling pool through a two-stage LLM-based
pipeline. Stage~1 uses Llama-3.3-70B-Instruct as a high-recall topic filter, chosen for high-throughput batched inference at low cost. Stage~2 produces labels under the decomposed interface from \S\ref{sec:label_space} for the more difficult subjective classification, using two frontier LLMs, closed-source GPT-5.2 and open-weight Qwen3.5-122B-A10B, to test whether findings generalize across LLMs. Implementation details are in App.\ref{app:infra}.
 
\paragraph{Stage 1: Topic Prefiltering.}
We applied \texttt{Llama-3.3-70B-Instruct}~\citep{grattafiori2024llama} with a binary prompt asking whether each comment discusses immigration, migrants, refugees, asylum seekers, or related policies (prompt: App.~\ref{app:llama-filter}). This reduced the corpus to \textbf{25{,}974 comments}.
 
\paragraph{Stage 2: Annotation.}
\textbf{GPT-5.2 (D).} We applied \texttt{GPT-5.2} via the OpenAI Batch API with a single prompt eliciting Q1 and Q2 as structured JSON (prompt: App.~\ref{app:gpt-prompt-decomposed}); all 25{,}974 comments received valid labels (Table~\ref{tab:llm_labels}), with the 21.8\% \textsc{anti-immigrant} rate close to the 21.7\% base rate in the human gold evaluation subset (\S\ref{sec:human-annotation}).

\textbf{Qwen3.5-122B-A10B (D).} Qwen3.5-122B-A10B \citep{qwen3.5} ran on the same 25{,}974 comments with reasoning enabled. 136 items hit the 16{,}384-token output budget before emitting valid JSON and were excluded from training; evaluation on the human-labeled test set is unaffected.

\textbf{Ablations.} Two variants isolate the contributions of the decomposed interface (GPT-5.2 holistic; prompt App.~\ref{app:gpt-prompt-holistic}, results App.~\ref{app:interface-ablation}) and of reasoning (Qwen3.5 reasoning disabled; results App.~\ref{app:reasoning-ablation}). We refer to the four variants throughout as \textbf{GPT-D}, \textbf{Qwen-DT}, \textbf{GPT-H}, and \textbf{Qwen-DNT}, and to the underlying models as GPT-5.2 and Qwen3.5 for short.

\begin{table}[t]
\centering
\small
\begin{tabular}{lccc}
\toprule
\textbf{LLM} & \textbf{Interface / mode} & \textbf{ANTI} & \textbf{\%} \\
\midrule
GPT-5.2  & D                  & 5{,}651 & 21.8 \\
Qwen3.5  & D + thinking       & 7{,}961 & 30.6 \\
GPT-5.2  & H                  & 7{,}847 & 30.2 \\
Qwen3.5  & D, no-thinking     & 8{,}315 & 32.0 \\
\bottomrule
\end{tabular}
\caption{\textsc{anti-immigrant} label counts and rates per LLM variant on the 25{,}974-comment prefiltered pool. Top two rows: main annotators; bottom two: ablations (App.~\ref{app:interface-ablation}, App.~\ref{app:reasoning-ablation}). Per-party breakdowns: App.~\ref{app:label-distributions}.}
\label{tab:llm_labels}
\vspace{-5mm}
\end{table}


\paragraph{Prefilter validation.}
To verify that the Stage~1 filter does not systematically discard relevant content, we re-annotated a party-stratified sample of 500 Llama-excluded comments with GPT-D. All 500 received Q1\,=\,NO (no immigration reference) and were labeled \textsc{not anti-immigrant}, indicating that the prefilter discards off-topic content rather than dropping \textsc{anti-immigrant} cases. This check is itself LLM-based; an LLM-independent validation,
e.g.\ human annotation of a sample of excluded comments, would strengthen
it further.
 
\paragraph{Annotation Corpus.}
From the GPT-H labeled pool, we sampled 5{,}000 comments for the human annotation study (\S\ref{sec:human-annotation}) using party-stratified sampling at a 40/60 \textsc{anti}/\textsc{not} target ratio, a moderate enrichment over that variant's 30.2\% base rate to ensure sufficient positive instances for AL. The same 5{,}000 comments are covered by all four LLM variants, enabling matched-pool comparison across sources. 

\enote{hs}{for the next paper, consider using present tense
instead of past tense, e.g., ``we sample'' instead of ``we
sampled''. this sounds more natural in English and saves space}

\enote{hs}{you can also save space by getting rid of the
oxford comma. this is preferred by many AE writers. (except
of course in the extremely rare cases where the oxford comma
improves readability and/or disambiguates}

\subsection{Human Annotation Study}
\label{sec:human-annotation}


We recruited annotators through Prolific, restricted to
proficient German speakers. Of eight recruited, six were
retained after qualification. Annotations were collected via
argilla.io on Hugging Face Spaces using the two-question
interface (\S\ref{sec:label_space}). The study comprised
three phases with decreasing overlap: a 50-item
qualification test (6-way, i.e.\ all six annotators label every item), a 200-item pilot (6-way), and a 4{,}750-item main round (950 triple-annotated, 3{,}800
single-annotated). The 1{,}200 multiply-annotated items
(test + pilot + triple-annotated main) form the \textbf{gold
evaluation set}; the 3{,}800 single-annotated items form
the \textbf{AL training pool}.
See App.~\ref{app:annotation-details} for
qualification criteria, attention checks, and item assignment.

\enote{hs}{bolding above: to make it easier for the reader
to find this in case they have questions later about the
exact construction of these sets}

Inter-annotator agreement
is moderate: Krippendorff's $\alpha = 0.49$ on the pilot (6-way, 200 items) and $\alpha = 0.43$ on the main triple-annotated items (3-way, 950 items). Agreement is higher on Q1 (reference, $\alpha = 0.57$--$0.59$) than on Q2 (valence, $\alpha = 0.43$--$0.49$): disagreement concentrates on valence rather than topic identification, consistent with prior work: $\alpha = 0.38$ for German refugee-crisis tweets~\citep{ross2017measuring}, Fleiss' $\kappa = 0.40$ with trained annotators on TikTok~\citep{fillies2025german}, and raw agreement of 0.70--0.83 in HatEval~\citep{basile2019semeval}, comparable to our class-conditional 81--98\% (App.~\ref{app:class-agreement}). Agreement is also strongly topic-dependent, from 56--69\% pairwise on contested clusters to over 90\% on clear-cut ones (App.~\ref{app:annotator-topics}). Hard labels are aggregated via majority vote (full pairwise agreement in App.~\ref{app:annotation-details}); the soft-label evaluation in \S\ref{sec:model-training} uses Bayesian posteriors over the raw annotator votes.

To assess whether the crowd majority is reliable enough to serve as the evaluation standard, an independent domain expert re-annotated a stratified 10\% sample of the gold set (120 items) using the same two-question interface, blind to the crowd labels. The expert agrees with the crowd majority on 88.3\% of items (Cohen's $\kappa = 0.73$), rising to 96.3\% on items the crowd annotated unanimously. Measured against individual annotators rather than the aggregate, the expert's mean pairwise $\kappa$ is 0.65, above the crowd's own mean pairwise $\kappa$ of 0.47 (pooled across phases on the 1{,}200 gold items; per-phase values in App.~\ref{app:annotation-details}). The expert therefore agrees with the crowd workers more than they agree with each other, which is the property that justifies aggregating individual judgements into a majority label. Full results are in App.~\ref{app:expert-validation}.


\subsection{Experimental Conditions}
\label{sec:conditions}

We compare seven conditions (Table~\ref{tab:conditions}) spanning three design axes: annotation source (human or LLM), sampling strategy (AL vs.\ random vs.\ full pool), and label volume. LLM conditions are instantiated under our two LLM annotators (GPT-D and Qwen-DT). Human-vs-LLM at matched sampling isolates label source; AL-vs-Random isolates acquisition value; \textsc{Full-LLM-25K} tests whether scale compensates for label quality. All conditions share the classifier, training recipe, and evaluation set (\S\ref{sec:model-training}); costs in \S\ref{sec:cost}.

\begin{table}[t]
\centering
\small
\begin{tabular}{lcc}
\toprule
\textbf{Condition} & \textbf{Annotator} & \textbf{$n_{\text{train}}$} \\
\midrule
\textsc{AL-Human}     & Human & $\leq$530  \\
\textsc{AL-LLM}       & LLM   & $\leq$530  \\
\textsc{Random-Human} & Human & $\leq$530  \\
\textsc{Random-LLM}   & LLM   & $\leq$530  \\
\textsc{Full-Human}   & Human & 3,800      \\
\textsc{Full-LLM}     & LLM   & 3,800      \\
\textsc{Full-LLM-25K} & LLM   & 24,774     \\
\bottomrule
\end{tabular}
\caption{Seven experimental conditions. Prefixes: \textsc{AL} = active learning, \textsc{Random} = random sampling, \textsc{Full} = full-pool training. AL conditions use four acquisitions (Entropy, BALD, Core-Set, BADGE; \S\ref{sec:al-design}) and report max $n_{\text{train}}$ at round 10 (30 seed + 500 acquired).}
\label{tab:conditions}
\vspace{-5mm}
\end{table}

\subsection{Active Learning Design}
\label{sec:al-design}

We implement pool-based AL for \textsc{AL-Human} and \textsc{AL-LLM}. Both start from a class-balanced, party-stratified seed set of 30 comments (5 per party), following \citet{fairstein2024class}, and simulate the AL loop by revealing pre-existing labels when queried. In each of 10 rounds, the top 50 most informative instances are selected, yielding up to 530 labeled examples at round 10. After each round, the classifier is retrained from scratch.

We evaluate four acquisition strategies: \textsc{Entropy}
and \textsc{BALD}~\citep{houlsby2011bayesian,
gal2016dropout} are uncertainty-based, selecting instances
on which the classifier is least
confident. 
Two are diversity-based,
selecting instances that span encoder
(\textsc{Core-Set}, \citet{sener2018active})
and gradient (\textsc{BADGE}, \citet{ash2020deep}) embedding
spaces. See
App.~\ref{app:al-acquisition} for details.

\subsection{Model, Training, and Evaluation}
\label{sec:model-training}

\paragraph{Models and Training.}
We evaluate four pre-trained transformer encoders:
ModernGBERT, german-bert, gbert-base, and xlm-r-base;
details
in App.~\ref{app:encoder-models}. xlm-r-base is a
multilingual baseline. Each is fine-tuned with a randomly
initialized binary classification head, with comments
tokenized and truncated to 128 tokens. We use
AdamW~\citep{loshchilov2019decoupled} with learning rate
$2 \times 10^{-5}$, batch size 16, weight decay $10^{-2}$,
and linear warmup (10\%) followed by decay. To address class
imbalance, we apply focal loss~\citep{lin2017focal} with
$\gamma = 2.0$ and per-class weights recomputed at each AL
round.
Training runs for up to 10 epochs with early stopping on dev macro-F1 (patience 5 for full conditions and the AL seed round, 3 for subsequent AL rounds).

\paragraph{Evaluation.}
The 1{,}200-item gold evaluation set is partitioned into a development set (15\%, for early stopping) and a test set (85\%, for final reporting), both stratified by class. We report macro-F1 supplemented by class-specific F1 for \textsc{ANTI-IMMIGRANT} (F1-Anti); for AL conditions, we additionally report the Area under the Learning Curve (ALC), normalized by the number of AL rounds. All experiments use 10 random seeds; we report mean and standard deviation. To test robustness to majority-vote aggregation, we additionally evaluate against two Bayesian posteriors over per-item labels~\citep{wright2025aggregating}: \textbf{Beta-Binomial (BB)}, treating annotators as exchangeable Bernoulli voters; and \textbf{Dawid-Skene (DS)}~\citep{dawid1979maximum}, estimating per-annotator competence via EM (details in App.~\ref{app:soft-label-construction}).

\subsection{Cost-Effectiveness Analysis}
\label{sec:cost}

Table~\ref{tab:cost} compares annotation costs across
 conditions. All costs are reported in USD.
Human annotation totaled \$416 across all 5{,}000 items
 (Prolific's ``good rate'' of \pounds9/hour).\footnote{Phase
 breakdown: \pounds18 (test, 20\,min) + \pounds54 (pilot,
 60\,min) + \pounds162 (main, 3\,hours) + \pounds77.92
 platform fee. (\pounds1\,=\,\$1.334)} GPT-5.2 annotation of the full
 25{,}974-comment pool cost \$28 via the OpenAI Batch API,
 plus \$6.14 for Llama prefiltering. Qwen3.5
 ran on local A100 GPUs. Token
 estimates, GPU-hours, and per-LLM breakdowns are in
 App.~\ref{app:infra}. The order-of-magnitude
 ($\sim$10$\times$) gap between human annotation and the
 GPT-5.2 reflects throughput differences rather
 than transient pricing;
see \S\ref{sec:cost-results} for cost-performance tradeoffs. This comparison targets annotation costs that scale with corpus size; design-stage costs are excluded on both sides and were similar by construction, as the guidelines were reused verbatim as the LLM prompt (App.~\ref{app:gpt-prompt-decomposed}).

\begin{table}[t]
\centering
\small
\begin{tabular}{lcrc}
\toprule
\textbf{Condition} & \textbf{$n_{\text{train}}$} &
\textbf{Cost} & \textbf{\$/label} \\
\midrule
\textsc{AL-Human}       & 530    & \$44    & \$0.083 \\
\textsc{AL-LLM}         & 530    & \$0.57  & \$0.001 \\
\textsc{Full-LLM}       & 3{,}800  & \$4.10  & \$0.001 \\
\textsc{Full-Human}     & 3{,}800  & \$316   & \$0.083 \\
\textsc{Full-LLM-25K}   & 24{,}774 & \$28    & \$0.001 \\
\midrule
Full human study        & 5{,}000  & \$416   & \$0.083 \\
\bottomrule
\end{tabular}
\caption{Annotation costs by condition.
Human cost scales proportionally from the \$416 study total; LLM costs are GPT-5.2, excluding a one-time \$6.14 for Llama prefiltering. Qwen3.5 costs (local A100s) are in App.~\ref{app:infra}. \textsc{Random} conditions are omitted (same cost as AL counterparts). The \textsc{Full-LLM-25K} cost covers annotation of all 25{,}974 pool comments; the 1{,}200 gold evaluation items are labeled but excluded from training, leaving 24{,}774 training instances.}
\label{tab:cost}
\end{table}

\enote{hs}{table 3: i probably would know the answer if i
had read the paper more carefully, but why is full human
study \$100 more expensive than FULL-HUMAN?}

\section{Results}

We evaluate all seven conditions across four encoders and 10
seeds on the held-out test set. The relative ordering of
annotation strategies is consistent across encoders; tables
report all four backbones and headline figures use
ModernGBERT (full per-backbone results in
App.~\ref{app:results}). LLM-supervised conditions use
GPT-D and Qwen-DT;
the ablations 
interface GPT-H and reasoning-mode Qwen-DNT  are in App.~\ref{app:interface-ablation} and App.~\ref{app:reasoning-ablation}.

\subsection{Annotation Source at Matched Volume and at Scale (RQ1)}
\label{sec:rq1}

\begin{figure}[t]
\centering
\includegraphics[width=\columnwidth]{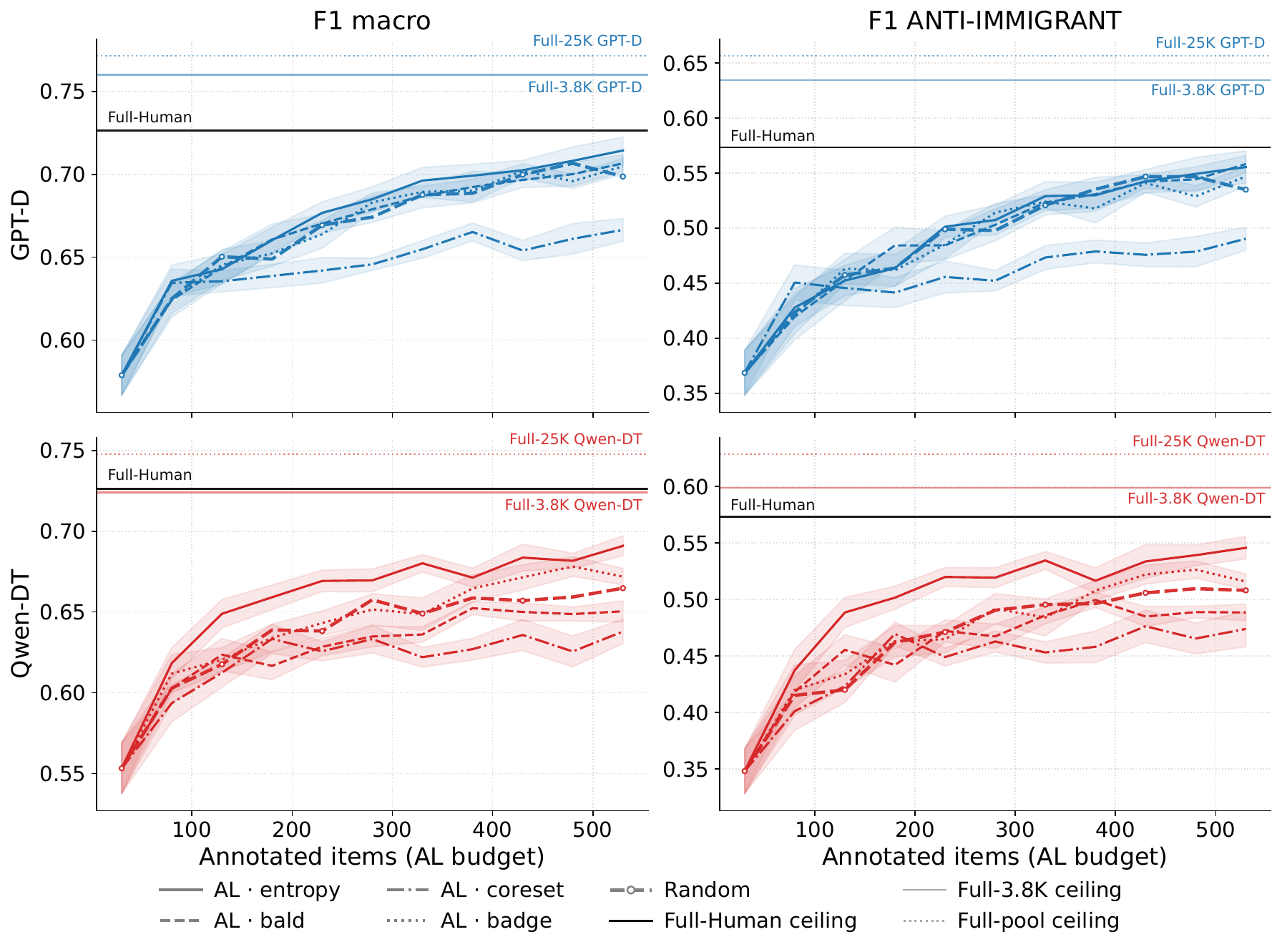}
\caption{ModernGBERT learning curves. Top: GPT-D (blue);
bottom: Qwen-DT (red). Horizontal lines: Full-LLM ceilings
at 3.8K and 24{,}774 labels; black: \textsc{Full-Human}.
Shaded bands: $\pm$1 SD across 10 seeds. Results for other encoders in App.~\ref{app:results}.}
\label{fig:curves}
\vspace{-5mm}
\end{figure}

\paragraph{Key finding.}
At matched volume (3{,}800 labels), \textsc{Full-LLM} under GPT-D matches or slightly exceeds \textsc{Full-Human}. At scale (24{,}774 labels),
\textsc{Full-LLM-25K} under GPT-D exceeds \textsc{Full-Human} in F1-Macro on
three of four encoders, and both LLM annotators improve minority-class F1.

On 3{,}800 labels, Full-LLM under GPT-D matches or slightly exceeds Full-Human
across all four backbones (Table~\ref{tab:section4_full_f1}; learning curves in
Figure~\ref{fig:curves}); Full-Human slightly outperforms Qwen-DT at this
volume. Scaling to 24{,}774 labels improves both LLM annotators, but not
uniformly: Full-LLM-25K under GPT-D gains +0.024 to +0.045 F1-Macro over
Full-Human on ModernGBERT, gbert-base, and xlm-r-base, while on german-bert
Full-Human remains ahead (0.740 vs.\ 0.718). Qwen-DT is at parity in F1-Macro
($-0.011$ to $+0.021$). The minority class is more consistent: F1-Anti exceeds
Full-Human under both annotators on every backbone except german-bert under
GPT-D (GPT-D 0.638--0.656 on the other three vs.\ Full-Human 0.573--0.599;
Qwen-DT 0.600--0.629 vs.\ 0.573--0.603). At $\sim$6.5$\times$ the volume and
one-tenth the cost (\S\ref{sec:cost-results}), LLM annotation is competitive
with human supervision on F1-Macro and stronger on the minority class.

\begin{table}[t]
\centering
\footnotesize
\setlength{\tabcolsep}{3.5pt}
\begin{tabular}{lcccc}
\toprule
\textbf{Condition} & \textbf{germ.} & \textbf{Mod.} & \textbf{gbert} & \textbf{xlm-r} \\
\midrule
\multicolumn{5}{c}{\textit{F1-Macro}} \\
\textsc{Full-Human}                       & \textbf{.740} & .726 & .735 & .725 \\
\textsc{Full-LLM}, GPT-D                  & .736 & \underline{.760} & \underline{.736} & .724 \\
\textsc{Full-LLM-25K}, GPT-D              & .718 & \textbf{.771} & \textbf{.759} & \textbf{.752} \\
\textsc{Full-LLM}, Qwen-DT                & .708 & .724 & .716 & .692 \\
\textsc{Full-LLM-25K}, Qwen-DT            & \underline{.739} & .748 & .724 & \underline{.730} \\
\midrule
\textsc{AL-Human} (Ent.)                  & .714 & .703 & .706 & .704 \\
\textsc{AL-LLM}, GPT-D (Ent.)             & .713 & .714 & .720 & .677 \\
\textsc{AL-LLM}, Qwen-DT (Ent.)           & .662 & .691 & .678 & .660 \\
\midrule
\textsc{Random-Human}                     & .695 & .677 & .690 & .702 \\
\textsc{Random-LLM}, GPT-D                & .713 & .699 & .701 & .703 \\
\textsc{Random-LLM}, Qwen-DT              & .671 & .665 & .672 & .662 \\
\midrule
\multicolumn{5}{c}{\textit{F1-Anti Immigrant}} \\
\textsc{Full-Human}                       & .603 & .573 & .599 & .594 \\
\textsc{Full-LLM}, GPT-D                  & \underline{.609} & \underline{.635} & \underline{.610} & .596 \\
\textsc{Full-LLM-25K}, GPT-D              & .560 & \textbf{.656} & \textbf{.638} & \textbf{.621} \\
\textsc{Full-LLM}, Qwen-DT                & .583 & .599 & .596 & .563 \\
\textsc{Full-LLM-25K}, Qwen-DT            & \textbf{.621} & .629 & .600 & \underline{.602} \\
\midrule
\textsc{AL-Human} (Ent.)                  & .554 & .537 & .545 & .562 \\
\textsc{AL-LLM}, GPT-D (Ent.)             & .561 & .555 & .577 & .501 \\
\textsc{AL-LLM}, Qwen-DT (Ent.)           & .516 & .546 & .538 & .486 \\
\midrule
\textsc{Random-Human}                     & .527 & .492 & .529 & .548 \\
\textsc{Random-LLM}, GPT-D                & .569 & .535 & .552 & .564 \\
\textsc{Random-LLM}, Qwen-DT              & .520 & .508 & .531 & .514 \\
\bottomrule
\end{tabular}
\caption{F1-Macro and F1-Anti at the final AL round (mean across 10 seeds). \textbf{Bold}/\underline{underline} = best/second-best per encoder. AL conditions report Entropy; Random samples uniformly.}
\label{tab:section4_full_f1}
\vspace{-5mm}
\end{table}

\subsection{Cost-Effectiveness}
\label{sec:cost-results}
 
\paragraph{Key finding.}
Full-LLM-25K under GPT-D achieves the strongest performance on three out of four
encoders at roughly one-tenth the cost of Full-Human.
 
\begin{figure}[t]
    \centering
    \includegraphics[width=\columnwidth]{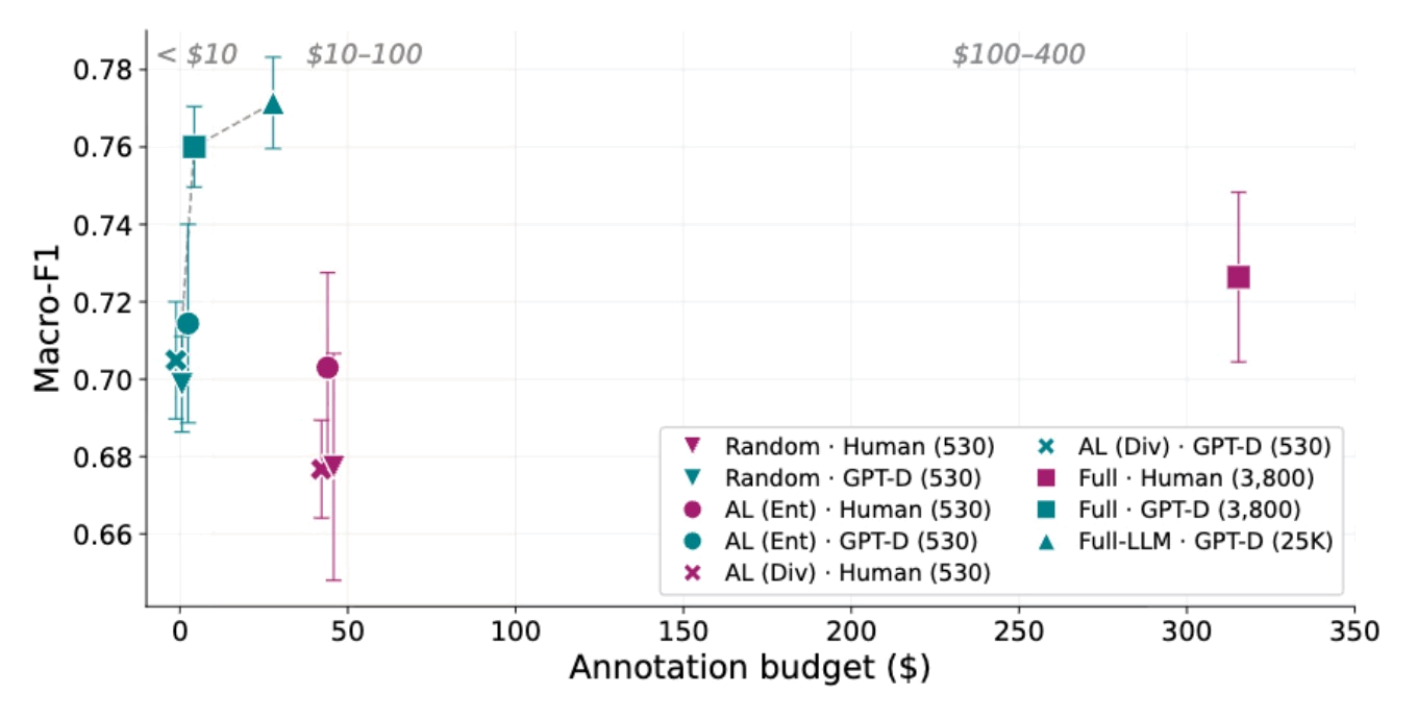}
    \caption{Annotation cost--performance frontier for ModernGBERT (mean $\pm$ 1 SD across 10 seeds). Three budget regimes: under \$10 (LLM-only), \$10--100 (LLM at scale), and \$100--400 (human annotation).}
    \label{fig:budget}
\vspace{-5mm}    
\end{figure}

Figure~\ref{fig:budget} shows the cost-performance frontier for ModernGBERT.
Three budget regimes: (i)~Under \$10, only LLM annotation is feasible;
Full-LLM at matched volume costs \$4 (F1-Macro 0.760). (ii)~At \$29,
Full-LLM-25K reaches 0.771. (iii)~At \$316, Full-Human reaches 0.726. On this
encoder Full-LLM-25K dominates Full-Human at every cost level; on german-bert
Full-Human remains ahead of Full-LLM-25K (\S\ref{sec:rq1}).

\enote{hs}{above: ``is dominated'' by what? (maybe clear
from context}

\subsection{AL vs.\ Random Sampling (RQ2)}
\label{sec:rq2}

\paragraph{Key finding.}
No acquisition strategy reliably outperforms random
 sampling.
 This holds for all four acquisition functions under both LLM and human annotators.

\enote{hs}{why do you say ``within'' here? are they
consistently worse or not?}

The four acquisition strategies cluster within 0.04 ALC of random sampling across encoders and label sources, with no consistent direction (App.~\ref{app:alc-full}). Entropy edges above random most consistently: across the three source variants (Human, GPT-D, Qwen-DT), it produces the highest ALC in 11 of 12 cells, though the gap remains small (AL-Human: 0.658 vs.\ 0.647; AL-LLM GPT-D: 0.671 vs.\ 0.664; AL-LLM Qwen-DT: 0.635 vs.\ 0.628). Diversity-based methods diverge: BADGE consistently outperforms Core-Set by 0.01--0.04 ALC, but neither reliably beats random sampling.

\paragraph{AL vs.\ LLM at matched cost.}
AL-Human at round 10 acquires 530 labels at \$44, while Full-LLM-25K (GPT-D)
reaches 0.771 F1-Macro at \$29: 47$\times$ more labels at comparable cost,
with F1-Macro higher by 0.048--0.068 on three of four encoders and comparable
on german-bert. AL reduces the cost of human annotation but not below LLM at
scale.

\subsection{Soft-Label Robustness (RQ3)}
\label{sec:rq3}
 
\paragraph{Key finding.}
Soft-label evaluation against Beta-Binomial and Dawid-Skene posteriors reproduces the hard-label pattern: \textsc{Full-LLM-25K} exceeds \textsc{Full-Human} on the same three encoders and falls short on german-bert, indicating the result is not an artifact of majority-vote aggregation.

The hard-label result of \S\ref{sec:rq1} could be questioned: majority-vote ground truth obscures annotator disagreement on subjective items. We address this by evaluating against two Bayesian posteriors of the per-item label (\S\ref{sec:model-training}) that weight items by annotator agreement. Both posteriors agree with the hard-label ranking encoder by encoder (App.~\ref{app:soft-eval-full}). Under DS, Full-LLM-25K (GPT-D) exceeds Full-Human by 0.024--0.057 expected macro-F1 on ModernGBERT, gbert-base, and xlm-r-base, and trails by 0.014 on german-bert; the BB gaps follow the same pattern with smaller magnitudes (+0.003 to +0.032). Full-LLM-25K (Qwen-DT) shows the same direction with smaller magnitudes still.

 

\subsection{Error Structure and Confidence (RQ4)}
\label{sec:rq4-errors}

\paragraph{Key finding.}
GPT-D's classifiers match the human FP:FN balance, while the other LLM
variants run moderately more FP-heavy. All LLM-supervised classifiers,
including those with fewer total errors, produce errors at higher confidence
than human-supervised ones.

\begin{figure}[t]
\centering
\includegraphics[width=\columnwidth]{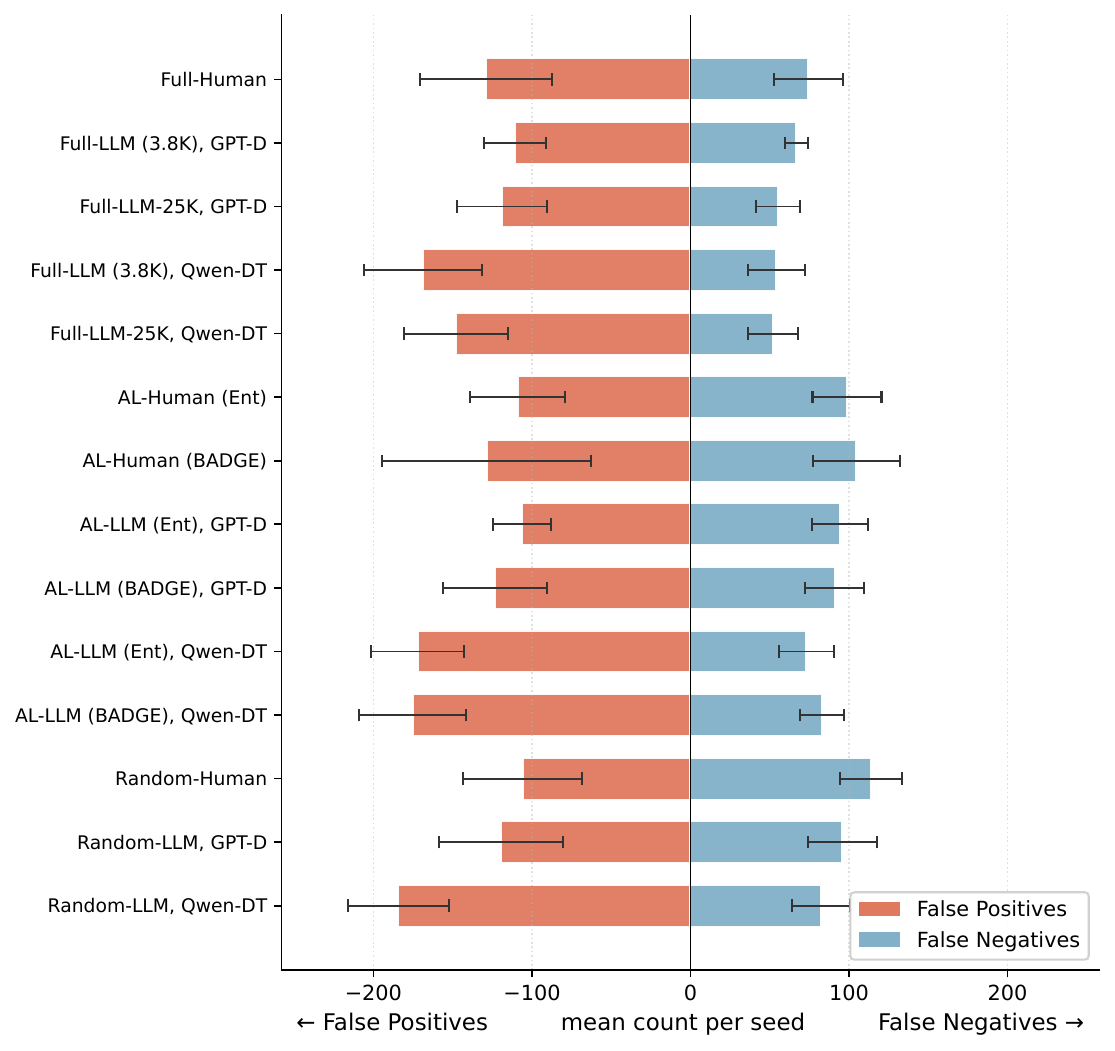}
\caption{FP/FN counts per condition for ModernGBERT (mean $\pm$ 1 SD across 10 seeds; dots = individual seeds). Confidence-on-errors distributions in App.~\ref{app:error-structure}.}
\label{fig:errors}
\vspace{-5mm}
\end{figure}

\paragraph{FP:FN asymmetry.}
Across four encoders (Fig.~\ref{fig:errors}), \textsc{Full-Human} produces
FP:FN ratios of 1.86--2.74 (mean 2.13). \textsc{Full-LLM-25K} (GPT-D) reaches
1.86--2.37 (mean 2.15), indistinguishable from human supervision: FP counts
are comparable (111--124 vs.\ 118--150) and FN counts are lower on two
encoders (55--79 vs.\ 65--77). The minority-class F1 gain of
\S\ref{sec:rq1} is therefore a recall gain, not an FP artifact
(App.~\ref{app:error-structure}). \textsc{Full-LLM-25K} (Qwen-DT) is
moderately more FP-heavy (2.79--3.39, mean 3.23), with FP counts 1.0--1.4$\times$
\textsc{Full-Human} (148--170) and lower FN counts (49--61). The ablations sit
in the same range (\textsc{GPT-H} 2.79--3.72, \textsc{Qwen-DNT} 3.06--3.88;
App.~\ref{app:interface-ablation}, \ref{app:reasoning-ablation}). All LLM
annotators improve minority-class recall; GPT-D does so while preserving the
human error balance, the others at a modest cost in precision.

\paragraph{Confidence on errors.}
Error counts tell only half the story; confidence matters too.
\textsc{Full-Human} assigns mean predicted-class probability 0.627--0.698 to
misclassified instances; \textsc{Full-LLM-25K} (GPT-D) has 0.692--0.751 and
(Qwen-DT) 0.719--0.804. The share of errors with confidence $>0.90$ rises from
2--14\% under \textsc{Full-Human} to 11--30\% (GPT-D) and 21--39\% (Qwen-DT).
LLM-supervised classifiers are thus more confident on the errors they make,
with the effect largest under Qwen-DT. This has practical consequences for
moderation pipelines that triage items by confidence: under LLM-supervised
classifiers, fewer items needing review surface as low-confidence cases. The
ablations show the same pattern (\textsc{GPT-H} 0.685--0.737, \textsc{Qwen-DNT}
0.689--0.776), indicating that confidence inflation tracks LLM supervision in
general rather than a specific interface or reasoning mode.

\paragraph{Annotation boundaries.}
A few instances are persistently misclassified (incorrect under 9 of 10 seeds across \textsc{Full-Human}, GPT-D, and Qwen-DT) and reflect genuine annotation ambiguity rather than model failure. Consider:
\textit{Wir nehmen die Menschen, wie sie sind und nicht wie sie sein sollten. Leitkultur \twemoji{clown face}}
{\small (`We take people as they are and not as they should be. Leitkultur \twemoji{clown face}')}
\noindent This juxtaposes
an emoji signaling
sarcasm with 
a familiar quote
and the term \textit{Leitkultur} (a politically charged concept
implying cultural assimilation). The sarcasm could target \textit{Leitkultur}
(mocking cultural assimilation, \textsc{not
anti-immigrant}) or the acceptance statement
(mocking 
tolerance,
\textsc{anti-immigrant}). All three supervision regimes
commit to the same incorrect reading with high confidence
(.88, .94, .92). \S\ref{sec:topics} examines: do such
cases concentrate in specific discourse themes?

\subsection{Topical Disagreement (RQ4)}
\label{sec:topics}

\paragraph{Key finding.}
Per-topic human-LLM disagreement decomposes into three distinct pathways: an interface-driven component that the decomposed interface (GPT-D) resolves, a topic-intrinsic component that no LLM variant resolves, and a model-specific component on which the two LLM annotators diverge under the same interface.

We embed the 25{,}974 prefiltered comments using a multilingual sentence transformer and apply BERTopic~\citep{grootendorst2022bertopic}, yielding 22 named discourse topics (App.~\ref{app:topics}). Per-topic agreement between human and LLM labels ranges from 37\% to 96\% for the two main annotators; full breakdowns are in App.~\ref{app:topics_full}, and per-annotator human rates by topic in App.~\ref{app:annotator-topics}.

\paragraph{Pathway 1: interface-driven.}
Several large clusters where the holistic interface
over-detected anti-immigrant content are resolved under the
decomposed interface: Welfare \& Economic Competition (gap
+28.7 under GPT-H drops to +0.7 under GPT-D), Rejection
Rhetoric (+31.8 $\rightarrow$ 0.0), Islam \& Islamization
(+21.1 $\rightarrow$ $-$2.6), Directed Personal Hostility
(+22.5 $\rightarrow$ +2.8), and Criminal Violence \& Public
Safety (+21.2 $\rightarrow$ +4.5). The Q1 (reference) / Q2
(valence) decomposition breaks a co-occurrence shortcut the
holistic prompt did not, directly accounting for a
substantial fraction of the F1-Anti and FP:FN gains
in \S\ref{sec:rq1} and \S\ref{sec:rq4-errors}.

\paragraph{Pathways 2 and 3: topic-intrinsic and model-specific.}
Two pathways remain unresolved. Border Control sees gaps of +49.0 (GPT-D), +54.9 (Qwen-DT),
+45.1 (GPT-H), and +62.8 (Qwen-DNT): every LLM variant
over-flags Border content, a topic-intrinsic source where
surface border vocabulary co-occurs with negative valence
regardless of stance; its impact on aggregate metrics is
negligible (App.~\ref{app:border-aggregate}). Independently,
GPT-D and Qwen-DT diverge on metaphor-heavy topics under the same interface: on Rejection Rhetoric ($n{=}22$), GPT-D matches the human rate exactly while Qwen-DT over-detects by 36 percentage points, with the effect extending to large clusters (Welfare under Qwen-DT: gap +15.9). 


\section{Discussion}
\label{sec:discussion}

\subsection{Why the Decomposed Interface Works}
\label{sec:why-decomposed}
The decomposed interface contributes independently of LLM scale, most clearly
at limited label volume: at 3,800 labels GPT-D exceeds GPT-H by 0.018--0.050
F1-Macro on every backbone, while at full scale the two converge to within
0.021. The topical analysis (\S\ref{sec:topics}) shows the mechanism directly at
the label level, before any classifier training: the holistic prompt applies
a co-occurrence shortcut where immigration-adjacent words plus negative
valence trigger an \textsc{anti} label, and the Q1/Q2 decomposition's forced
separate judgment breaks it. Decomposition moves the
GPT \textsc{anti} rate toward the human rate on 17 of 22 discourse clusters
(Welfare 62\%$\rightarrow$34\%, Rejection Rhetoric 36\%$\rightarrow$4\%,
Directed Hostility 35\%$\rightarrow$16\%), raising mean per-cluster agreement
from 74\% to 79\% (App.~\ref{app:topics_full_decomposed}). Policy-critique and
economic-anxiety comments pass Q1 with a \textsc{no} and are correctly
classified. The same decomposition was originally introduced for human
annotation to address low inter-annotator agreement (Fleiss $\kappa = 0.175$
under a holistic prompt; \S\ref{sec:label_space}): the structural argument
applies to both annotator types.

\subsection{The Scale-Quality Trade-off}
At costs relevant for resource-constrained research, LLM-at-scale is favorable
on most encoders: Full-LLM-25K (GPT-D) at \$29 outperforms Full-Human at \$316
on three of four encoders (german-bert is the exception). At a comparable cost
to AL-Human's 530 labels (\$44), Full-LLM-25K provides 47$\times$ more labels
and 0.048--0.068 higher F1-Macro on those three encoders. The pattern holds
under soft-label evaluation (\S\ref{sec:rq3}): Full-LLM-25K (GPT-D)
exceeds Full-Human under both BB and DS posteriors on the same three encoders,
indicating the comparison is not an artifact of majority-vote aggregation.

\subsection{Two LLMs, Two Distinct Error Structures}
Despite similar aggregate F1, the two LLM annotators produce different error
profiles: GPT-D matches the human FP:FN balance (1.86--2.37 vs.\ Full-Human
1.86--2.74; means 2.15 and 2.13), while Qwen-DT runs moderately more FP-heavy
(2.79--3.39) and over-flags metaphor-heavy topics. Both LLMs also produce
errors at higher confidence than human-supervised classifiers: the share of
errors above 0.90 confidence rises from 2--14\% under Full-Human to 11--30\%
(GPT-D) and 21--39\% (Qwen-DT). This impacts deployment: pipelines that triage
low-confidence predictions for human review will surface fewer errors needing
review for classifiers trained on LLM labels. The topical analysis
(\S\ref{sec:topics}) adds two qualifications: Border Control content is
over-flagged by all four LLMs (a topic-intrinsic limitation no interface or
model choice resolves), and GPT-D and Qwen-DT diverge by 25--36 points on
metaphor-heavy clusters. The choice of LLM, not only the interface, materially
affects which discourse themes are flagged.

\subsection{Active Learning in Pre-Enriched Pools}

\enote{hs}{i would tone the following paragraph down: what
you find about the underperforamnce of active learning may
not generalize to other settings}

The Llama-prefiltered pool already concentrates task-relevant content, reducing the marginal benefit of acquisition: Core-Set and BADGE in particular fared worse than random, consistent with the absence of meaningful diversity. When pool construction does relevance work, acquisition complexity provides limited additional benefit, though this may not generalize to unfiltered pools where relevance is itself the acquisition target.

\section{Conclusion}
\label{sec:conclusion}
We asked whether LLM annotation at scale can replace human-in-the-loop AL for
training subjective classifiers. Under matched interfaces, the answer is
largely yes: on three of four encoders, LLM annotation produces stronger
classifiers than human supervision at one-tenth the cost, a pattern that holds
under soft-label evaluation as well as majority-vote labels, and both a
closed-source (GPT-5.2) and an open-weight (Qwen3.5) LLM improve
minority-class F1. Three conditions on this finding matter for practitioners.
Interface is the prerequisite: decomposed unlocks the cost advantage that
holistic does not. The LLM choice shapes error structure: GPT-D matches the
human FP:FN balance, while the other variants run more FP-heavy. And AL
provides little advantage over random sampling in prefiltered pools. Humans
remain essential as gold-standard evaluators; for training labels at scale, the
question has shifted from ``human labels or LLM labels?'' to ``which LLM, which
interface, and what shape of pool?''.

\section{Limitations}
\label{sec:limitations}

\paragraph{Pre-enriched annotation pool.}
Our pool is constructed via LLM-based topic prefiltering, which concentrates content around the target domain and reduces the marginal benefit of uncertainty- and diversity-based acquisition. Our AL-versus-random finding (\S\ref{sec:rq2}) should be interpreted in this context and may not transfer to unfiltered pools.

\paragraph{LLM-derived sampling of the annotation pool.}
The 5{,}000-comment human annotation sample, and hence the gold set, was
enriched using GPT-H labels (\S\ref{sec:llm-pool}), so item selection is not
independent of LLM judgments. Three observations bound this concern: the
enrichment variant is the holistic ablation, not the headline GPT-D
annotator; the crowd re-annotated every item from scratch, confirming only
49\% of GPT-H's and 61\% of GPT-D's positive labels; and Qwen-DT, absent
from pool construction, also outperforms \textsc{Full-Human}
(\S\ref{sec:rq1}). An independently sampled pool could still shift absolute
scores and topic composition; our observations support the relative
human-versus-LLM ordering.

\paragraph{Simulated active learning.}
We simulate the AL loop by pre-collecting labels and revealing them when queried, standard practice in pool-based AL evaluation. A live deployment may differ if annotators calibrate across rounds.

\paragraph{Single task and language.}
All experiments concern a single binary task in German on TikTok. The three-pathway taxonomy (\S\ref{sec:topics}) may transfer to other subjective tasks, but the specific clusters that fall into each pathway will not.

\paragraph{Text-only annotation.}
TikTok comments respond to a video, but humans, LLMs, and the trained classifiers all operate on comment text alone. This is partly by design, matching annotation to deployment conditions and keeping the human--LLM comparison symmetric, and partly by necessity, as the TikTok Research API provides comment text and video metadata but not the video media itself. Some comments are only fully interpretable with the video, which bounds the ecological validity of any text-only label; the expert's notes (App.~\ref{app:expert-validation}) illustrate this.

\paragraph{Two LLM annotators.}
Stage 2 annotation uses two frontier LLMs (closed-source GPT-5.2 and open-weight Qwen3.5). The model-specific pathway in \S\ref{sec:topics} relies on this two-LLM contrast; additional models or alternative interface designs may surface further variation.

\paragraph{Crowdsourced rather than expert annotation.}
Human annotations were collected from Prolific crowdworkers, not domain experts. Crowdworker judgements may differ systematically from trained experts on borderline cases, and expert annotation could shift the decision boundary and alter the human-versus-LLM comparison.

\paragraph{Evaluation against the human gold standard.}
All conditions are evaluated against the majority-vote human label, which structurally disadvantages classifiers whose decision boundary diverges from the human one. Soft-label evaluation (\S\ref{sec:rq3}) partially addresses this; expert adjudication on the contested clusters in \S\ref{sec:topics} would disentangle genuine model errors from evaluation artifacts.

\paragraph{Annotator demographics.}
Our six annotators are predominantly white, all resident in Germany, and not pre-screened for political orientation (App.~\ref{app:annotation-details}). A more diverse panel might shift the gold standard, particularly on contested topics~\citep{sap2022annotators}.

\paragraph{Single round of LLM annotation.}
Each LLM annotates each comment once at temperature 0. Aggregating multiple samples (e.g., self-consistency) might reduce model-specific topical bias but increases cost; we did not explore this trade-off.

\paragraph{In-distribution evaluation only.}
Our evaluation set is sampled from the same corpus and time window as the training pool. Performance on out-of-distribution data (later time slices, other platforms, other German political contexts) may differ. Whether the three-pathway taxonomy (\S\ref{sec:topics}) transfers to OOD evaluation is an open question.

\paragraph{Cost figures are a snapshot.}
Crowdwork rates and LLM API pricing both evolve; absolute dollar comparisons reflect late-2025 pricing and should be interpreted as illustrative of the current cost structure rather than as fixed ratios.

\section{Ethical Considerations}

\paragraph{Annotator exposure and compensation.}
Our annotation task required crowdworkers to read and classify comments containing xenophobic language, stereotypes, and dehumanizing rhetoric. Annotators were informed about the nature of the content before accepting the task, could pause or withdraw at any time without penalty, and were compensated at Prolific's recommended ``good rate'' of \pounds9/hour regardless of completion speed (full cost breakdown in \S\ref{sec:cost-results} and App.~\ref{app:infra}). We did not use rejection-based payment models. Nonetheless, sustained exposure to hostile content carries documented psychological risks~\citep{steiger2021wellbeing}, and reducing the need for human annotation of harmful content is an explicit motivation for this work.

\paragraph{Data collection and privacy.}
Comments were collected via the TikTok Research API, in accordance with its terms of service for academic research. All comments are publicly posted content. We do not collect or store user profile information beyond the comment text and the party account under which it appeared. Usernames and user IDs are stripped from the released dataset. The annotation study was conducted with informed consent from all annotators.

\paragraph{Risks of automated content classification.}
Classifiers trained to detect anti-immigrant content carry dual-use risks. In moderation contexts, they may improve platform safety; without appropriate human oversight, they may suppress legitimate political speech, including immigration-policy criticism, particularly given the systematic over-detection patterns we document in Sections~\ref{sec:rq4-errors} and~\ref{sec:topics} (high-confidence false positives concentrated on policy-critique content). Our findings on FP\,:\, FN asymmetry and confidence inflation are intended to inform deployment decisions, not to enable unsupervised filtering of political speech. Released code and data are intended for research; downstream deployment should include human review of flagged items, particularly in topic clusters with high human-LLM disagreement.

\paragraph{Use of generative tools.}
We used LLM-based tools to assist with literature retrieval in the related work section and as coding assistants during experiment implementation. All generated content was verified by the authors.

\section*{Acknowledgments}

This research was supported by the German Research Foundation (DFG, grant SCHU 2246/14-1). Ahmad Dawar Hakimi acknowledges travel support from ELIAS (GA no 101120237) and ELSA (GA no 101070617). We would like to thank the members of the SchützeLab, CopeNLU and MaiNLP for their valuable feedback and discussions, and Leonor Veloso and Greta Warren for their insights on the setup and design of the human annotation study.

\bibliography{custom}

@article{settles2009active,
  title={Active learning literature survey},
  author={Settles, Burr},
  year={2009},
  publisher={University of Wisconsin-Madison Department of Computer Sciences}
}

@article{schroeder2020survey,
  author       = {Christopher Schr{\"{o}}der and
                  Andreas Niekler},
  title        = {A Survey of Active Learning for Text Classification using Deep Neural
                  Networks},
  journal      = {CoRR},
  volume       = {abs/2008.07267},
  year         = {2020},
  url          = {https://arxiv.org/abs/2008.07267},
  eprinttype   = {arXiv},
  eprint       = {2008.07267},
  bibsource    = {dblp computer science bibliography, https://dblp.org}
}

@inproceedings{romberg2025have,
  author       = {Julia Romberg and
                  Christopher Schr{\"{o}}der and
                  Julius Gonsior and
                  Katrin Tomanek and
                  Fredrik Olsson},
  editor       = {Vera Demberg and
                  Kentaro Inui and
                  Llu{\'{\i}}s Marquez},
  title        = {Reassessing Active Learning Adoption in Contemporary {NLP:} {A} Community
                  Survey},
  booktitle    = {Proceedings of the 19th Conference of the European Chapter of the
                  Association for Computational Linguistics, {EACL} 2026 - Volume 1:
                  Long Papers, Rabat, Morocco, March 24-29, 2026},
  pages        = {2621--2647},
  publisher    = {Association for Computational Linguistics},
  year         = {2026},
  url          = {https://aclanthology.org/2026.eacl-long.120/},
  bibsource    = {dblp computer science bibliography, https://dblp.org}
}

@article{houlsby2011bayesian,
  author       = {Neil Houlsby and
                  Ferenc Huszar and
                  Zoubin Ghahramani and
                  M{\'{a}}t{\'{e}} Lengyel},
  title        = {Bayesian Active Learning for Classification and Preference Learning},
  journal      = {CoRR},
  volume       = {abs/1112.5745},
  year         = {2011},
  url          = {http://arxiv.org/abs/1112.5745},
  eprinttype   = {arXiv},
  eprint       = {1112.5745},
  bibsource    = {dblp computer science bibliography, https://dblp.org}
}

@inproceedings{xia2025selection,
  author       = {Yu Xia and
                  Subhojyoti Mukherjee and
                  Zhouhang Xie and
                  Junda Wu and
                  Xintong Li and
                  Ryan Aponte and
                  Hanjia Lyu and
                  Joe Barrow and
                  Hongjie Chen and
                  Franck Dernoncourt and
                  Branislav Kveton and
                  Tong Yu and
                  Ruiyi Zhang and
                  Jiuxiang Gu and
                  Nesreen K. Ahmed and
                  Yu Wang and
                  Xiang Chen and
                  Hanieh Deilamsalehy and
                  Sungchul Kim and
                  Zhengmian Hu and
                  Yue Zhao and
                  Nedim Lipka and
                  Seunghyun Yoon and
                  Ting{-}Hao Kenneth Huang and
                  Zichao Wang and
                  Puneet Mathur and
                  Soumyabrata Pal and
                  Koyel Mukherjee and
                  Zhehao Zhang and
                  Namyong Park and
                  Thien Huu Nguyen and
                  Jiebo Luo and
                  Ryan A. Rossi and
                  Julian J. McAuley},
  editor       = {Wanxiang Che and
                  Joyce Nabende and
                  Ekaterina Shutova and
                  Mohammad Taher Pilehvar},
  title        = {From Selection to Generation: {A} Survey of LLM-based Active Learning},
  booktitle    = {Proceedings of the 63rd Annual Meeting of the Association for Computational
                  Linguistics (Volume 1: Long Papers), {ACL} 2025, Vienna, Austria,
                  July 27 - August 1, 2025},
  pages        = {14552--14569},
  publisher    = {Association for Computational Linguistics},
  year         = {2025},
  url          = {https://aclanthology.org/2025.acl-long.708/},
  bibsource    = {dblp computer science bibliography, https://dblp.org}
}

@article{gilardi2023chatgpt,
  author       = {Fabrizio Gilardi and
                  Meysam Alizadeh and
                  Ma{\"{e}}l Kubli},
  title        = {ChatGPT Outperforms Crowd-Workers for Text-Annotation Tasks},
  journal      = {CoRR},
  volume       = {abs/2303.15056},
  year         = {2023},
  url          = {https://doi.org/10.48550/arXiv.2303.15056},
  doi          = {10.48550/ARXIV.2303.15056},
  eprinttype   = {arXiv},
  eprint       = {2303.15056},
  bibsource    = {dblp computer science bibliography, https://dblp.org}
}

@article{tornberg2024llms,
  title={Large language models outperform expert coders and supervised classifiers at annotating political social media messages},
  author={T{\"o}rnberg, Petter},
  journal={Social Science Computer Review},
  volume={43},
  number={6},
  pages={1181--1195},
  year={2025},
  publisher={Sage Publications Sage CA: Los Angeles, CA}
}

@article{alizadeh2025opensource,
  author       = {Meysam Alizadeh and
                  Ma{\"{e}}l Kubli and
                  Zeynab Samei and
                  Shirin Dehghani and
                  Mohammadmasiha Zahedivafa and
                  Juan Diego Bermeo and
                  Maria Korobeynikova and
                  Fabrizio Gilardi},
  title        = {Open-source LLMs for text annotation: a practical guide for model
                  setting and fine-tuning},
  journal      = {J. Comput. Soc. Sci.},
  volume       = {8},
  number       = {1},
  pages        = {17},
  year         = {2025},
  url          = {https://doi.org/10.1007/s42001-024-00345-9},
  doi          = {10.1007/S42001-024-00345-9},
  bibsource    = {dblp computer science bibliography, https://dblp.org}
}

@inproceedings{he2024annollm,
  author       = {Xingwei He and
                  Zhenghao Lin and
                  Yeyun Gong and
                  A{-}Long Jin and
                  Hang Zhang and
                  Chen Lin and
                  Jian Jiao and
                  Siu Ming Yiu and
                  Nan Duan and
                  Weizhu Chen},
  editor       = {Yi Yang and
                  Aida Mostafazadeh Davani and
                  Avi Sil and
                  Anoop Kumar},
  title        = {AnnoLLM: Making Large Language Models to Be Better Crowdsourced Annotators},
  booktitle    = {Proceedings of the 2024 Conference of the North American Chapter of
                  the Association for Computational Linguistics: Human Language Technologies:
                  Industry Track, {NAACL} 2024, Mexico City, Mexico, June 16-21, 2024},
  pages        = {165--190},
  publisher    = {Association for Computational Linguistics},
  year         = {2024},
  url          = {https://doi.org/10.18653/v1/2024.naacl-industry.15},
  doi          = {10.18653/V1/2024.NAACL-INDUSTRY.15},
  bibsource    = {dblp computer science bibliography, https://dblp.org}
}

@inproceedings{zhang2023llmaaa,
  author       = {Ruoyu Zhang and
                  Yanzeng Li and
                  Yongliang Ma and
                  Ming Zhou and
                  Lei Zou},
  editor       = {Houda Bouamor and
                  Juan Pino and
                  Kalika Bali},
  title        = {LLMaAA: Making Large Language Models as Active Annotators},
  booktitle    = {Findings of the Association for Computational Linguistics: {EMNLP}
                  2023, Singapore, December 6-10, 2023},
  series       = {Findings of {ACL}},
  pages        = {13088--13103},
  publisher    = {Association for Computational Linguistics},
  year         = {2023},
  url          = {https://doi.org/10.18653/v1/2023.findings-emnlp.872},
  doi          = {10.18653/V1/2023.FINDINGS-EMNLP.872},
  bibsource    = {dblp computer science bibliography, https://dblp.org}
}

@inproceedings{waseem2016hateful,
  author       = {Zeerak Waseem and
                  Dirk Hovy},
  title        = {Hateful Symbols or Hateful People? Predictive Features for Hate Speech
                  Detection on Twitter},
  booktitle    = {Proceedings of the Student Research Workshop, SRW@HLT-NAACL 2016,
                  The 2016 Conference of the North American Chapter of the Association
                  for Computational Linguistics: Human Language Technologies, San Diego
                  California, USA, June 12-17, 2016},
  pages        = {88--93},
  publisher    = {The Association for Computational Linguistics},
  year         = {2016},
  url          = {https://doi.org/10.18653/v1/n16-2013},
  doi          = {10.18653/V1/N16-2013},
  bibsource    = {dblp computer science bibliography, https://dblp.org}
}

@article{ross2017measuring,
  author       = {Bj{\"{o}}rn Ross and
                  Michael Rist and
                  Guillermo Carbonell and
                  Benjamin Cabrera and
                  Nils Kurowsky and
                  Michael Wojatzki},
  title        = {Measuring the Reliability of Hate Speech Annotations: The Case of
                  the European Refugee Crisis},
  journal      = {CoRR},
  volume       = {abs/1701.08118},
  year         = {2017},
  url          = {http://arxiv.org/abs/1701.08118},
  eprinttype   = {arXiv},
  eprint       = {1701.08118},
  bibsource    = {dblp computer science bibliography, https://dblp.org}
}

@inproceedings{schmidt2017survey,
  author       = {Anna Schmidt and
                  Michael Wiegand},
  editor       = {Lun{-}Wei Ku and
                  Cheng{-}Te Li},
  title        = {A Survey on Hate Speech Detection using Natural Language Processing},
  booktitle    = {Proceedings of the Fifth International Workshop on Natural Language
                  Processing for Social Media, SocialNLP@EACL 2017, Valencia, Spain,
                  April 3, 2017},
  pages        = {1--10},
  publisher    = {Association for Computational Linguistics},
  year         = {2017},
  url          = {https://doi.org/10.18653/v1/w17-1101},
  doi          = {10.18653/V1/W17-1101},
  bibsource    = {dblp computer science bibliography, https://dblp.org}
}

@article{fortuna2018survey,
  author       = {Paula Fortuna and
                  S{\'{e}}rgio Nunes},
  title        = {A Survey on Automatic Detection of Hate Speech in Text},
  journal      = {{ACM} Comput. Surv.},
  volume       = {51},
  number       = {4},
  pages        = {85:1--85:30},
  year         = {2018},
  url          = {https://doi.org/10.1145/3232676},
  doi          = {10.1145/3232676},
  bibsource    = {dblp computer science bibliography, https://dblp.org}
}

@article{fillies2025german,
  title={A novel german tiktok hate speech dataset: far-right comments against politicians, women, and others},
  author={Fillies, Jan and Theisen, Esther and Hoffmann, Michael and Jung, Robert and Jung, Elena and Fischer, Nele and Paschke, Adrian},
  journal={Discover Data},
  volume={3},
  number={1},
  pages={4},
  year={2025},
  publisher={Springer}
}

@inproceedings{plank2022problem,
  author       = {Barbara Plank},
  editor       = {Yoav Goldberg and
                  Zornitsa Kozareva and
                  Yue Zhang},
  title        = {The "Problem" of Human Label Variation: On Ground Truth in Data, Modeling
                  and Evaluation},
  booktitle    = {Proceedings of the 2022 Conference on Empirical Methods in Natural
                  Language Processing, {EMNLP} 2022, Abu Dhabi, United Arab Emirates,
                  December 7-11, 2022},
  pages        = {10671--10682},
  publisher    = {Association for Computational Linguistics},
  year         = {2022},
  url          = {https://doi.org/10.18653/v1/2022.emnlp-main.731},
  doi          = {10.18653/V1/2022.EMNLP-MAIN.731},
  bibsource    = {dblp computer science bibliography, https://dblp.org}
}

@article{davani2022dealing,
  author       = {Aida Mostafazadeh Davani and
                  Mark D{\'{\i}}az and
                  Vinodkumar Prabhakaran},
  title        = {Dealing with Disagreements: Looking Beyond the Majority Vote in Subjective
                  Annotations},
  journal      = {Trans. Assoc. Comput. Linguistics},
  volume       = {10},
  pages        = {92--110},
  year         = {2022},
  url          = {https://doi.org/10.1162/tacl\_a\_00449},
  doi          = {10.1162/TACL\_A\_00449},
  bibsource    = {dblp computer science bibliography, https://dblp.org}
}

@inproceedings{rottger2022two,
  author       = {Paul R{\"{o}}ttger and
                  Bertie Vidgen and
                  Dirk Hovy and
                  Janet B. Pierrehumbert},
  editor       = {Marine Carpuat and
                  Marie{-}Catherine de Marneffe and
                  Iv{\'{a}}n Vladimir Meza Ru{\'{\i}}z},
  title        = {Two Contrasting Data Annotation Paradigms for Subjective {NLP} Tasks},
  booktitle    = {Proceedings of the 2022 Conference of the North American Chapter of
                  the Association for Computational Linguistics: Human Language Technologies,
                  {NAACL} 2022, Seattle, WA, United States, July 10-15, 2022},
  pages        = {175--190},
  publisher    = {Association for Computational Linguistics},
  year         = {2022},
  url          = {https://doi.org/10.18653/v1/2022.naacl-main.13},
  doi          = {10.18653/V1/2022.NAACL-MAIN.13},
  bibsource    = {dblp computer science bibliography, https://dblp.org}
}

@inproceedings{steiger2021wellbeing,
  author       = {Miriah Steiger and
                  Timir J. Bharucha and
                  Sukrit Venkatagiri and
                  Martin J. Riedl and
                  Matthew Lease},
  editor       = {Yoshifumi Kitamura and
                  Aaron Quigley and
                  Katherine Isbister and
                  Takeo Igarashi and
                  Pernille Bj{\o}rn and
                  Steven Mark Drucker},
  title        = {The Psychological Well-Being of Content Moderators: The Emotional
                  Labor of Commercial Moderation and Avenues for Improving Support},
  booktitle    = {{CHI} '21: {CHI} Conference on Human Factors in Computing Systems,
                  Virtual Event / Yokohama, Japan, May 8-13, 2021},
  pages        = {341:1--341:14},
  publisher    = {{ACM}},
  year         = {2021},
  url          = {https://doi.org/10.1145/3411764.3445092},
  doi          = {10.1145/3411764.3445092},
  bibsource    = {dblp computer science bibliography, https://dblp.org}
}

@article{grootendorst2022bertopic,
  author       = {Maarten Grootendorst},
  title        = {BERTopic: Neural topic modeling with a class-based {TF-IDF} procedure},
  journal      = {CoRR},
  volume       = {abs/2203.05794},
  year         = {2022},
  url          = {https://doi.org/10.48550/arXiv.2203.05794},
  doi          = {10.48550/ARXIV.2203.05794},
  eprinttype   = {arXiv},
  eprint       = {2203.05794},
  bibsource    = {dblp computer science bibliography, https://dblp.org}
}

@inproceedings{kargaran2023glotlid,
  author       = {Amir Hossein Kargaran and
                  Ayyoob Imani and
                  Fran{\c{c}}ois Yvon and
                  Hinrich Sch{\"{u}}tze},
  editor       = {Houda Bouamor and
                  Juan Pino and
                  Kalika Bali},
  title        = {GlotLID: Language Identification for Low-Resource Languages},
  booktitle    = {Findings of the Association for Computational Linguistics: {EMNLP}
                  2023, Singapore, December 6-10, 2023},
  series       = {Findings of {ACL}},
  pages        = {6155--6218},
  publisher    = {Association for Computational Linguistics},
  year         = {2023},
  url          = {https://doi.org/10.18653/v1/2023.findings-emnlp.410},
  doi          = {10.18653/V1/2023.FINDINGS-EMNLP.410},
  bibsource    = {dblp computer science bibliography, https://dblp.org}
}

@inproceedings{lin2017focal,
  title={Focal loss for dense object detection},
  author={Lin, Tsung-Yi and Goyal, Priya and Girshick, Ross and He, Kaiming and Doll{\'a}r, Piotr},
  booktitle={Proceedings of the IEEE international conference on computer vision},
  pages={2980--2988},
  year={2017}
}

@inproceedings{loshchilov2019decoupled,
  author       = {Ilya Loshchilov and
                  Frank Hutter},
  title        = {Decoupled Weight Decay Regularization},
  booktitle    = {7th International Conference on Learning Representations, {ICLR} 2019,
                  New Orleans, LA, USA, May 6-9, 2019},
  publisher    = {OpenReview.net},
  year         = {2019},
  url          = {https://openreview.net/forum?id=Bkg6RiCqY7},
  bibsource    = {dblp computer science bibliography, https://dblp.org}
}

@article{grattafiori2024llama,
  author       = {{Llama Team}},
  title        = {The Llama 3 Herd of Models},
  journal      = {CoRR},
  volume       = {abs/2407.21783},
  year         = {2024},
  url          = {https://doi.org/10.48550/arXiv.2407.21783},
  doi          = {10.48550/ARXIV.2407.21783},
  eprinttype   = {arXiv},
  eprint       = {2407.21783},
  bibsource    = {dblp computer science bibliography, https://dblp.org}
}

@phdthesis{kwon2025vllm,
  title={vLLM: An Efficient Inference Engine for Large Language Models},
  author={Kwon, Woosuk},
  year={2025},
  school={UC Berkeley}
}

@inproceedings{gal2016dropout,
  author       = {Yarin Gal and
                  Zoubin Ghahramani},
  editor       = {Maria{-}Florina Balcan and
                  Kilian Q. Weinberger},
  title        = {Dropout as a Bayesian Approximation: Representing Model Uncertainty
                  in Deep Learning},
  booktitle    = {Proceedings of the 33nd International Conference on Machine Learning,
                  {ICML} 2016, New York City, NY, USA, June 19-24, 2016},
  series       = {{JMLR} Workshop and Conference Proceedings},
  pages        = {1050--1059},
  publisher    = {JMLR.org},
  year         = {2016},
  url          = {http://proceedings.mlr.press/v48/gal16.html},
  bibsource    = {dblp computer science bibliography, https://dblp.org}
}

@inproceedings{fairstein2024class,
  title={Class balancing for efficient active learning in imbalanced datasets},
  author={Fairstein, Yaron and Kalinsky, Oren and Karnin, Zohar and Kushilevitz, Guy and Libov, Alexander and Tolmach, Sofia},
  booktitle={Proceedings of The 18th Linguistic Annotation Workshop (LAW-XVIII)},
  pages={77--86},
  year={2024}
}

@inproceedings{kholodna2024llms,
  author       = {Nataliia Kholodna and
                  Sahib Julka and
                  Mohammad Khodadadi and
                  Muhammed Nurullah Gumus and
                  Michael Granitzer},
  editor       = {Albert Bifet and
                  Tomas Krilavicius and
                  Ioanna Miliou and
                  Slawomir Nowaczyk},
  title        = {LLMs in the Loop: Leveraging Large Language Model Annotations for
                  Active Learning in Low-Resource Languages},
  booktitle    = {Machine Learning and Knowledge Discovery in Databases. Applied Data
                  Science Track - European Conference, {ECML} {PKDD} 2024, Vilnius,
                  Lithuania, September 9-13, 2024, Proceedings, Part {X}},
  series       = {Lecture Notes in Computer Science},
  pages        = {397--412},
  publisher    = {Springer},
  year         = {2024},
  url          = {https://doi.org/10.1007/978-3-031-70381-2\_25},
  doi          = {10.1007/978-3-031-70381-2\_25},
  bibsource    = {dblp computer science bibliography, https://dblp.org}
}

@inproceedings{lowell2019practical,
  author       = {David Lowell and
                  Zachary C. Lipton and
                  Byron C. Wallace},
  editor       = {Kentaro Inui and
                  Jing Jiang and
                  Vincent Ng and
                  Xiaojun Wan},
  title        = {Practical Obstacles to Deploying Active Learning},
  booktitle    = {Proceedings of the 2019 Conference on Empirical Methods in Natural
                  Language Processing and the 9th International Joint Conference on
                  Natural Language Processing, {EMNLP-IJCNLP} 2019, Hong Kong, China,
                  November 3-7, 2019},
  pages        = {21--30},
  publisher    = {Association for Computational Linguistics},
  year         = {2019},
  url          = {https://doi.org/10.18653/v1/D19-1003},
  doi          = {10.18653/V1/D19-1003},
  bibsource    = {dblp computer science bibliography, https://dblp.org}
}

@article{pangakis2023automated,
  author       = {Nicholas Pangakis and
                  Samuel Wolken and
                  Neil Fasching},
  title        = {Automated Annotation with Generative {AI} Requires Validation},
  journal      = {CoRR},
  volume       = {abs/2306.00176},
  year         = {2023},
  url          = {https://doi.org/10.48550/arXiv.2306.00176},
  doi          = {10.48550/ARXIV.2306.00176},
  eprinttype   = {arXiv},
  eprint       = {2306.00176},
  bibsource    = {dblp computer science bibliography, https://dblp.org}
}

@inproceedings{kostikova2024fine,
  author       = {Aida Kostikova and
                  Dominik Beese and
                  Benjamin Paassen and
                  Ole P{\"{u}}tz and
                  Gregor Wiedemann and
                  Steffen Eger},
  editor       = {Yaser Al{-}Onaizan and
                  Mohit Bansal and
                  Yun{-}Nung Chen},
  title        = {Fine-Grained Detection of Solidarity for Women and Migrants in 155
                  Years of German Parliamentary Debates},
  booktitle    = {Proceedings of the 2024 Conference on Empirical Methods in Natural
                  Language Processing, {EMNLP} 2024, Miami, FL, USA, November 12-16,
                  2024},
  pages        = {5884--5907},
  publisher    = {Association for Computational Linguistics},
  year         = {2024},
  url          = {https://doi.org/10.18653/v1/2024.emnlp-main.337},
  doi          = {10.18653/V1/2024.EMNLP-MAIN.337},
  bibsource    = {dblp computer science bibliography, https://dblp.org}
}

@article{vallejo2025llms,
  title={LLMs as annotators: the effect of party cues on labelling decisions by large language models},
  author={Vallejo Vera, Sebasti{\'a}n and Driggers, Hunter},
  journal={Humanities and Social Sciences Communications},
  volume={12},
  number={1},
  pages={1--11},
  year={2025},
  publisher={Palgrave}
}

@inproceedings{sap2020social,
  author       = {Maarten Sap and
                  Saadia Gabriel and
                  Lianhui Qin and
                  Dan Jurafsky and
                  Noah A. Smith and
                  Yejin Choi},
  editor       = {Dan Jurafsky and
                  Joyce Chai and
                  Natalie Schluter and
                  Joel R. Tetreault},
  title        = {Social Bias Frames: Reasoning about Social and Power Implications
                  of Language},
  booktitle    = {Proceedings of the 58th Annual Meeting of the Association for Computational
                  Linguistics, {ACL} 2020, Online, July 5-10, 2020},
  pages        = {5477--5490},
  publisher    = {Association for Computational Linguistics},
  year         = {2020},
  url          = {https://doi.org/10.18653/v1/2020.acl-main.486},
  doi          = {10.18653/V1/2020.ACL-MAIN.486},
  bibsource    = {dblp computer science bibliography, https://dblp.org}
}

@article{wright2025aggregating,
  title={Aggregating soft labels from crowd annotations improves uncertainty estimation under distribution shift},
  author={Wright, Dustin and Augenstein, Isabelle},
  journal={PLoS One},
  volume={20},
  number={6},
  pages={e0323064},
  year={2025},
  publisher={Public Library of Science San Francisco, CA USA}
}

@article{ni2025llm,
  author       = {Jingwei Ni and
                  Yu Fan and
                  Vil{\'{e}}m Zouhar and
                  Donya Rooein and
                  Alexander Miserlis Hoyle and
                  Mrinmaya Sachan and
                  Markus Leippold and
                  Dirk Hovy and
                  Elliott Ash},
  title        = {Can Large Language Models Capture Human Annotator Disagreements?},
  journal      = {CoRR},
  volume       = {abs/2506.19467},
  year         = {2025},
  url          = {https://doi.org/10.48550/arXiv.2506.19467},
  doi          = {10.48550/ARXIV.2506.19467},
  eprinttype   = {arXiv},
  eprint       = {2506.19467},
  bibsource    = {dblp computer science bibliography, https://dblp.org}
}

@inproceedings{sap2022annotators,
  author       = {Maarten Sap and
                  Swabha Swayamdipta and
                  Laura Vianna and
                  Xuhui Zhou and
                  Yejin Choi and
                  Noah A. Smith},
  editor       = {Marine Carpuat and
                  Marie{-}Catherine de Marneffe and
                  Iv{\'{a}}n Vladimir Meza Ru{\'{\i}}z},
  title        = {Annotators with Attitudes: How Annotator Beliefs And Identities Bias
                  Toxic Language Detection},
  booktitle    = {Proceedings of the 2022 Conference of the North American Chapter of
                  the Association for Computational Linguistics: Human Language Technologies,
                  {NAACL} 2022, Seattle, WA, United States, July 10-15, 2022},
  pages        = {5884--5906},
  publisher    = {Association for Computational Linguistics},
  year         = {2022},
  url          = {https://doi.org/10.18653/v1/2022.naacl-main.431},
  doi          = {10.18653/V1/2022.NAACL-MAIN.431},
  bibsource    = {dblp computer science bibliography, https://dblp.org}
}

@inproceedings{melis2025modular,
    title = "A Modular Taxonomy for Hate Speech Definitions and Its Impact on Zero-Shot {LLM} Classification Performance",
    author = "Melis, Matteo  and
      Lapesa, Gabriella  and
      Assenmacher, Dennis",
    editor = "Calabrese, Agostina  and
      de Kock, Christine  and
      Nozza, Debora  and
      Plaza-del-Arco, Flor Miriam  and
      Talat, Zeerak  and
      Vargas, Francielle",
    booktitle = "Proceedings of the The 9th Workshop on Online Abuse and Harms (WOAH)",
    month = aug,
    year = "2025",
    address = "Vienna, Austria",
    publisher = "Association for Computational Linguistics",
    url = "https://aclanthology.org/2025.woah-1.45/",
    pages = "490--521",
    ISBN = "979-8-89176-105-6"
}

@inproceedings{masud2024hate,
    title = "Hate Personified: Investigating the role of {LLM}s in content moderation",
    author = "Masud, Sarah  and
      Singh, Sahajpreet  and
      Hangya, Viktor  and
      Fraser, Alexander  and
      Chakraborty, Tanmoy",
    editor = "Al-Onaizan, Yaser  and
      Bansal, Mohit  and
      Chen, Yun-Nung",
    booktitle = "Proceedings of the 2024 Conference on Empirical Methods in Natural Language Processing",
    month = nov,
    year = "2024",
    address = "Miami, Florida, USA",
    publisher = "Association for Computational Linguistics",
    url = "https://aclanthology.org/2024.emnlp-main.886/",
    doi = "10.18653/v1/2024.emnlp-main.886",
    pages = "15847--15863"
}

@inproceedings{basile2019semeval,
    title = "{S}em{E}val-2019 Task 5: Multilingual Detection of Hate Speech Against Immigrants and Women in {T}witter",
    author = "Basile, Valerio  and
      Bosco, Cristina  and
      Fersini, Elisabetta  and
      Nozza, Debora  and
      Patti, Viviana  and
      Rangel Pardo, Francisco Manuel  and
      Rosso, Paolo  and
      Sanguinetti, Manuela",
    editor = "May, Jonathan  and
      Shutova, Ekaterina  and
      Herbelot, Aurelie  and
      Zhu, Xiaodan  and
      Apidianaki, Marianna  and
      Mohammad, Saif M.",
    booktitle = "Proceedings of the 13th International Workshop on Semantic Evaluation",
    month = jun,
    year = "2019",
    address = "Minneapolis, Minnesota, USA",
    publisher = "Association for Computational Linguistics",
    url = "https://aclanthology.org/S19-2007/",
    doi = "10.18653/v1/S19-2007",
    pages = "54--63"
}

@inproceedings{sanguinetti2018italian,
    title = "An {I}talian {T}witter Corpus of Hate Speech against Immigrants",
    author = "Sanguinetti, Manuela  and
      Poletto, Fabio  and
      Bosco, Cristina  and
      Patti, Viviana  and
      Stranisci, Marco",
    editor = "Calzolari, Nicoletta  and
      Choukri, Khalid  and
      Cieri, Christopher  and
      Declerck, Thierry  and
      Goggi, Sara  and
      Hasida, Koiti  and
      Isahara, Hitoshi  and
      Maegaard, Bente  and
      Mariani, Joseph  and
      Mazo, H{\'e}l{\`e}ne  and
      Moreno, Asuncion  and
      Odijk, Jan  and
      Piperidis, Stelios  and
      Tokunaga, Takenobu",
    booktitle = "Proceedings of the Eleventh International Conference on Language Resources and Evaluation ({LREC} 2018)",
    month = may,
    year = "2018",
    address = "Miyazaki, Japan",
    publisher = "European Language Resources Association (ELRA)",
    url = "https://aclanthology.org/L18-1443/"
}

@inproceedings{karamcheti2021mind,
    title = "Mind Your Outliers! Investigating the Negative Impact of Outliers on Active Learning for Visual Question Answering",
    author = "Karamcheti, Siddharth  and
      Krishna, Ranjay  and
      Fei-Fei, Li  and
      Manning, Christopher",
    editor = "Zong, Chengqing  and
      Xia, Fei  and
      Li, Wenjie  and
      Navigli, Roberto",
    booktitle = "Proceedings of the 59th Annual Meeting of the Association for Computational Linguistics and the 11th International Joint Conference on Natural Language Processing (Volume 1: Long Papers)",
    month = aug,
    year = "2021",
    address = "Online",
    publisher = "Association for Computational Linguistics",
    url = "https://aclanthology.org/2021.acl-long.564/",
    doi = "10.18653/v1/2021.acl-long.564",
    pages = "7265--7281"
}

@inproceedings{snijders2023investigating,
    title = "Investigating Multi-source Active Learning for Natural Language Inference",
    author = "Snijders, Ard  and
      Kiela, Douwe  and
      Margatina, Katerina",
    editor = "Vlachos, Andreas  and
      Augenstein, Isabelle",
    booktitle = "Proceedings of the 17th Conference of the European Chapter of the Association for Computational Linguistics",
    month = may,
    year = "2023",
    address = "Dubrovnik, Croatia",
    publisher = "Association for Computational Linguistics",
    url = "https://aclanthology.org/2023.eacl-main.160/",
    doi = "10.18653/v1/2023.eacl-main.160",
    pages = "2187--2209"
}

@article{engelmann2024annotating,
  author       = {Paul Engelmann and
                  Peter Brunsgaard Trolle and
                  Christian Hardmeier},
  title        = {A Dataset for the Detection of Dehumanizing Language},
  journal      = {CoRR},
  volume       = {abs/2402.08764},
  year         = {2024},
  url          = {https://doi.org/10.48550/arXiv.2402.08764},
  doi          = {10.48550/ARXIV.2402.08764},
  eprinttype   = {arXiv},
  eprint       = {2402.08764},
  bibsource    = {dblp computer science bibliography, https://dblp.org}
}

@article{assenmacher2025bilingual,
  author       = {Dennis Assenmacher and
                  Paloma Piot and
                  Katarina Laken and
                  David Jurgens and
                  Claudia Wagner},
  title        = {Beyond the Explicit: {A} Bilingual Dataset for Dehumanization Detection
                  in Social Media},
  journal      = {CoRR},
  volume       = {abs/2510.18582},
  year         = {2025},
  url          = {https://doi.org/10.48550/arXiv.2510.18582},
  doi          = {10.48550/ARXIV.2510.18582},
  eprinttype   = {arXiv},
  eprint       = {2510.18582},
  bibsource    = {dblp computer science bibliography, https://dblp.org}
}

@inproceedings{davidson2017automated,
  author       = {Thomas Davidson and
                  Dana Warmsley and
                  Michael W. Macy and
                  Ingmar Weber},
  title        = {Automated Hate Speech Detection and the Problem of Offensive Language},
  booktitle    = {Proceedings of the Eleventh International Conference on Web and Social
                  Media, {ICWSM} 2017, Montr{\'{e}}al, Qu{\'{e}}bec, Canada,
                  May 15-18, 2017},
  pages        = {512--515},
  publisher    = {{AAAI} Press},
  year         = {2017},
  url          = {https://aaai.org/ocs/index.php/ICWSM/ICWSM17/paper/view/15665},
  bibsource    = {dblp computer science bibliography, https://dblp.org}
}

@inproceedings{antypas2023robust,
    title = "Robust Hate Speech Detection in Social Media: A Cross-Dataset Empirical Evaluation",
    author = "Antypas, Dimosthenis  and
      Camacho-Collados, Jose",
    editor = {Chung, Yi-Ling  and
      R{\"o}ttger, Paul  and
      Nozza, Debora  and
      Talat, Zeerak  and
      Mostafazadeh Davani, Aida},
    booktitle = "The 7th Workshop on Online Abuse and Harms (WOAH)",
    month = jul,
    year = "2023",
    address = "Toronto, Canada",
    publisher = "Association for Computational Linguistics",
    url = "https://aclanthology.org/2023.woah-1.25/",
    doi = "10.18653/v1/2023.woah-1.25",
    pages = "231--242"
}

@inproceedings{lee2024exploring,
    title = "Exploring Cross-Cultural Differences in {E}nglish Hate Speech Annotations: From Dataset Construction to Analysis",
    author = "Lee, Nayeon  and
      Jung, Chani  and
      Myung, Junho  and
      Jin, Jiho  and
      Camacho-Collados, Jose  and
      Kim, Juho  and
      Oh, Alice",
    editor = "Duh, Kevin  and
      Gomez, Helena  and
      Bethard, Steven",
    booktitle = "Proceedings of the 2024 Conference of the North American Chapter of the Association for Computational Linguistics: Human Language Technologies (Volume 1: Long Papers)",
    month = jun,
    year = "2024",
    address = "Mexico City, Mexico",
    publisher = "Association for Computational Linguistics",
    url = "https://aclanthology.org/2024.naacl-long.236/",
    doi = "10.18653/v1/2024.naacl-long.236",
    pages = "4205--4224"
}

@inproceedings{piot2024metahate,
  author       = {Paloma Piot and
                  Patricia Mart{\'{\i}}n{-}Rodilla and
                  Javier Parapar},
  editor       = {Yu{-}Ru Lin and
                  Yelena Mejova and
                  Meeyoung Cha},
  title        = {MetaHate: {A} Dataset for Unifying Efforts on Hate Speech Detection},
  booktitle    = {Proceedings of the Eighteenth International {AAAI} Conference on Web
                  and Social Media, {ICWSM} 2024, Buffalo, New York, USA, June 3-6,
                  2024},
  pages        = {2025--2039},
  publisher    = {{AAAI} Press},
  year         = {2024},
  url          = {https://doi.org/10.1609/icwsm.v18i1.31445},
  doi          = {10.1609/ICWSM.V18I1.31445},
  bibsource    = {dblp computer science bibliography, https://dblp.org}
}

@inproceedings{vidgen2021learning,
    title = "Learning from the Worst: Dynamically Generated Datasets to Improve Online Hate Detection",
    author = "Vidgen, Bertie  and
      Thrush, Tristan  and
      Waseem, Zeerak  and
      Kiela, Douwe",
    editor = "Zong, Chengqing  and
      Xia, Fei  and
      Li, Wenjie  and
      Navigli, Roberto",
    booktitle = "Proceedings of the 59th Annual Meeting of the Association for Computational Linguistics and the 11th International Joint Conference on Natural Language Processing (Volume 1: Long Papers)",
    month = aug,
    year = "2021",
    address = "Online",
    publisher = "Association for Computational Linguistics",
    url = "https://aclanthology.org/2021.acl-long.132/",
    doi = "10.18653/v1/2021.acl-long.132",
    pages = "1667--1682"
}

@inproceedings{sener2018active,
  author       = {Ozan Sener and
                  Silvio Savarese},
  title        = {Active Learning for Convolutional Neural Networks: {A} Core-Set Approach},
  booktitle    = {6th International Conference on Learning Representations, {ICLR} 2018,
                  Vancouver, BC, Canada, April 30 - May 3, 2018, Conference Track Proceedings},
  publisher    = {OpenReview.net},
  year         = {2018},
  url          = {https://openreview.net/forum?id=H1aIuk-RW},
  bibsource    = {dblp computer science bibliography, https://dblp.org}
}

@inproceedings{ash2020deep,
  author       = {Jordan T. Ash and
                  Chicheng Zhang and
                  Akshay Krishnamurthy and
                  John Langford and
                  Alekh Agarwal},
  title        = {Deep Batch Active Learning by Diverse, Uncertain Gradient Lower Bounds},
  booktitle    = {8th International Conference on Learning Representations, {ICLR} 2020,
                  Addis Ababa, Ethiopia, April 26-30, 2020},
  publisher    = {OpenReview.net},
  year         = {2020},
  url          = {https://openreview.net/forum?id=ryghZJBKPS},
  bibsource    = {dblp computer science bibliography, https://dblp.org}
}

@article{dawid1979maximum,
  title={Maximum likelihood estimation of observer error-rates using the EM algorithm},
  author={Dawid, Alexander Philip and Skene, Allan M},
  journal={Journal of the Royal Statistical Society: Series C (Applied Statistics)},
  volume={28},
  number={1},
  pages={20--28},
  year={1979},
  publisher={Wiley Online Library}
}

@inproceedings{arthur2007kmeans,
  author       = {David Arthur and
                  Sergei Vassilvitskii},
  editor       = {Nikhil Bansal and
                  Kirk Pruhs and
                  Clifford Stein},
  title        = {k-means++: the advantages of careful seeding},
  booktitle    = {Proceedings of the Eighteenth Annual {ACM-SIAM} Symposium on Discrete
                  Algorithms, {SODA} 2007, New Orleans, Louisiana, USA, January 7-9,
                  2007},
  pages        = {1027--1035},
  publisher    = {{SIAM}},
  year         = {2007},
  url          = {http://dl.acm.org/citation.cfm?id=1283383.1283494},
  bibsource    = {dblp computer science bibliography, https://dblp.org}
}

@misc{qwen3.5,
    title  = {{Qwen3.5}: Towards Native Multimodal Agents},
    author = {{Qwen Team}},
    month  = {February},
    year   = {2026},
    url    = {https://qwen.ai/blog?id=qwen3.5}
}

@inproceedings{bjerva-etal-2020-subjqa,
    title = {{{SubjQA}: {A} {D}ataset for {S}ubjectivity and {R}eview {C}omprehension}},
    author = "Bjerva, Johannes  and
      Bhutani, Nikita  and
      Golshan, Behzad  and
      Tan, Wang-Chiew  and
      Augenstein, Isabelle",
    editor = "Webber, Bonnie  and
      Cohn, Trevor  and
      He, Yulan  and
      Liu, Yang",
    booktitle = "Proceedings of the 2020 Conference on Empirical Methods in Natural Language Processing (EMNLP)",
    month = nov,
    year = "2020",
    address = "Online",
    publisher = "Association for Computational Linguistics",
    url = "https://aclanthology.org/2020.emnlp-main.442/",
    doi = "10.18653/v1/2020.emnlp-main.442",
    pages = "5480--5494"
}

@inproceedings{sen-etal-2022-counterfactually,
    title = {{Counterfactually Augmented Data and Unintended Bias: The Case of Sexism and Hate Speech Detection}},
    author = "Sen, Indira  and
      Samory, Mattia  and
      Wagner, Claudia  and
      Augenstein, Isabelle",
    editor = "Carpuat, Marine  and
      de Marneffe, Marie-Catherine  and
      Meza Ruiz, Ivan Vladimir",
    booktitle = "Proceedings of the 2022 Conference of the North American Chapter of the Association for Computational Linguistics: Human Language Technologies",
    month = jul,
    year = "2022",
    address = "Seattle, United States",
    publisher = "Association for Computational Linguistics",
    url = "https://aclanthology.org/2022.naacl-main.347/",
    doi = "10.18653/v1/2022.naacl-main.347",
    pages = "4716--4726"
}
\bibliographystyle{acl_natbib}

\appendix

\section{Dataset and Preprocessing}
\label{app:data}

Table~\ref{tab:party_counts} reports per-party raw video and comment counts from the TikTok Research API collection (January 2024 to September 2025), prior to preprocessing. Table~\ref{tab:preprocessing} reports per-step removal counts during preprocessing.

\begin{table}[h]
\centering
\small
\begin{tabular}{lrrr}
\toprule
Party & Videos & Comments & Comm./video \\
\midrule
AfD   & 259 & 131{,}541 & 508 \\
CDU   & 585 & 149{,}399 & 255 \\
SPD   & 583 & 70{,}366  & 121 \\
Linke & 437 & 84{,}701  & 194 \\
Grüne & 360 & 23{,}599  & 66  \\
FDP   & 326 & 8{,}054   & 25  \\
\midrule
\textbf{Total} & \textbf{2{,}550} & \textbf{467{,}660} & avg 183 \\
\bottomrule
\end{tabular}
\caption{Per-party raw video and comment counts from the TikTok Research API, January 2024 to September 2025. Accounts: AfD = @afdfraktionimbundestag, CDU = @insidecdu, SPD = @deinespd, Linke = @die.linke, Grüne = @diegruenen, FDP = @fdpbt.}
\label{tab:party_counts}
\end{table}

\begin{table}[h!]
\centering
\small
\begin{tabular}{lr}
\toprule
\textbf{Step} & \textbf{Removed} \\
\midrule
Duplicate IDs                  & 20,390 \\
Emoji-only                     & 86,883 \\
Empty                          &  1,989 \\
Mention-only                   &  1,273 \\
URL-only                       &    447 \\
Hashtag-only                   &    406 \\
Number-only                    &    783 \\
Laughter-only                  &    270 \\
Duplicate text (within-party)  & 28,543 \\
Duplicate text (cross-party)   &  4,068 \\
Non-German (GlotLID)           & 44,706 \\
\midrule
\textbf{Total removed}  & \textbf{189,758} \\
\midrule
\textbf{Remaining}      & \textbf{277,902} \\
\bottomrule
\end{tabular}
\caption{Per-step preprocessing counts applied to the 467,660 raw comments. GlotLID~\cite{kargaran2023glotlid} is used for language identification to retain only German-language comments.}
\label{tab:preprocessing}
\end{table}

Comments are short: median 5 words in the raw corpus (mean 8.3, 95\textsuperscript{th} percentile 27), rising to a median of 13 words (mean 16.3, 95\textsuperscript{th} percentile 34) in the 25{,}974-comment immigration-relevant subset retained after Llama prefiltering. The 95\textsuperscript{th} percentile in the prefiltered subset motivates the 128-token truncation used in \S\ref{sec:model-training}.

\section{Human Annotation Study}
\label{app:annotation-details}

Table~\ref{tab:annotation-structure} summarises the phase structure introduced in \S\ref{sec:human-annotation}.

\begin{table}[h]
\centering
\small
\begin{tabular}{lccl}
\toprule
\textbf{Phase} & \textbf{Overlap} & \textbf{Items} & \textbf{Use} \\
\midrule
Test           & 6-way & 50    & Eval \\
Pilot          & 6-way & 200   & Eval \\
Main (triple)  & 3-way & 950   & Eval \\
Main (single)  & 1-way & 3{,}800 & Train \\
\midrule
Total eval     &       & 1{,}200 & \\
Total train    &       & 3{,}800 & \\
\bottomrule
\end{tabular}
\caption{Annotation study structure with 6 annotators. Multiply-annotated items form the gold evaluation set; single-annotated items form the AL training pool.}
\label{tab:annotation-structure}
\end{table}

\subsection{Annotated Examples}
\label{app:example-comments}

Table~\ref{tab:example-comments} shows four gold-set comments from the
6-way pilot round with their per-question annotator votes, covering the
four characteristic cases: clear \textsc{anti-immigrant}, immigrant
reference without negative portrayal (Q1 = \textsc{yes}, Q2 =
\textsc{no}), off-topic (Q1 = \textsc{no}), and a genuinely contested
item.

\begin{table*}[t]
\centering
\footnotesize
\setlength{\tabcolsep}{4pt}
\begin{tabular}{p{5.4cm} p{5.4cm} cc l l}
\toprule
\textbf{Comment (German)} & \textbf{English gloss} & \textbf{Q1} & \textbf{Q2} & \textbf{Maj.} & \textbf{GPT-D} \\
 & & \scriptsize y--n & \scriptsize y--n & & \\
\midrule
``Die meisten, die \textquotedbl einwandern\textquotedbl, arbeiten aber nicht.''
 & `But most of those who ``immigrate'' don't work.'
 & 6--0 & 6--0 & \textsc{anti} & \textsc{anti} \\
\addlinespace
``ich bin selber Ausl\"ander 50 Jahre ich wahr nicht Arbeitlos und kein Cent bekommen''
 & `I am a foreigner myself, 50 years; I was not unemployed and didn't receive a cent'
 & 6--0 & 0--6 & \textsc{not} & \textsc{not} \\
\addlinespace
``furchtbar wie eure Politik''
 & `terrible, the way you do politics'
 & 0--6 & 0--6 & \textsc{not} & \textsc{not} \\
\addlinespace
``afd ist gegen kriminelle Ukrainer''
 & `the AfD is against criminal Ukrainians'
 & 6--0 & 4--2 & \textsc{anti} & \textsc{anti} \\
\bottomrule
\end{tabular}
\caption{Example comments from the 6-way-annotated pilot with per-question
vote splits (Q1 = immigrant/immigration reference, Q2 = negative
portrayal; y--n = \textsc{yes}--\textsc{no} votes across the six
annotators), the resulting majority label, and GPT-D's label for the same
item. The four rows illustrate a clear \textsc{anti-immigrant} case, an
immigrant reference passing Q1 but not Q2, an off-topic comment failing
Q1, and a contested item splitting the annotators on valence (all six
affirm Q1; the 4--2 split is entirely on Q2). Comments are reproduced
verbatim, including original spelling.}
\label{tab:example-comments}
\end{table*}

\subsection{Annotator Demographics}
\label{app:annotator-demographics}

The six annotators retained across all study phases (\S\ref{sec:human-annotation}) were recruited through Prolific. They are referred to throughout this appendix as A1, A2, A3, A4, A5, and A8 (Annotator\_6 and Annotator\_7 were excluded after the qualification round, see App.~\ref{app:annotator-qc}; original identifiers are preserved for transparency). Ages ranged from 19 to 60 (mean: 40). Four annotators identified as female and two as male; five identified as White and one as Asian. All six were resident in Germany at the time of the study. Four were native German speakers, and two were highly proficient non-native speakers with first languages Portuguese and Japanese. Employment status comprised three full-time workers, one part-time worker, one unemployed, and one not in paid work. Prolific approval counts ranged from 31 to 1102 (median: 158), indicating a mix of relatively new and experienced platform workers.

Political orientation was not included in the pre-screening criteria. This is a limitation, since annotator political leanings may correlate with how anti-immigrant content is interpreted (\citet{vallejo2025llms} for parallel findings on LLM annotators). Future replications should consider stratified pre-screening on political orientation to characterize this potential source of bias.

\subsection{Holistic-Prompt Pilot and the Two-Question Framework}
\label{app:holistic-pilot}

The two-question decomposition (\S\ref{sec:label_space}) was adopted following an initial holistic-prompt pilot. Ten Prolific annotators received the binary labeling task (``Is this comment anti-immigrant: \textsc{yes}/\textsc{no}?'') with the same written guidelines later used in the main study, at a scale comparable to the qualification round. The pilot yielded a Fleiss $\kappa = 0.175$, indicating very low agreement. Manual review suggested that annotators inconsistently applied the boundary between policy critique and group-directed hostility; the single-question format did not surface this distinction during annotation. We therefore designed the main study with the label decomposed into two sequential yes/no questions (Q1: presence of immigrant reference; Q2: negative portrayal), scaffolding the decision and forcing engagement with each criterion. The main study used a freshly recruited cohort of 8 annotators, of whom 6 were retained after qualification (App.~\ref{app:annotator-qc}). Pilot costs are not included in the cost analysis of \S\ref{sec:cost}; they were comparable to the qualification round.

\subsection{Attention Checks}
\label{app:annotator-qc}

Instruction manipulation checks (IMCs) were interleaved throughout all phases at densities of approximately 10 per 100 items (test), 5 per 100 (pilot), and 3 per 100 (main). Each check instructs the annotator to select specific Q1 and Q2 responses, testing whether comment text is being read. A minimum of two checks was enforced per annotator per phase.

In the qualification test, Annotator\_7 failed to meet the $\geq$80\% accuracy threshold on attention checks (2/3, 67\%) and was excluded. Annotator\_6 passed attention checks but showed extremely low inter-annotator agreement on the valence question (average pairwise $\kappa = 0.13$ on Q2) and was also excluded. The remaining six annotators achieved 83--100\% accuracy on attention checks across all phases, with all six scoring 100\% in the main round.

\subsection{Item Assignment}
\label{app:item-assignment}

In the main round, 950 items received triple annotation for the gold evaluation set. Items were distributed via round-robin assignment over all $\binom{6}{3} = 20$ possible 3-annotator combinations to ensure balanced workload and annotator-pair coverage. The remaining 3,800 items received single annotation and were distributed evenly across annotators.

\subsection{Inter-Annotator Agreement}
\label{app:iaa}

Tables~\ref{tab:iaa-pilot} and~\ref{tab:iaa-main} report agreement statistics for the pilot and main phases. Tables~\ref{tab:pairwise-pilot} and~\ref{tab:pairwise-main} provide full pairwise Cohen's $\kappa$ matrices for the derived binary label.

\begin{table}[h!]
\centering
\small
\begin{tabular}{lccc}
\toprule
\textbf{Metric} & \textbf{Q1} & \textbf{Q2} & \textbf{Label} \\
\midrule
Krippendorff's $\alpha$ & 0.574 & 0.493 & 0.494 \\
Cohen's $\kappa$ (avg)  & 0.580 & 0.488 & 0.490 \\
Cohen's $\kappa$ range  & .356--.807 & .310--.758 & .320--.741 \\
\bottomrule
\end{tabular}
\caption{Inter-annotator agreement on the pilot phase (200 items, 6-way overlap).}
\label{tab:iaa-pilot}
\end{table}

\begin{table}[h!]
\centering
\small
\begin{tabular}{lccc}
\toprule
\textbf{Metric} & \textbf{Q1} & \textbf{Q2} & \textbf{Label} \\
\midrule
Krippendorff's $\alpha$ & 0.591 & 0.432 & 0.433 \\
Cohen's $\kappa$ (avg)  & 0.599 & 0.437 & 0.439 \\
Cohen's $\kappa$ range  & .384--.704 & .221--.588 & .221--.588 \\
\bottomrule
\end{tabular}
\caption{Inter-annotator agreement on the main phase (950 triple-annotated items, 3-way overlap). Krippendorff's $\alpha$ is used as Fleiss' $\kappa$ is not applicable for variable annotator subsets.}
\label{tab:iaa-main}
\end{table}

\begin{table}[h!]
\centering
\small
\begin{tabular}{lcccccc}
\toprule
 & \textbf{A1} & \textbf{A2} & \textbf{A3} & \textbf{A4} & \textbf{A5} & \textbf{A8} \\
\midrule
\textbf{A1} & ---   & .470 & .515 & .518 & .386 & .320 \\
\textbf{A2} & .470 & ---   & .738 & .741 & .545 & .408 \\
\textbf{A3} & .515 & .738 & ---   & .615 & .551 & .395 \\
\textbf{A4} & .518 & .741 & .615 & ---   & .425 & .351 \\
\textbf{A5} & .386 & .545 & .551 & .425 & ---   & .363 \\
\textbf{A8} & .320 & .408 & .395 & .351 & .363 & ---   \\
\bottomrule
\end{tabular}
\caption{Pairwise Cohen's $\kappa$ for the derived binary label on the pilot phase (200 items, 6-way overlap).}
\label{tab:pairwise-pilot}
\end{table}

\begin{table}[h!]
\centering
\small
\begin{tabular}{lcccccc}
\toprule
 & \textbf{A1} & \textbf{A2} & \textbf{A3} & \textbf{A4} & \textbf{A5} & \textbf{A8} \\
\midrule
\textbf{A1} & ---   & .535 & .419 & .272 & .358 & .357 \\
\textbf{A2} & .535 & ---   & .588 & .471 & .560 & .480 \\
\textbf{A3} & .419 & .588 & ---   & .536 & .487 & .521 \\
\textbf{A4} & .272 & .471 & .536 & ---   & .416 & .359 \\
\textbf{A5} & .358 & .560 & .487 & .416 & ---   & .221 \\
\textbf{A8} & .357 & .480 & .521 & .359 & .221 & ---   \\
\bottomrule
\end{tabular}
\caption{Pairwise Cohen's $\kappa$ for the derived binary label on the main phase (950 triple-annotated items, 3-way overlap). Each pair is computed over items where both annotators provided labels.}
\label{tab:pairwise-main}
\end{table}

\subsection{Per-Annotator Statistics}
\label{app:per-annotator}

Table~\ref{tab:annotator-stats} reports per-annotator average $\kappa$ (with all other annotators) and \textsc{anti-immigrant} label rates across the pilot and main phases. Annotator\_1 shows a consistently higher positive label rate (34--37\%) than other annotators (16--29\%), reflecting a more inclusive interpretation of what constitutes anti-immigrant content.

\begin{table}[h!]
\centering
\small
\begin{tabular}{lcccc}
\toprule
 & \multicolumn{2}{c}{\textbf{Pilot}} 
 & \multicolumn{2}{c}{\textbf{Main}} \\
\cmidrule(lr){2-3} \cmidrule(lr){4-5}
\textbf{Annotator} & $\bar{\kappa}$ & ANTI\% 
                    & $\bar{\kappa}$ & ANTI\% \\
\midrule
A1 & .442 & 34.0 & .388 & 37.0 \\
A2 & .581 & 26.5 & .527 & 28.7 \\
A3 & .563 & 29.0 & .510 & 21.6 \\
A4 & .530 & 30.0 & .411 & 16.5 \\
A5 & .454 & 18.5 & .408 & 16.3 \\
A8 & .367 & 15.5 & .388 & 17.0 \\
\bottomrule
\end{tabular}
\caption{Per-annotator average Cohen's $\kappa$ (on the derived binary label, averaged over all pairwise comparisons) and \textsc{anti-immigrant} label rate, by study phase.}
\label{tab:annotator-stats}
\end{table}

\subsection{Per-Annotator and Per-Topic Rates}
\label{app:annotator-topics}

Table~\ref{tab:annotator-topic-rates} breaks the per-annotator
\textsc{anti-immigrant} rates of Table~\ref{tab:annotator-stats} down by
discourse topic (\S\ref{sec:topics}). A1's higher overall rate is not a
uniform offset: it concentrates in economic-competition and
immigration-policy framing (+40pp over the other annotators on Welfare \&
Economic Competition, +28pp on Ukraine \& Ukrainians), while on clear-cut
mobilization or nationality-specific comments A1 tracks the other
annotators within a few points. Human strictness differences are thus
topic-specific rather than global, the human counterpart to the
model-specific topical biases in \S\ref{sec:topics}.

Table~\ref{tab:topic-agreement} reports crowd agreement per topic on the
multiply-annotated gold items. Agreement is strongly topic-dependent:
pairwise agreement ranges from 56--69\% on contested clusters (Criminal
Violence, Border Control, Refugees \& Asylum, Islam \& Islamization) to
91--95\% on clear-cut ones (Rejection Rhetoric, Pro-AfD Mobilization),
though the most contested clusters have small gold-set $n$ (10--20), so
their point estimates are noisy. Disagreement sits where the judgment is genuinely hard, the pattern expected from a subjective task rather than from unreliable annotators.

\begin{table*}[t]
\centering
\footnotesize
\setlength{\tabcolsep}{4pt}
\begin{tabular}{ll r rrrrrr}
\toprule
\textbf{ID} & \textbf{Topic} & $n$ & \textbf{A1} & \textbf{A2} & \textbf{A3} & \textbf{A4} & \textbf{A5} & \textbf{A8} \\
\midrule
C0  & Germany in Decline \& National Identity  & 1{,}069 & 33.7 & 25.9 & 24.2 & 18.2 & 16.5 & 17.2 \\
C1  & Pro-AfD Mobilization                     & 438  & 6.1  & 6.5  & 7.9  & 2.7  & 2.4  & 5.6 \\
C2  & Democracy \& Governance Critique         & 356  & 16.3 & 15.2 & 4.2  & 11.7 & 10.5 & 9.7 \\
C3  & Welfare State \& Economic Competition    & 293  & 65.2 & 32.0 & 19.2 & 33.0 & 17.9 & 20.9 \\
C6  & Migration Policy \& Terminology          & 221  & 70.7 & 51.6 & 52.6 & 40.0 & 36.7 & 31.4 \\
C5  & Merkel/CDU \& 2015 Legacy                & 218  & 25.7 & 34.4 & 32.8 & 13.3 & 22.2 & 9.5 \\
C4  & Directed Personal Hostility              & 213  & 26.1 & 9.4  & 16.7 & 7.3  & 7.8  & 13.2 \\
C7  & Ukraine \& Ukrainians                    & 168  & 83.0 & 68.1 & 45.7 & 55.6 & 53.5 & 51.2 \\
C8  & Afghanistan, Syria \& Turkey             & 81   & 50.0 & 50.0 & 51.9 & 47.8 & 55.6 & 26.3 \\
C12 & Refugees \& Asylum Policy                & 76   & 57.9 & 70.4 & 38.9 & 65.0 & 47.8 & 45.8 \\
C9  & Left-Wing Party Critique (Gr\"une/Linke) & 70   & 31.6 & 31.8 & 50.0 & 10.0 & 16.7 & 11.1 \\
C10 & Criminal Violence \& Public Safety       & 66   & 62.5 & 52.9 & 45.5 & 25.0 & 16.7 & 50.0 \\
\midrule
 & \textit{All items} & 5{,}000 & 36.9 & 28.4 & 22.8 & 19.0 & 17.7 & 17.5 \\
\bottomrule
\end{tabular}
\caption{Per-annotator \textsc{anti-immigrant} rate (\%) by discourse topic
(\S\ref{sec:topics}), over all items each annotator labeled (all phases
pooled); $n$ = distinct annotated comments in the topic. Shown are the 12
topics where all six annotators have $n \geq 15$ items; the full 22-topic
breakdown is in the released data. A1 is the strictest annotator overall
(36.9\%), but the gap is topic-specific rather than uniform.}
\label{tab:annotator-topic-rates}
\end{table*}

\begin{table}[t]
\centering
\footnotesize
\setlength{\tabcolsep}{2pt}
\begin{tabular}{@{}ll r rr r@{}}
\toprule
\textbf{ID} & \textbf{Topic} & $n$ & \textbf{Pair.} & \textbf{Unan.} & $\alpha$ \\
 & & & (\%) & (\%) & \\
\midrule
C18 & Israel/Palestine \& M.\ East & 3$^\dagger$ & 42.9 & 0.0  & $-$.13 \\
C19 & War \& Militarism            & 3$^\dagger$ & 55.6 & 33.3 & .20 \\
C10 & Criminal Violence            & 16  & 56.0  & 31.2 & .08 \\
C13 & Border Control               & 13  & 60.8  & 23.1 & .06 \\
C7  & Ukraine \& Ukrainians        & 41  & 64.7  & 43.9 & .29 \\
C12 & Refugees \& Asylum           & 20  & 65.0  & 40.0 & .25 \\
C21 & Brandmauer Metaphors         & 2$^\dagger$ & 66.7 & 50.0 & .00 \\
C14 & Islam \& Islamization        & 10  & 68.5  & 30.0 & .15 \\
C16 & Female Politicians           & 7$^\dagger$ & 71.1 & 57.1 & .10 \\
C3  & Welfare \& Econ.\ Comp.      & 64  & 72.8  & 48.4 & .31 \\
C6  & Migration Policy             & 48  & 72.9  & 56.2 & .51 \\
C5  & Merkel/CDU \& 2015           & 65  & 76.3  & 69.2 & .50 \\
C0  & Germany in Decline           & 264 & 79.4  & 67.0 & .43 \\
C15 & EU Politics                  & 9$^\dagger$ & 79.5 & 55.6 & .19 \\
C9  & Left-Wing Critique           & 18  & 83.3  & 66.7 & .56 \\
C2  & Democracy Critique           & 85  & 84.2  & 78.8 & .37 \\
C4  & Directed Hostility           & 46  & 86.7  & 76.1 & .37 \\
C8  & Afghanistan/Syria/Turkey     & 22  & 87.3  & 77.3 & .73 \\
C20 & Rejection Rhetoric           & 12  & 91.7  & 83.3 & $-$.03 \\
C11 & Racism \& Fascism            & 8$^\dagger$ & 94.4 & 87.5 & .00 \\
C1  & Pro-AfD Mobilization         & 108 & 95.1  & 91.7 & .50 \\
C17 & Poll Numbers                 & 7$^\dagger$ & 100.0 & 100.0 & -- \\
\midrule
 & \textit{22 topics} & 871     & 79.5 & 66.8 & .46 \\
 & \textit{All gold}  & 1{,}200 & 80.0 & 67.2 & .46 \\
\bottomrule
\end{tabular}
\caption{Per-topic crowd agreement on the multiply-annotated gold items.
Pair.\ = mean pairwise \% agreement on the final binary label; Unan.\ = \%
of items with a unanimous label; $\alpha$ = Krippendorff's nominal alpha
(unreliable for small $n$). Sorted by pairwise agreement, ascending.
$^\dagger$$n < 10$: agreement estimate unreliable. Topic names abbreviated;
full names in Table~\ref{tab:topics-full}. The 22 named topics cover 871
of the 1{,}200 gold items; the remainder fall in the residual noise
cluster (\textit{All gold} row).}
\label{tab:topic-agreement}
\end{table}

\subsection{Class-Conditional Agreement}
\label{app:class-agreement}

Agreement is higher for the majority class across both phases. On the pilot (200 items, majority vote), items labeled \textsc{not anti-immigrant} (n=156) show 90.2\% average agreement with 63\% unanimity, while items labeled \textsc{anti-immigrant} (n=44) show 81.4\% average agreement with 30\% unanimity. On the main round (950 triple-annotated items), agreement is high overall: 96.3\% for \textsc{anti-immigrant} items (89\% unanimous) and 98.5\% for \textsc{not anti-immigrant} items (96\% unanimous), reflecting that the main round's 3-way overlap yields higher raw agreement than the pilot's 6-way overlap.

\subsection{Expert Validation of the Gold Standard}
\label{app:expert-validation}

An independent domain expert re-annotated a stratified 10\% sample of the gold
set (120 of 1{,}200 items) using the same two-question interface
(\S\ref{sec:label_space}), blind to the crowd labels. The sample contains 38
\textsc{anti-immigrant} items (31.7\%) and 82 \textsc{not anti-immigrant}
(68.3\%); 81 items had been labeled unanimously by the crowd annotators and 39
were split. All 120 responses parsed to valid labels.

\paragraph{Agreement with the crowd majority.}
Table~\ref{tab:expert-validation} reports agreement between the expert and the
crowd majority label. The expert agrees on 88.3\% of items (Cohen's
$\kappa = 0.73$), and the mean overlap with the full annotator pool, that is
the fraction of individual crowd annotators sharing the expert's label, is
0.846. Agreement is higher on Q1 (reference, 91.7\%, $\kappa = 0.83$) than on
Q2 (valence, 88.3\%, $\kappa = 0.73$), mirroring the crowd's own pattern
(\S\ref{sec:human-annotation}): disagreement concentrates on valence rather
than topic identification.

The expert reproduces the crowd consensus almost exactly where that consensus
exists: 96.3\% agreement on unanimously labeled items against 71.8\% on split
items. Of the 14 disagreements, 11 fall on items the crowd had already split,
leaving 3 genuine interpretive differences against a unanimous crowd label. The
expert applies \textsc{anti-immigrant} marginally less often than the crowd
majority (36 vs.\ 38 of 120 items). Two properties make this a conservative
estimate: the sample is enriched for the positive class relative to the full
gold set (31.7\% vs.\ 21.7\% \textsc{anti}), and positive items are the
contested ones, so agreement on a uniform random sample would be expected to be
somewhat higher.

{\setlength{\intextsep}{6pt plus 2pt minus 2pt}
\begin{table}[h!]
\centering
\footnotesize
\begin{tabular}{lrrr}
\toprule
& $n$ & \textbf{Agreement} & \textbf{Cohen's} $\kappa$ \\
\midrule
Q1 (reference)  & 120 & 91.7\% & 0.83 \\
Q2 (valence)    & 120 & 88.3\% & 0.73 \\
Final label     & 120 & 88.3\% & 0.73 \\
\midrule
Crowd unanimous & 81 & 96.3\% & --- \\
Crowd split     & 39 & 71.8\% & --- \\
\bottomrule
\end{tabular}
\caption{Expert validation on a stratified 10\% sample of the gold set. Agreement is between the independent expert and the crowd majority label.}
\label{tab:expert-validation}
\end{table}

\paragraph{Agreement with the LLM annotators.}
The same 120 items allow a comparison against the LLM labels. The expert agrees
with Qwen-DT on 75.8\% of items (91/120), against 88.3\% with the crowd
majority. Of the 29 disagreements, 26 are cases where Qwen assigns
\textsc{anti-immigrant} and the expert does not; only 3 run the other way. This
independently corroborates the over-flagging pattern reported in
\S\ref{sec:rq4-errors} and \S\ref{sec:topics}, on a sample and an annotator
that play no role in the classifier experiments.

The expert's free-text notes (recorded on 45 of 120 items) indicate two
recurring mechanisms, both consistent with the topical analysis. The first is
implicit-reference over-attribution: on terse or ambiguous comments the expert
marks Q1 = \textsc{no}, noting that the reference is unclear or that the
connection cannot be established without video context, where the LLM infers an
immigrant referent that is not present in the text. The second is policy
criticism read as hostility: a cluster of items marked Q1 = \textsc{yes},
Q2 = \textsc{no} with notes such as ``restrictive position, but no clear
derogation'', where the expert reserves \textsc{anti-immigrant} for derogation
of migrants as a group and separates it from a restrictive stance on
immigration policy. This is the same distinction the Q1/Q2 decomposition is
designed to enforce (\S\ref{sec:why-decomposed}).

\section{LLM Annotation Pipeline}
\label{app:prompts}

\subsection{Stage 1: Topic Prefiltering Prompt}
\label{app:llama-filter}

\begin{figure}[h!]
\begin{promptbox}{Llama-3.3-70B-Instruct Topic Prefiltering}
\footnotesize\ttfamily
Does this German TikTok comment discuss immigration, migrants, refugees, asylum seekers, foreigners, or related policies?\\[4pt]
Output JSON: \{"label": "YES" or "NO"\}\\[4pt]
Comment: \{comment\}
\end{promptbox}
\caption{Prefiltering prompt applied to all 277,902 comments (temperature = 0, JSON output).}
\label{prompt:llama-filter}
\end{figure}

\vspace{4pt}

\subsection{Stage 2: Holistic Annotation Prompt (Interface Ablation)}
\label{app:gpt-prompt-holistic}

\begin{figure}[h!]
\begin{promptbox}[breakable=false]{GPT-5.2 Anti-Immigrant Annotation (Holistic)}
\small\ttfamily
You are an expert in sentiment analysis and political discourse, specializing in German-language text. Your task is to classify TikTok comments posted under videos from German political parties.\\[4pt]
Classify the comment as either "ANTI-IMMIGRANT" or "NOT ANTI-IMMIGRANT".\\[4pt]
\textbf{Definitions:}\\
- ANTI-IMMIGRANT: Expresses negative sentiments, hostility, fear, or opposition towards immigrants, refugees, asylum seekers, or immigration policies. Includes xenophobic language, stereotypes, calls for stricter borders, deportation advocacy, or blaming immigrants for social/economic issues. Subtle implications count if they clearly lean negative.\\
- NOT ANTI-IMMIGRANT: Neutral, positive towards immigrants/immigration, or unrelated to immigration altogether.\\[4pt]
\textbf{Guidelines:}\\
- Analyze the comment in its original German.\\
- Base classification solely on the comment's content, not party context.\\
- If ambiguous or sarcastic, default to "NOT ANTI-IMMIGRANT".\\
- Base decisions only on visible text; no external inference.\\[4pt]
Output JSON: \{"label": "ANTI-IMMIGRANT" or "NOT ANTI-IMMIGRANT"\}\\[4pt]
Comment: \{comment\}
\end{promptbox}
\captionof{figure}{Holistic classification prompt applied to 25,974 prefiltered comments via GPT-5.2 Batch API (temperature = 0, structured JSON output). Used for the GPT-H interface ablation.}
\label{fig:prompt-holistic}
\end{figure}

\subsection{Stage 2: Decomposed Annotation Prompt}
\label{app:gpt-prompt-decomposed}

This prompt is used by GPT-D, Qwen-DT, and Qwen-DNT. The system message specifies the JSON schema; the user message provides the annotation guidelines (translated from the Argilla interface used by human annotators in \S\ref{sec:human-annotation}) followed by the comment to annotate. Serving details for each model (endpoint, sampling parameters, reasoning-mode flag) are in \S\ref{app:infra}.

\begin{figure*}[h!]
\begin{promptbox}[breakable=false]{Decomposed Annotation}
\small\ttfamily\raggedright
\textbf{System:}\\[2pt]
You are an annotator answering two questions about a German TikTok comment posted under a video of a German political party.\\[4pt]
You see only the comment text, with no video context.\\[4pt]
Return your answer as a single JSON object with these fields:\\
- "q1":           "YES" or "NO"\\
- "q1\_reasoning": one short sentence in German\\
- "q2":           "YES" or "NO"\\
- "q2\_reasoning": one short sentence in German\\[8pt]
\textbf{User:}\\[2pt]
Annotationsrichtlinie:\\[2pt]
\# TikTok-Kommentare -- Annotationsrichtlinie\\[2pt]
\#\# 1. Aufgabenübersicht\\
Sie werden deutsche TikTok-Kommentare, die unter Videos deutscher politischer Parteien gepostet wurden, anhand von **zwei Fragen** bewerten.\\[2pt]
**Wichtig:**\\
- Sie sehen nur den Kommentartext, ohne Video-Kontext\\[2pt]
\#\# 2. Die zwei Fragen\\[2pt]
\#\#\# Q1: Bezieht sich der Kommentar auf Einwanderer, Flüchtlinge, Asylsuchende, Ausländer oder Einwanderung/Asylpolitik?\\
- **YES:** Der Kommentar bezieht sich auf Einwanderer, Flüchtlinge oder Asylsuchende -- sei es direkt, durch Slang/Codewörter oder durch typische Begriffe aus der Migrationsdebatte\\
- **NO:** Kein Bezug zu Einwanderung oder Einwanderergruppen\\[2pt]
\#\#\# Q2: Werden Einwanderer/Flüchtlinge/Ausländer in diesem Kommentar abwertend dargestellt -- durch Feindseligkeit, Angst, Ablehnung, fremdenfeindliche Sprache, Stereotype, Forderungen nach strengeren Grenzen, Abschiebungsrhetorik oder Schuldzuweisungen für soziale/wirtschaftliche Probleme?\\
- **YES:** Feindseligkeit, Angst, Ablehnung, Stereotype, Abschiebungsrhetorik oder Schuldzuweisungen für soziale/wirtschaftliche Probleme\\
- **NO:** Der Kommentar ist sachlich, neutral, positiv gegenüber Einwanderern/Einwanderung, oder hat keinen Bezug zu Einwanderung.\\[4pt]
Kommentar:\\
"\{comment\}"\\[4pt]
Antworten Sie mit einem JSON-Objekt im folgenden Format:\\
\{\\
\ \ "q1": "YES" or "NO",\\
\ \ "q1\_reasoning": "...",\\
\ \ "q2": "YES" or "NO",\\
\ \ "q2\_reasoning": "..."\\
\}
\end{promptbox}
\caption{Decomposed annotation prompt, identical across the three decomposed LLM variants. The user-message guidelines are verbatim from the Argilla interface used by human annotators (\S\ref{sec:human-annotation}).}
\label{prompt:decomposed}
\end{figure*}

\paragraph{Label derivation.}
The binary label is derived deterministically from the JSON response, outside the model: $\textsc{anti-immigrant} \Leftrightarrow q_1 = \text{YES} \wedge q_2 = \text{YES}$; otherwise \textsc{not anti-immigrant}. Items for which the model exceeds the output token budget before emitting valid JSON are excluded from downstream classifier training; evaluation on the human-labeled test set is unaffected. Per-model counts are reported in \S\ref{app:infra}.

\subsection{Infrastructure Details}
\label{app:infra}

\paragraph{Llama-3.3-70B-Instruct (Stage 1).}
The model was run locally on four NVIDIA A100 80GB GPUs using vLLM~\cite{kwon2025vllm} with tensor parallelism across all four devices, bfloat16 precision, and a maximum sequence length of 512 tokens. Filtering all 277{,}902 comments completed in 0.55 wall-clock hours (2.20 GPU-hours total) at a throughput of approximately 140 comments/second. At the Lambda Labs A100 on-demand rate of \$2.79/GPU-hour, the cloud-equivalent cost is $\approx$\$6.14 USD. Researchers without GPU access can replicate Stage~1 via the DeepInfra API (\texttt{meta-llama/Llama-3.3-70B-Instruct}) at approximately \$5.28 for the same job (\$0.10/1M input tokens, \$0.40/1M output tokens; 41.7M input and 2.8M output tokens estimated).

\paragraph{GPT-5.2 (Stage 2).}
We submitted all 25{,}974 prefiltered comments through the OpenAI Batch API. Each request comprised the decomposed annotation prompt (\S\ref{app:gpt-prompt-decomposed}) plus the comment text, yielding approximately 8.1M input tokens and 2.1M output tokens across the run. The Batch API charges 50\% of standard rates and processes within a 24-hour window; the total invoice was \textbf{\$28 USD} ($\approx$\$1.08 per 1{,}000 comments). All 25{,}974 comments received valid JSON responses with no content-policy refusals.

\paragraph{Qwen3.5-122B-A10B (Stage 2).}
We additionally annotated the same 25{,}974 prefiltered comments with Qwen3.5-122B-A10B, an open-weight 122B-parameter mixture-of-experts model with $\sim$10B active parameters per token, distributed under Apache-2.0. The model was served locally on four NVIDIA A100 80GB GPUs using vLLM~\cite{kwon2025vllm} with tensor parallelism, bfloat16 precision, and the same decomposed prompt as GPT-5.2 (\S\ref{app:gpt-prompt-decomposed}). Two configurations were run: the main \textsc{Qwen-DT} variant with reasoning enabled (\texttt{enable\_thinking=True}, max output 16{,}384 tokens), which exceeded the output token budget on 136 of 25{,}974 comments (0.5\%); and the ablation \textsc{Qwen-DNT} variant with reasoning disabled (max output 4{,}096 tokens), which exceeded the budget on a comparable share of comments ($\sim$0.5\%).
The reasoning-enabled run completed in 26 GPU-hours; the no-thinking ablation in 2.5 GPU-hours. At the Lambda Labs A100 on-demand rate of \$2.79/GPU-hour, the cloud-equivalent costs are $\approx$\$73 (Qwen-DT) and $\approx$\$7 (Qwen-DNT); on owned hardware, the marginal cost is electricity only.

\paragraph{Human Annotation Cost Breakdown.}
Annotators were compensated at \pounds9/hour across three phases. Phase-level costs: \pounds18 (test, 20\,min per annotator $\times$ 6) + \pounds54 (pilot, 60\,min $\times$ 6) + \pounds162 (main, 3\,hours $\times$ 6) = \pounds234 base compensation. With Prolific's 33.3\% academic platform fee (\pounds77.92), the grand total was \pounds312 (\$416 USD). Per-condition costs are computed proportionally: \textsc{Full-Human} uses 3{,}800 of 5{,}000 labels ($3{,}800 / 5{,}000 \times \$416 = \$316$); \textsc{AL-Human} uses a maximum of 530 ($530 / 5{,}000 \times \$416 = \$44$).

\subsection{LLM Label Distributions}
\label{app:label-distributions}

Table~\ref{tab:label-distributions} reports the per-party distribution of ANTI-IMMIGRANT labels under each of the four LLM annotators. Two observations are worth flagging. First, CDU/CSU comments receive the highest ANTI-IMMIGRANT rate among the four major parties under every LLM variant (24.9\% to 37.6\%), exceeding AfD (21.6\% to 29.7\%) in each case. A plausible explanation is that CDU/CSU video topics often center on migration policy debates, drawing migration-themed comment threads from both supportive and critical audiences. Table~\ref{tab:per-party-human-llm} shows that this ordering is specific to the LLM labels: on the 5{,}000 crowd-annotated comments the crowd rate is roughly flat across parties (19--26\%) and does not place CDU/CSU above AfD. Second, the ANTI-IMMIGRANT rate is 7--10 percentage points lower under the decomposed interface than under the holistic one across all six party segments, with the relative ordering of parties preserved: the Q1/Q2 decomposition functions as a roughly uniform correction across party segments, though not across discourse themes (\S\ref{sec:topics}).

\begin{table*}[t]
\centering
\small
\setlength{\tabcolsep}{5pt}
\begin{tabular}{lr rr rr rr rr}
\toprule
\textbf{Party} & \textbf{$n_{\textsc{party}}$} & \multicolumn{2}{c}{\textbf{GPT-D}} & \multicolumn{2}{c}{\textbf{GPT-H}} & \multicolumn{2}{c}{\textbf{Qwen-DT}} & \multicolumn{2}{c}{\textbf{Qwen-DNT}} \\
\cmidrule(lr){3-4} \cmidrule(lr){5-6} \cmidrule(lr){7-8} \cmidrule(lr){9-10}
 &  & \textbf{$n$} & \textbf{\%} & \textbf{$n$} & \textbf{\%} & \textbf{$n$} & \textbf{\%} & \textbf{$n$} & \textbf{\%} \\
\midrule
AfD & 9,379 & 2,022 & 21.6 & 2,691 & 28.7 & 2,786 & 29.7 & 2,780 & 29.6 \\
CDU/CSU & 7,316 & 1,821 & 24.9 & 2,508 & 34.3 & 2,557 & 35.0 & 2,753 & 37.6 \\
SPD & 4,324 & 981 & 22.7 & 1,430 & 33.1 & 1,384 & 32.0 & 1,426 & 33.0 \\
Linke & 3,311 & 570 & 17.2 & 831 & 25.1 & 841 & 25.4 & 928 & 28.0 \\
Grüne & 1,244 & 144 & 11.6 & 246 & 19.8 & 260 & 20.9 & 274 & 22.0 \\
FDP & 400 & 113 & 28.2 & 141 & 35.2 & 133 & 33.2 & 154 & 38.5 \\
\midrule
\textbf{Total} & 25,974 & 5,651 & 21.8 & 7,847 & 30.2 & 7,961 & 30.6 & 8,315 & 32.0 \\
\bottomrule
\end{tabular}
\caption{Per-party distribution of ANTI-IMMIGRANT labels across the four LLM annotation variants on the 25{,}974-comment Llama-prefiltered pool (complementing Table~1 in the main paper). $n_{\textsc{party}}$ is the number of comments from each party's TikTok account; $n$ and \% report the count and share of those comments labeled ANTI-IMMIGRANT under each LLM. Counts and percentages are computed over the full 25{,}974-comment pool; Qwen-DT excluded 136 items (0.5\%) where the output exceeded the token budget (\S\ref{app:infra}), while Qwen-DNT's over-budget items were successfully re-parsed, so all 25{,}974 received labels. GPT-D and GPT-H denote the decomposed and holistic GPT-5.2 prompts; Qwen-DT and Qwen-DNT denote Qwen3.5-122B-A10B with and without enabled thinking.}
\label{tab:label-distributions}
\end{table*}

\begin{table*}[t]
\centering
\footnotesize
\setlength{\tabcolsep}{4pt}
\begin{tabular}{lrrrrrrrrrr}
\toprule
& & \textbf{Human} & \multicolumn{2}{c}{\textbf{GPT-D}} & \multicolumn{2}{c}{\textbf{GPT-H}} & \multicolumn{2}{c}{\textbf{Qwen-DT}} & \multicolumn{2}{c}{\textbf{Qwen-DNT}} \\
\cmidrule(lr){4-5}\cmidrule(lr){6-7}\cmidrule(lr){8-9}\cmidrule(lr){10-11}
\textbf{Party} & $n$ & \textsc{anti}\% & \textsc{anti}\% & agr.\% & \textsc{anti}\% & agr.\% & \textsc{anti}\% & agr.\% & \textsc{anti}\% & agr.\% \\
\midrule
AfD      & 1{,}792 & 23.3 & 28.1 & 83.9 & 38.3 & 79.0 & 36.5 & 80.1 & 35.6 & 79.6 \\
CDU/CSU  & 1{,}435 & 24.3 & 31.7 & 80.1 & 44.5 & 72.5 & 41.7 & 74.6 & 45.4 & 71.5 \\
SPD      &   844 & 21.6 & 29.0 & 82.6 & 43.2 & 72.9 & 38.2 & 74.4 & 39.7 & 73.3 \\
Linke    &   622 & 20.7 & 22.8 & 83.1 & 34.1 & 76.7 & 31.4 & 78.6 & 34.2 & 77.2 \\
Grüne    &   229 & 19.2 & 15.7 & 88.6 & 27.5 & 85.6 & 28.9 & 85.5 & 30.6 & 82.5 \\
FDP      &    78 & 25.6 & 32.1 & 80.8 & 44.9 & 73.1 & 37.2 & 75.6 & 46.2 & 74.4 \\
\midrule
Total    & 5{,}000 & 22.8 & 28.1 & 82.6 & 40.0 & 76.0 & 37.3 & 77.6 & 38.9 & 76.0 \\
\bottomrule
\end{tabular}
\caption{Per-party \textsc{anti-immigrant} rates for the crowd majority and the four LLM variants on the 5{,}000 human-annotated comments, with per-variant label agreement against the crowd majority. Rates are computed over the annotated subset and are therefore higher than the full-pool rates in Table~\ref{tab:label-distributions}, which covers all 25{,}974 comments. The crowd rate is roughly flat across parties (19--26\%) and is not concentrated on AfD posts; every LLM variant assigns a higher rate than the crowd for every party, and GPT-D is closest to the crowd on all six. FDP ($n=78$) is small and its rates are correspondingly noisy.}
\label{tab:per-party-human-llm}
\end{table*}

\section{Models and Acquisition Functions}
\label{app:models-and-acquisitions}

\subsection{Encoder Model Details}
\label{app:encoder-models}

Table~\ref{tab:encoder-models} summarizes the four pre-trained encoders evaluated in our supervised pipeline. All four are encoder-only BERT-style transformers fine-tuned with a randomly initialized binary classification head; parameter counts cover the full pre-trained encoder and exclude the two-class output head ($2 \times 768 + 2 = 1{,}538$ parameters for the base-sized models). Three of the four models are German-specific, with xlm-r-base serving as a multilingual baseline. All four encoders are fine-tuned end-to-end with identical training hyperparameters (\S\ref{sec:model-training}): all encoder parameters are updated jointly with the binary classification head, with no frozen layers.

\begin{table}[t!]
\centering
\small
\begin{tabular}{lcc}
\toprule
\textbf{Model} & \textbf{Params} & \textbf{Pretraining language(s)} \\
\midrule
german-bert   & 110M & German \\
ModernGBERT   & 134M & German \\
gbert-base    & 110M & German \\
xlm-r-base    & 278M & 100 languages incl.\ German \\
\bottomrule
\end{tabular}
\caption{Pre-trained encoder models. All four are fine-tuned end-to-end with identical training hyperparameters (\S\ref{sec:model-training}). Parameter counts exclude the binary classification head. Model sources: \href{https://hf.co/dbmdz/bert-base-german-uncased}{german-bert}, \href{https://hf.co/LSX-UniWue/ModernGBERT_134M}{ModernGBERT}, \href{https://hf.co/deepset/gbert-base}{gbert-base}, \href{https://hf.co/FacebookAI/xlm-roberta-base}{xlm-r-base}.}
\label{tab:encoder-models}
\end{table}

All downstream classifier training was run on NVIDIA RTX A6000 48GB GPUs (one GPU per training run; up to 8 A6000s dispatched in parallel). The complete experimental sweep comprises 1{,}360 training runs across 10 random seeds and 4 encoders for each of: \textsc{Full-Human}, \textsc{Full-LLM} (3.8K) under 4 LLM variants, \textsc{Full-LLM-25K} under 4 LLM variants, \textsc{Random-Human}, \textsc{Random-LLM} under 4 LLM variants, \textsc{AL-Human} under 4 acquisition strategies, and \textsc{AL-LLM} under 4 acquisitions $\times$ 4 LLM variants. Total downstream compute is \textbf{244.6 A6000 GPU-hours}, with \textsc{AL-LLM} (109.2 GPU-hours; 45\%) and \textsc{Full-LLM-25K} (82.9 GPU-hours; 34\%) accounting for the bulk of the footprint. Per-backbone totals are 53.8 (\texttt{german\_bert}), 74.1 (\texttt{ModernGBERT}), 57.0 (\texttt{gbert-base}), and 59.7 (\texttt{xlm-r-base}) GPU-hours. Stage 1 (Llama prefiltering, 4$\times$A100 80GB) and Stage 2 (Qwen3.5 annotation, 4$\times$A100 80GB) add a further $\sim$31 A100 GPU-hours on a separate machine; GPT-5.2 annotation runs on OpenAI's hosted infrastructure and is not included in the local GPU-hour total.

\subsection{Active Learning Acquisition Functions}
\label{app:al-acquisition}

We evaluate four pool-based acquisition strategies, two uncertainty-based (Entropy, BALD) and two diversity-based (Core-Set, BADGE), covering the four most widely used acquisition families in the AL literature. At round $t$, all four operate on the current classifier $f_{\theta_t}$, the labeled set $\mathcal{L}_t$, and the unlabeled pool $\mathcal{U}_t$, and select the top-$k$ instances (with $k=50$ in our setup) to be added to $\mathcal{L}_{t+1}$.

\paragraph{Entropy.}
The predictive entropy of the classifier's softmax output over the two classes:
\[
\text{Score}_\text{Ent}(x) = -\sum_{c \in \{0,1\}} p_\theta(c \mid x) \log p_\theta(c \mid x).
\]
The 50 instances with the highest score are selected. Entropy is the standard uncertainty baseline and reduces to selecting instances closest to the decision boundary in our binary setting.

\paragraph{BALD~\citep{houlsby2011bayesian}.}
Bayesian Active Learning by Disagreement selects instances on which an ensemble of stochastic predictions disagrees most:
\[
\text{Score}_\text{BALD}(x) = H[\bar{p}(\cdot \mid x)] - \mathbb{E}_{\theta'}\bigl[H[p_{\theta'}(\cdot \mid x)]\bigr],
\]
where $\bar{p}$ is the average prediction across stochastic forward passes and $H[\cdot]$ is the entropy. We estimate the ensemble via $T=10$ MC Dropout~\citep{gal2016dropout} passes through the classifier with dropout enabled at inference time; the inner expectation is approximated as the mean entropy across the $T$ passes. BALD prioritizes instances where individual stochastic predictions disagree (high $H[\bar{p}]$) but each prediction is itself sharp (low expected entropy), capturing epistemic uncertainty rather than aleatoric uncertainty.

\paragraph{Core-Set~\citep{sener2018active}.}
A pure diversity-based acquisition that selects a subset of $\mathcal{U}_t$ providing the best coverage of the unlabeled-pool embedding space. We use the k-Center-Greedy algorithm: initialized with the current labeled set $\mathcal{L}_t$, it iteratively selects the unlabeled instance with maximum Euclidean distance to its nearest already-selected neighbour, repeating until $k$ instances are picked:
\[
\text{Score}_\text{CS}(x) = \min_{x' \in \mathcal{L}_t \cup \mathcal{S}} \lVert \phi(x) - \phi(x') \rVert_2,
\]
where $\phi(x)$ is the \texttt{[CLS]} embedding from the classifier's encoder and $\mathcal{S}$ is the set of instances selected so far in the current round. Core-Set does not use the classifier's predictions at all; it picks based on geometry of the representation space.

\paragraph{BADGE~\citep{ash2020deep}.}
Batch Active learning by Diverse Gradient Embeddings combines uncertainty and diversity. For each unlabeled instance, the gradient of the cross-entropy loss with respect to the final classifier layer is computed, assuming the model's most likely predicted label as the pseudo-target:
\[
\begin{aligned}
g(x)
  &= \nabla_{w_{\mathrm{cls}}}
     \mathcal{L}(f_\theta(x), \hat{y}), \\
\hat{y}
  &= \arg\max_c p_\theta(c \mid x).
\end{aligned}
\]
The gradient magnitude captures uncertainty (high for instances the model would update most strongly on), while the gradient direction captures the type of update. BADGE applies k-means++ initialization~\citep{arthur2007kmeans} on the gradient embeddings $\{g(x) : x \in \mathcal{U}_t\}$ and selects the $k$ chosen centroid instances. This procedure tends to pick instances that are both uncertain and dissimilar from each other.

\paragraph{Implementation details.}
All acquisitions are computed once per round on the current trained classifier. We use stratified mini-batch processing of the pool to avoid GPU memory issues, with the same batch size as during training (16). Embeddings, entropies, and gradient embeddings are recomputed at every round on the latest classifier. Ties are broken by comment ID order to ensure deterministic acquisition across seeds. For BADGE, we use the scikit-learn \texttt{kmeans++} implementation; for Core-Set, our own implementation of k-Center-Greedy.

\section{Annotation Ablations}
\label{app:annotation-ablations}

This appendix reports full downstream-classifier results under the two annotation ablations: the holistic interface (GPT-H) and the reasoning-disabled Qwen variant (Qwen-DNT). Both share the data splits, training recipe, and evaluation protocol of the main-paper experiments; the only changes are at the LLM-annotation stage.

\subsection{Interface Ablation: GPT-H}
\label{app:interface-ablation}

The holistic (H) interface queries GPT-5.2 with a single prompt asking for the binary label directly, rather than the two-question (Q1 reference, Q2 valence) decomposition used by GPT-D. The H variant labels 7{,}847 of 25{,}974 prefiltered comments as ANTI-IMMIGRANT (30.2\%), compared with 21.8\% under GPT-D (\S\ref{sec:llm-pool}).

\paragraph{Aggregate performance.}
Table~\ref{tab:gpth-ablation} reports F1-Macro and F1-Anti under the H
interface. \textsc{Full-LLM-25K} (GPT-H) reaches 0.739 to 0.751 F1-Macro
across the four backbones, compared with 0.718 to 0.771 for
\textsc{Full-LLM-25K} (GPT-D). At full scale the two interfaces are within
0.021 F1-Macro of each other on every backbone, with GPT-D ahead on
ModernGBERT, gbert-base, and xlm-r-base and GPT-H ahead on german-bert. At
matched volume (3,800 labels) the interface effect is larger and consistent:
GPT-D exceeds GPT-H by 0.018 to 0.050 F1-Macro on all four backbones. The
decomposition therefore contributes most when label volume is limited.

Two properties of the holistic labels help explain this. GPT-H assigns
\textsc{anti-immigrant} to 30.2\% of the pool against a human gold rate of
21.7\%, and agrees with the human gold label on 77.3\% of the 1,200 evaluation
items compared with 84.4\% for GPT-D, with almost all of the difference coming
from false positives. Classifiers trained on GPT-H labels correspondingly
over-predict the minority class and require a higher decision threshold
(tuned to 0.55--0.65 on gold-dev, against 0.47--0.56 for GPT-D).

\begin{table}[h]
\centering
\footnotesize
\setlength{\tabcolsep}{3.5pt}
\begin{tabular}{lcccc}
\toprule
\textbf{Condition} & \textbf{germ.} & \textbf{Mod.} & \textbf{gbert} & \textbf{xlm-r} \\
\midrule
\multicolumn{5}{c}{\textit{F1-Macro}} \\
\textsc{Full-LLM} (3.8K)  & .705 & .710 & .706 & .703 \\
\textsc{Full-LLM-25K}     & .739 & .751 & .749 & .742 \\
\textsc{AL-LLM} (Ent.)    & .674 & .685 & .677 & .677 \\
\textsc{Random-LLM}       & .670 & .668 & .664 & .677 \\
\midrule
\multicolumn{5}{c}{\textit{F1-Anti}} \\
\textsc{Full-LLM} (3.8K)  & .587 & .585 & .587 & .586 \\
\textsc{Full-LLM-25K}     & .616 & .632 & .632 & .617 \\
\textsc{AL-LLM} (Ent.)    & .539 & .532 & .533 & .528 \\
\textsc{Random-LLM}       & .517 & .531 & .520 & .531 \\
\bottomrule
\end{tabular}
\caption{F1-Macro and F1-Anti under the holistic (H) interface; all rows use GPT-H. Mean across 10 seeds.}
\label{tab:gpth-ablation}
\end{table}

\paragraph{Error structure.}
Under the H interface, \textsc{Full-LLM-25K} produces a FP:FN ratio of 2.79 to 3.72 across backbones, above \textsc{Full-Human} (1.86 to 2.74) and above \textsc{Full-LLM-25K} (GPT-D) at 1.86 to 2.37. The H classifiers under-produce false negatives (49.5 to 59.5 FNs per backbone, compared with 55.1 to 78.6 under GPT-D and 64.9 to 74.4 under Full-Human), consistent with the H interface driving over-prediction of the ANTI-IMMIGRANT class. Mean confidence on errors is comparable under the two interfaces (0.685 to 0.737 under H, 0.692 to 0.751 under GPT-D), so the confidence inflation relative to human supervision (0.627 to 0.698) is not specific to either prompt design. Combined with the per-topic gap reductions in \S\ref{sec:topics}, the H interface remains the primary source of the over-detection patterns.

\subsection{Reasoning-Mode Ablation: Qwen-DNT}
\label{app:reasoning-ablation}

The Qwen-DNT variant queries Qwen3.5-122B-A10B via the same decomposed prompt as the main-paper Qwen-DT variant, but with the \texttt{enable\_thinking} flag set to \texttt{False}, eliminating the intermediate reasoning trace before the JSON response. The DNT variant labels 8{,}315 of 25{,}974 prefiltered comments as ANTI-IMMIGRANT (32.0\%), comparable to DT at 30.6\% (\S\ref{sec:llm-pool}).

\paragraph{Aggregate performance.}
Table~\ref{tab:qwendnt-ablation} reports F1-Macro and F1-Anti under DNT.
\textsc{Full-LLM-25K} (Qwen-DNT) reaches 0.722 to 0.746 F1-Macro across the
four backbones, against 0.724 to 0.748 for Qwen-DT. At full scale the two
reasoning modes are within 0.008 F1-Macro of each other on every backbone,
with DT ahead on ModernGBERT and xlm-r-base and DNT ahead on german-bert and
gbert-base. Reasoning therefore has no consistent effect at this label volume,
a much smaller factor than the interface comparison in
App.~\ref{app:interface-ablation}.

\begin{table}[h]
\centering
\footnotesize
\setlength{\tabcolsep}{3.5pt}
\begin{tabular}{lcccc}
\toprule
\textbf{Condition} & \textbf{germ.} & \textbf{Mod.} & \textbf{gbert} & \textbf{xlm-r} \\
\midrule
\multicolumn{5}{c}{\textit{F1-Macro}} \\
\textsc{Full-LLM} (3.8K)  & .716 & .722 & .715 & .688 \\
\textsc{Full-LLM-25K}     & .744 & .746 & .732 & .722 \\
\textsc{AL-LLM} (Ent.)    & .650 & .688 & .682 & .687 \\
\textsc{Random-LLM}       & .669 & .667 & .669 & .659 \\
\midrule
\multicolumn{5}{c}{\textit{F1-Anti}} \\
\textsc{Full-LLM} (3.8K)  & .594 & .593 & .585 & .573 \\
\textsc{Full-LLM-25K}     & .621 & .631 & .613 & .595 \\
\textsc{AL-LLM} (Ent.)    & .498 & .548 & .540 & .536 \\
\textsc{Random-LLM}       & .530 & .523 & .522 & .520 \\
\bottomrule
\end{tabular}
\caption{F1-Macro and F1-Anti under Qwen-DNT; all rows use Qwen-DNT. Mean across 10 seeds.}
\label{tab:qwendnt-ablation}
\end{table}

\paragraph{Error structure.}
\textsc{Full-LLM-25K} (Qwen-DNT) produces FP:FN ratios of 3.06 to 3.88 across
backbones, slightly above Qwen-DT (2.79 to 3.39) and above the
human-supervised baseline (1.86 to 2.74). Mean confidence on errors is lower
under DNT (0.689 to 0.776) than under DT (0.719 to 0.804), and the share of
errors above 0.90 confidence falls from 21 to 39\% under DT to 16 to 35\%
under DNT. Disabling reasoning thus yields classifiers with comparable
performance and error balance and somewhat better calibration on their errors,
though both variants remain more confident on errors than human supervision
(2 to 14\%). Confidence inflation is therefore a property of LLM supervision
in this setting rather than of the reasoning mode.

\paragraph{Label properties.}
Qwen-DNT assigns \textsc{anti-immigrant} to 32.0\% of the pool against a human
gold rate of 21.7\%, and agrees with the human gold label on 78.9\% of the
1,200 evaluation items (Qwen-DT: 30.8\% and 80.3\%). As with the GPT-H, classifiers trained on these labels over-predict the minority class
at a 0.5 decision boundary and select a higher threshold on gold-dev
(0.58--0.60).

\section{Soft-Label Robustness}
\label{app:soft-label-robustness}

This appendix reports the methodology and full results for the soft-label evaluation discussed in \S\ref{sec:rq3}. The hard-label majority-vote ground truth used in the main results obscures annotator disagreement on subjective items; the soft-label evaluation re-scores each classifier against two Bayesian posteriors over the per-item labels, weighting items by inter-annotator agreement rather than collapsing them to a single hard label.

\subsection{Posterior Construction}
\label{app:soft-label-construction}

Both Bayesian posteriors are computed from raw annotator votes on the 1{,}200 evaluation items (3 to 6 annotators per item).

\textbf{Beta-Binomial (BB).} Applies a Laplace prior to the \textsc{ANTI-IMMIGRANT} vote count: $q_i = (k_i + 1) / (N_i + 2)$, where $k_i$ is the number of positive votes and $N_i$ the number of annotators on item $i$. This treats annotators as exchangeable Bernoulli voters.

\textbf{Dawid-Skene (DS).} Jointly estimates per-annotator confusion matrices $\pi_j$ and per-item posteriors $\mu_i$ via EM with a Dirichlet prior ($\alpha = 1$) on each $\pi_j$, treating each (phase, annotator) pair as a distinct worker. EM converges within approximately 50 iterations on our data.

\textbf{Annotator exclusions.} Posterior estimation uses the same exclusions applied to the hard-label gold standard (\S\ref{app:annotator-qc}): Annotator\_6 and Annotator\_7. The BB and DS posteriors are therefore computed over the same vote pool that defines the majority vote.

\textbf{Soft-label metrics.} Expected confusion-matrix entries are computed as $\mathbb{E}[\mathrm{TP}] = \sum_i q_i \cdot \mathbf{1}[\hat{y}_i = 1]$, $\mathbb{E}[\mathrm{FP}] = \sum_i (1 - q_i) \cdot \mathbf{1}[\hat{y}_i = 1]$, $\mathbb{E}[\mathrm{FN}] = \sum_i q_i \cdot \mathbf{1}[\hat{y}_i = 0]$, and $\mathbb{E}[\mathrm{TN}] = \sum_i (1 - q_i) \cdot \mathbf{1}[\hat{y}_i = 0]$, where $q_i$ is the soft label under the relevant posterior. Precision, recall, and F1 are computed from these expected counts in the usual way; macro-F1 averages over the two classes.

\subsection{Per-Backbone Soft-Label Evaluation}
\label{app:soft-eval-full}

Table~\ref{tab:soft-label} reports the full per-backbone soft-label evaluation discussed in \S\ref{sec:rq3}. Both posteriors reproduce the hard-label ranking encoder by encoder: Full-LLM-25K under GPT-D exceeds Full-Human on ModernGBERT, gbert-base, and xlm-r-base and trails on german-bert, and Qwen-DT shows the same pattern with smaller magnitudes.

\begin{table*}[t]
\centering
\footnotesize
\setlength{\tabcolsep}{4pt}
\begin{tabular}{lcccccccccccc}
\toprule
& \multicolumn{4}{c}{\textbf{Hard F1-Macro}} & \multicolumn{4}{c}{\textbf{BB exp-F1}} & \multicolumn{4}{c}{\textbf{DS exp-F1}} \\
\cmidrule(lr){2-5}\cmidrule(lr){6-9}\cmidrule(lr){10-13}
\textbf{Condition} & germ. & Mod. & gbert & xlm-r & germ. & Mod. & gbert & xlm-r & germ. & Mod. & gbert & xlm-r \\
\midrule
\textsc{Full-Human}
& \textbf{.740} & .726 & .735 & .725
& .613 & .600 & .614 & .617
& .735 & .714 & .732 & .728 \\
\textsc{Full-LLM}, GPT-D
& .736 & \underline{.760} & \underline{.736} & .724
& \underline{.620} & .622 & .620 & \underline{.619}
& \underline{.742} & .753 & \underline{.740} & .731 \\
\textsc{Full-LLM-25K}, GPT-D
& .718 & \textbf{.771} & \textbf{.759} & \textbf{.752}
& .601 & \textbf{.632} & \textbf{.628} & \textbf{.620}
& .720 & \textbf{.771} & \textbf{.762} & \textbf{.752} \\
\textsc{Full-LLM}, Qwen-DT
& .708 & .724 & .716 & .692
& .616 & .620 & .619 & .608
& .724 & .733 & .732 & .708 \\
\textsc{Full-LLM-25K}, Qwen-DT
& \underline{.739} & .748 & .724 & \underline{.730}
& \textbf{.628} & \underline{.628} & \underline{.623} & \underline{.619}
& \textbf{.750} & \underline{.755} & .737 & \underline{.737} \\
\bottomrule
\end{tabular}
\caption{Soft-label evaluation under Beta-Binomial (BB) and Dawid-Skene (DS) posteriors of the eval-set labels, all four encoder backbones. Hard F1 included for reference. Mean across 10 seeds. \textbf{Bold}/\underline{underline} = best/second-best per backbone per metric.}
\label{tab:soft-label}
\end{table*}

\section{Error Structure}
\label{app:error-structure}

This appendix reports the full error-structure analysis for the three full-pool conditions discussed in \S\ref{sec:rq4-errors}: FP:FN ratios and confidence on errors (Table~\ref{tab:error-structure} and Figure~\ref{fig:confidence-errors-appendix}). Equivalent error-structure analyses for the interface ablation (GPT-H) and reasoning-mode ablation (Qwen-DNT) are in App.~\ref{app:interface-ablation} and App.~\ref{app:reasoning-ablation}.

\subsection{FP:FN Ratios and Confidence Summary}

The three full-pool conditions differ more in confidence than in error balance.
\textsc{Full-Human} produces FP:FN ratios of 1.86--2.74 and moderate confidence
on errors (0.627--0.698). \textsc{Full-LLM-25K} (GPT-D) is comparable on
balance (1.86--2.37) but more confident on the errors it makes (0.692--0.751),
with 11--30\% of errors above 0.90 confidence.
\textsc{Full-LLM-25K} (Qwen-DT) is somewhat more FP-heavy (2.79--3.39) and the
most confident of the three (0.719--0.804), with 21--39\% of errors above 0.90.
The confidence ordering (Full-Human $<$ GPT-D $<$ Qwen-DT) holds on all four
encoders; the FP:FN ordering is less consistent, with GPT-D more balanced than
\textsc{Full-Human} on xlm-r-base.

\begin{table*}[t]
\centering
\footnotesize
\setlength{\tabcolsep}{4pt}
\begin{tabular}{lcccccccccccc}
\toprule
& \multicolumn{3}{c}{\textbf{german\_bert}} & \multicolumn{3}{c}{\textbf{ModernGBERT}} & \multicolumn{3}{c}{\textbf{gbert-base}} & \multicolumn{3}{c}{\textbf{xlm-r-base}} \\
\cmidrule(lr){2-4}\cmidrule(lr){5-7}\cmidrule(lr){8-10}\cmidrule(lr){11-13}
\textbf{Condition} & ratio & $\bar{c}$ & \%$>$.9 & ratio & $\bar{c}$ & \%$>$.9 & ratio & $\bar{c}$ & \%$>$.9 & ratio & $\bar{c}$ & \%$>$.9 \\
\midrule
\textsc{Full-Human}
& 1.86 & .674 & 7.7
& 2.07 & .698 & 13.8
& 1.86 & .691 & 8.9
& 2.74 & .627 & 2.2 \\
\textsc{Full-LLM-25K} (GPT-D)
& 2.10 & .722 & 19.6
& 2.37 & .751 & 29.7
& 2.27 & .692 & 11.3
& 1.86 & .705 & 18.3 \\
\textsc{Full-LLM-25K} (Qwen-DT)
& 3.39 & .777 & 29.4
& 3.34 & .751 & 25.9
& 3.39 & .804 & 38.6
& 2.79 & .719 & 21.0 \\
\bottomrule
\end{tabular}
\caption{Error structure for the three full-pool conditions, mean across 10 seeds. \textbf{ratio} = FP:FN ratio; $\bar{c}$ = mean predicted-class confidence on misclassified instances; \%$>$\textbf{.9} = share of errors with confidence $>$ 0.90.}
\label{tab:error-structure}
\end{table*}

\subsection{Confidence-Ranked Errors}

Figure~\ref{fig:confidence-errors-appendix} visualizes the same pattern as confidence-ranked errors on ModernGBERT. Each curve plots the predicted-class confidence of all misclassified instances in descending order; the horizontal line at $0.90$ marks the high-confidence threshold, and the intersection of each curve with this line gives the \%$>$.9 value from Table~\ref{tab:error-structure}. The LLM-supervised curves cross $0.90$ later and sit above the \textsc{Full-Human} curve across most of the ranking, but in all three conditions the majority of errors fall below $0.90$ and the curves taper toward $0.50$ in the long tail.

\begin{figure*}[h]
\centering
\includegraphics[width=\linewidth]{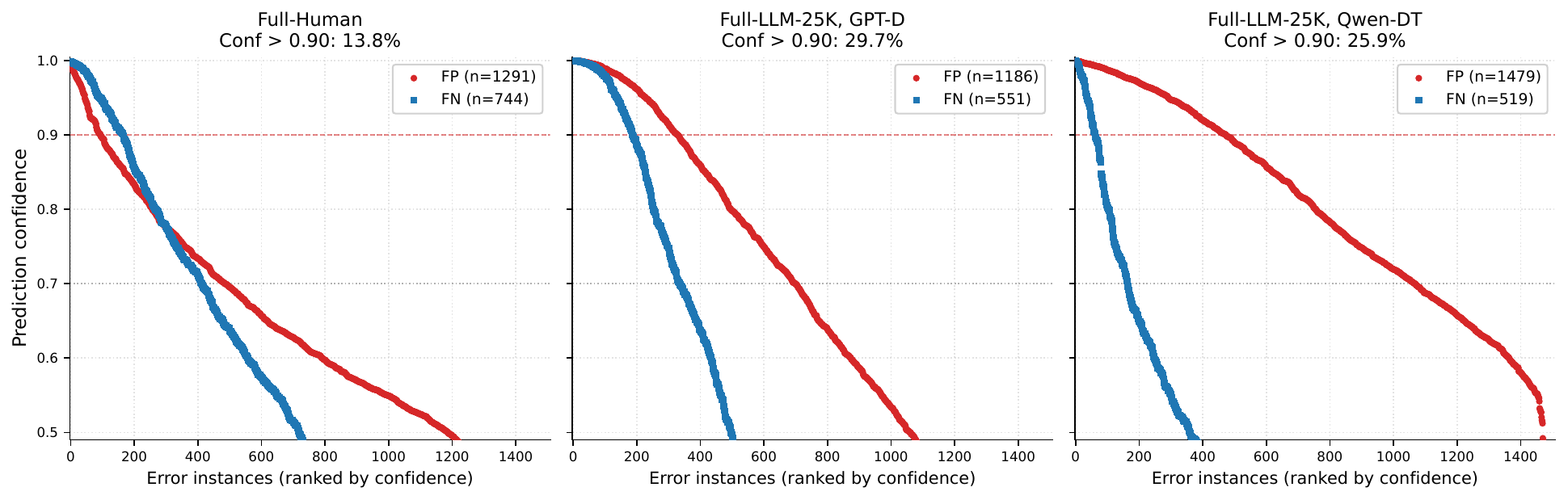}
\caption{Errors ranked by predicted-class confidence on ModernGBERT for \textsc{Full-Human}, \textsc{Full-LLM-25K} (GPT-D), and \textsc{Full-LLM-25K} (Qwen-DT). The horizontal red line marks confidence $= 0.90$; intersection with each curve gives the share of errors that fall above that threshold (cf.\ the \%$>$.9 column in Table~\ref{tab:error-structure}).}
\label{fig:confidence-errors-appendix}
\end{figure*}




\subsection{BERTopic Cluster Structure}
\label{app:topics}

Figure~\ref{fig:tsne} shows the t-SNE projection of all 25{,}974 immigration-relevant comments, clustered into 22 named discourse topics via BERTopic~\cite{grootendorst2022bertopic}. Each cluster is labeled by its dominant keyword pattern. Topics span explicit hostility (C10: Criminal Violence \& Public Safety; C14: Islam \& Islamization), coded political language (C05: Merkel/CDU \& 2015 Legacy; C16: Female Politicians), policy-adjacent discourse (C02: Democracy \& Governance; C06: Migration Policy \& Terminology), and international framing (C11: Racism \& Fascism Discourse; C18: Israel, Palestine \& Middle East; C19: War \& Militarism).

\begin{figure*}[h!]
    \centering
    \includegraphics[width=\textwidth]{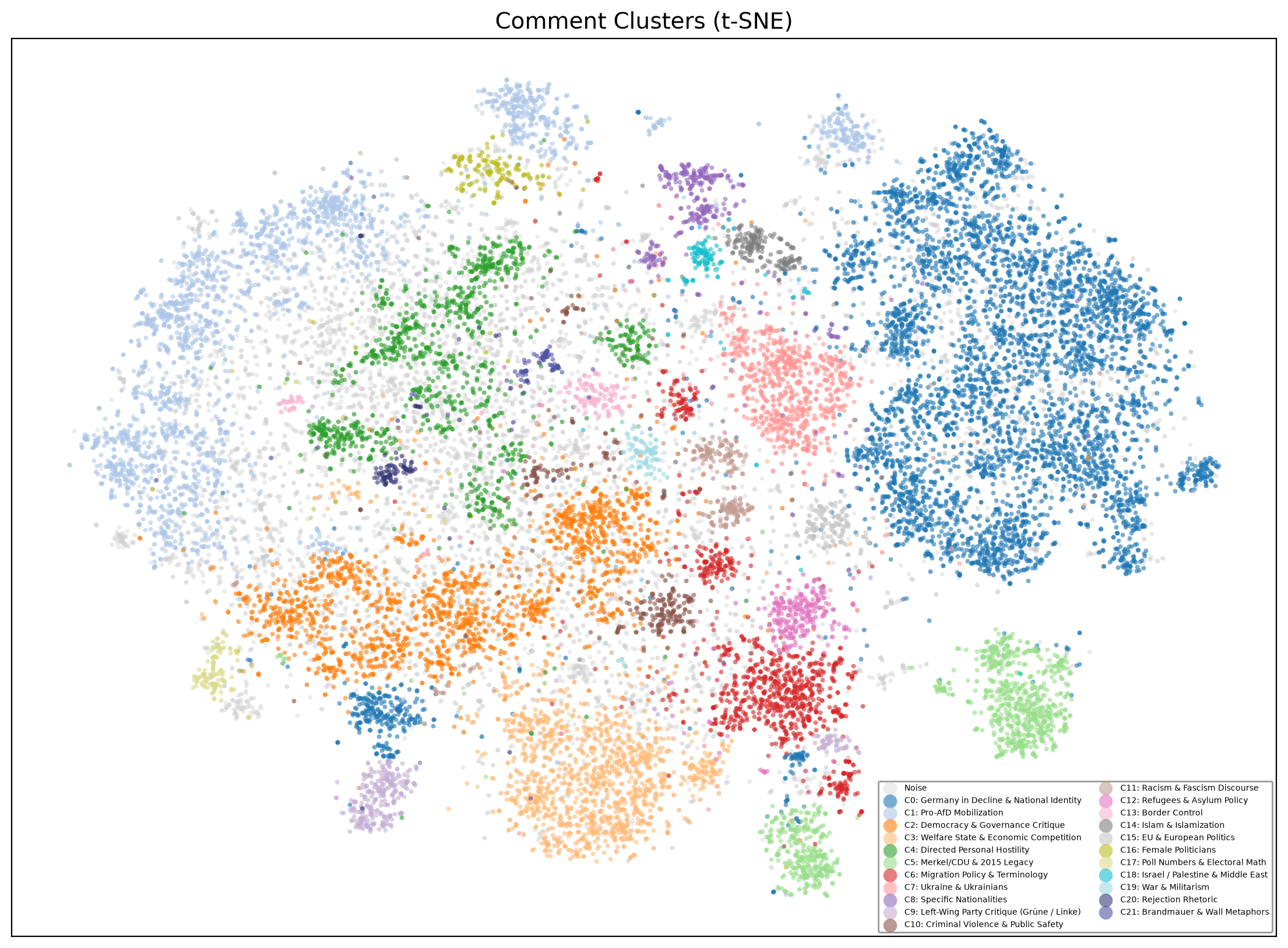}
    \begin{minipage}{\textwidth}
        \centering
        \captionsetup{margin=0pt, width=\linewidth}
        \caption{t-SNE projection of all 25{,}974 immigration-relevant comments, clustered into 22 named discourse topics via BERTopic. Colours denote topic membership; grey points indicate the noise category. The clusters showing the strongest FP-dominant bias under LLM conditions are discussed in \S\ref{sec:topics}.}
        \label{fig:tsne}
    \end{minipage}
\end{figure*}

\subsection{Per-Cluster Human-LLM Agreement}
\label{app:topics_full}

To localize where the holistic-interface (H) over-flagging in \S\ref{sec:why-decomposed} concentrates, Table~\ref{tab:topics-full} reports per-cluster human--GPT-H agreement across all 22 BERTopic discourse topics on the overlapping human-annotated subset. The lowest-agreement clusters (C13 Border Control, C3 Welfare State, C16 Female Politicians, C10 Criminal Violence, C14 Islam \& Islamization) are dominated by the policy-adjacent themes where the co-occurrence shortcut from \S\ref{sec:why-decomposed} fires: immigration-adjacent vocabulary combined with negative valence triggers an \textsc{anti-immigrant} label under the holistic prompt even when the target is policy or institutions rather than people. The Q1/Q2 decomposition under GPT-D resolves these cases by forcing a separate reference judgment (see Pathway~1 in \S\ref{sec:topics}).

\begin{table*}[t]
\centering
\small
\setlength{\tabcolsep}{3pt}

\resizebox{0.99\textwidth}{!}{%
\begin{tabular}{@{}llrr*{4}{rrr}@{}}
\toprule
 & & & & \multicolumn{3}{c}{\textbf{GPT-D}} & \multicolumn{3}{c}{\textbf{GPT-H}} & \multicolumn{3}{c}{\textbf{Qwen-DT}} & \multicolumn{3}{c}{\textbf{Qwen-DNT}} \\
\cmidrule(lr){5-7} \cmidrule(lr){8-10} \cmidrule(lr){11-13} \cmidrule(lr){14-16}
\textbf{ID} & \textbf{Topic} & $n_H$ & \textbf{H\%} & L\% & Agr & Gap & L\% & Agr & Gap & L\% & Agr & Gap & L\% & Agr & Gap \\
\midrule
C0  & Germany in Decline \& National Identity & 1{,}069 & 22 & 28 & 84 & $+6$  & 43 & 73 & $+21$ & 37 & 76 & $+15$ & 37 & 75 & $+15$ \\
C1  & Pro-AfD Mobilization                    & 438 & 5  & 6  & 96 & $+1$  & 11 & 92 & $+6$  & 8  & 94 & $+3$  & 8  & 95 & $+3$ \\
C2  & Democracy \& Governance Critique        & 356 & 10 & 14 & 91 & $+3$  & 20 & 87 & $+10$ & 20 & 87 & $+10$ & 21 & 85 & $+11$ \\
C3  & Welfare State \& Economic Competition   & 293 & 33 & 34 & 75 & $+1$  & 62 & 60 & $+29$ & 49 & 71 & $+16$ & 53 & 67 & $+19$ \\
C6  & Migration Policy \& Terminology         & 221 & 48 & 59 & 73 & $+10$ & 71 & 70 & $+22$ & 59 & 71 & $+10$ & 65 & 71 & $+17$ \\
C5  & Merkel/CDU \& 2015 Legacy               & 218 & 18 & 39 & 78 & $+21$ & 44 & 72 & $+27$ & 43 & 72 & $+25$ & 58 & 59 & $+40$ \\
C4  & Directed Personal Hostility             & 213 & 13 & 16 & 87 & $+3$  & 35 & 73 & $+22$ & 40 & 67 & $+27$ & 38 & 70 & $+26$ \\
C7  & Ukraine \& Ukrainians                   & 168 & 64 & 61 & 74 & $-4$  & 61 & 75 & $-4$  & 69 & 79 & $+4$  & 70 & 74 & $+6$ \\
C8  & Afghanistan, Syria \& Turkey            & 81  & 47 & 62 & 75 & $+15$ & 64 & 78 & $+17$ & 64 & 73 & $+17$ & 72 & 73 & $+25$ \\
C12 & Refugees \& Asylum Policy               & 76  & 55 & 68 & 68 & $+13$ & 70 & 78 & $+14$ & 59 & 72 & $+4$  & 64 & 72 & $+9$ \\
C9  & Left-Wing Party Critique (Gr\"une/Linke)& 70  & 23 & 34 & 83 & $+11$ & 39 & 84 & $+16$ & 27 & 90 & $+4$  & 31 & 86 & $+8$ \\
C10 & Criminal Violence \& Public Safety      & 66  & 44 & 48 & 65 & $+5$  & 65 & 67 & $+21$ & 68 & 69 & $+24$ & 67 & 68 & $+23$ \\
C11 & Racism \& Fascism Discourse             & 52  & 6  & 2  & 96 & $-4$  & 8  & 94 & $+2$  & 10 & 92 & $+4$  & 8  & 90 & $+2$ \\
C13 & Border Control           & 51  & 33 & 82 & 51 & $+49$ & 78 & 43 & $+45$ & 88 & 37 & $+55$ & 96 & 33 & $+63$ \\
C14 & Islam \& Islamization                   & 38  & 66 & 63 & 50 & $-3$  & 87 & 68 & $+21$ & 92 & 68 & $+26$ & 90 & 66 & $+24$ \\
C15 & EU \& European Politics                 & 35  & 23 & 26 & 80 & $+3$  & 31 & 74 & $+8$  & 31 & 80 & $+8$  & 37 & 74 & $+14$ \\
C16 & Female Politicians (Faeser/Storch)      & 31  & 13 & 29 & 71 & $+16$ & 45 & 61 & $+32$ & 32 & 74 & $+19$ & 45 & 61 & $+32$ \\
C19 & War \& Militarism                       & 29  & 34 & 41 & 86 & $+7$  & 48 & 79 & $+14$ & 48 & 79 & $+14$ & 55 & 72 & $+21$ \\
C17 & Poll Numbers \& Electoral Math          & 22  & 0  & 4  & 96 & $+4$  & 14 & 86 & $+14$ & 24 & 76 & $+24$ & 27 & 73 & $+27$ \\
C20 & Rejection Rhetoric                      & 22  & 4  & 4  & 91 & $0$   & 36 & 68 & $+32$ & 41 & 64 & $+36$ & 41 & 64 & $+36$ \\
C18 & Israel, Palestine \& Middle East        & 18  & 39 & 33 & 72 & $-6$  & 44 & 72 & $+6$  & 44 & 83 & $+6$  & 44 & 72 & $+6$ \\
C21 & Brandmauer \& Wall Metaphors            & 15  & 13 & 7  & 93 & $-7$  & 20 & 80 & $+7$  & 40 & 60 & $+27$ & 47 & 53 & $+33$ \\
\bottomrule
\end{tabular}
}
\caption{Per-topic human--LLM comparison across all 22 BERTopic discourse
topics under all four LLM variants. Every rate is computed over the $n_H$
human-annotated items in each cluster: H\% = human \textsc{anti-immigrant}
rate, L\% = LLM rate, Agr = human--LLM label agreement, Gap = L\%$-$H\%.
Sorted by $n_H$. GPT-D yields the smallest gaps of the four variants
($\leq$$+$10 on most topics). Border Control is the only
cluster that every variant over-flags heavily (Gap $+45$ to $+63$); on
Islam \& Islamization the GPT-D rate matches the human rate (Gap $-3$)
while item-level agreement falls, the same rate-corrected but misaligned
pattern as Criminal Violence.}
\label{tab:topics-full}
\end{table*}

\paragraph{Border Control}
(agreement 43\%; 29 human--GPT-H disagreements on the $n_H = 51$
human-annotated items, 26 of them GPT-H over-flags). The over-flagged
comments are border-specific remarks whose negativity predominantly targets
political actors or policy decisions rather than immigrant groups:
rhetorical questions over the 2015 border opening (``Wer war denn 2015 der
Ausl\"oser?''\footnote{Quoted comments in this appendix are lightly trimmed
for presentation; full texts are in the released dataset.}, `Who caused
this in 2015?') and comments on border protection itself (``Diese Grenze war vorher auch gesch\"utzt'', `This border was protected before too').

\paragraph{Islam \& Islamization}
(agreement 68\%, Gap $+21$ under GPT-H; 12 disagreements and 23
both-\textsc{anti} cases on the $n_H = 38$ human-annotated items, 10 of
the disagreements GPT-H over-flags). Disagreement comments target Muslims
as a religious group (``muslim free bitte'', `Muslim-free please'; ``Die
CDU hat Millionen Moslems reingeholt'', `The CDU brought in millions of
Muslims'). By contrast, the both-\textsc{anti} cases typically combine
religious references with explicit immigration vocabulary
(``Islamistenwanderung'') or clearly dehumanizing language.

\subsection{Per-Cluster Agreement under the Decomposed Interface}
\label{app:topics_full_decomposed}

Table~\ref{tab:topics-full} also reports per-cluster agreement under the
decomposed prompt (GPT-D). Relative to GPT-H, agreement rises on 15 of 22
clusters (mean 74\% to 79\%), with the largest gains on the policy-adjacent
clusters that drove the holistic interface's over-detection: Welfare \&
Economic Competition (Agr.\ 60\% to 75\%, rate 62\% to 34\%), Rejection
Rhetoric (68\% to 91\%, 36\% to 4\%), and Directed Personal Hostility
(73\% to 87\%, 35\% to 16\%). Criminal Violence shows the rate correction
without a corresponding agreement gain (rate 65\% to 48\%, agreement 67\%
to 65\%): the decomposition lowers the flag rate but shifts which comments
are flagged. Islam \& Islamization follows the same pattern in stronger
form: the GPT-D rate (63\%) almost matches the human rate (66\%), yet
agreement falls from 68\% to 50\%, so the rate calibrates without item-level
alignment. Border Control alone resists entirely: the rate
stays high under every variant (78--96\%, Gap $+45$ to $+63$), supporting
the topic-intrinsic disagreement pathway identified in \S\ref{sec:topics}.

\subsection{Aggregate Impact of Border-Control Over-Flagging}
\label{app:border-aggregate}

We quantify the aggregate impact of Border-Control (C13) over-flagging
(Table~\ref{tab:border-aggregate}). The over-flagging propagates to the
classifiers: models trained on LLM labels concentrate 4--5$\times$ more of
their false positives in C13 than its $\sim$1\% share of the test set would
predict, versus 2.4$\times$ for the human-trained model. However, because
C13 comprises only $\sim$11 test items, removing it changes macro-F1 by at
most $+0.004$, within per-seed variance, and the LLM-versus-human macro-F1
advantage is unchanged or slightly larger with C13 excluded. At the
annotation level, C13 accounts for only $\sim$3\% of human--LLM
disagreements. Border-Control over-flagging is thus a localized effect that
does not alter the aggregate metrics or conclusions.

\begin{table}[t]
\centering
\footnotesize
\setlength{\tabcolsep}{2.5pt}
\begin{tabular}{@{}l c c c c c@{}}
\toprule
\textbf{Condition} & \textbf{Macro-F1} & $\Delta$\textbf{F1} & \textbf{FP} & \textbf{FP} & \textbf{Over-} \\
 & (all) & ($-$C13) & \textbf{C13} & \textbf{C13+14} & \textbf{rep.} \\
\midrule
Full-Human & .731\,{\scriptsize$\pm$.020} & $+$.002 & 2.5\% & 3.7\% & 2.4$\times$ \\
GPT-D      & .750\,{\scriptsize$\pm$.053} & $+$.004 & 5.3\% & 6.3\% & 5.1$\times$ \\
Qwen-DT    & .735\,{\scriptsize$\pm$.018} & $+$.003 & 4.3\% & 5.4\% & 4.1$\times$ \\
\bottomrule
\end{tabular}
\caption{Aggregate impact of Border-Control (C13) over-flagging on the gold
test split (mean\,$\pm$\,sd over 4 encoders $\times$ 10 seeds); GPT-D and
Qwen-DT denote the \textsc{Full-LLM-25K} conditions. $\Delta$F1 = macro-F1
with C13 removed minus all-items macro-F1. FP C13 = share of the model's
false positives falling in C13; over-rep.\ = that share divided by C13's
share of test items ($\sim$1\%, $\sim$11 items). LLM-trained models
concentrate their false positives in C13 (4--5$\times$ vs 2.4$\times$ for
human supervision), but removing C13 shifts macro-F1 within seed noise, and
the LLM-versus-human advantage holds (GPT-D $+.019 \rightarrow +.020$,
Qwen-DT $+.004 \rightarrow +.006$).}
\label{tab:border-aggregate}
\end{table}

\section{Additional Results}
\label{app:results}

This appendix reports the full learning-curve sweep across all four LLM annotation sources and all four encoders (\S\ref{app:per-source-curves}), complementing Figure~\ref{fig:curves} in \S\ref{sec:rq1}, and the full normalized ALC table for all acquisition strategies, LLM annotators, and encoder backbones (\S\ref{app:alc-full}).

\subsection{Per-Source Learning Curves}
\label{app:per-source-curves}

For completeness we report the full learning-curve sweep across all four encoders. Each panel shows the four active-learning acquisition strategies (entropy, BALD, Core-Set, BADGE), the random-sampling baseline, and the source's own full-pool ceilings at 3.8K and 26K labels. The \textsc{Full-Human} full-pool ceiling is drawn as a solid black reference line.

\begin{figure*}[t]
\centering
\begin{subfigure}{0.49\textwidth}
  \includegraphics[width=\linewidth]{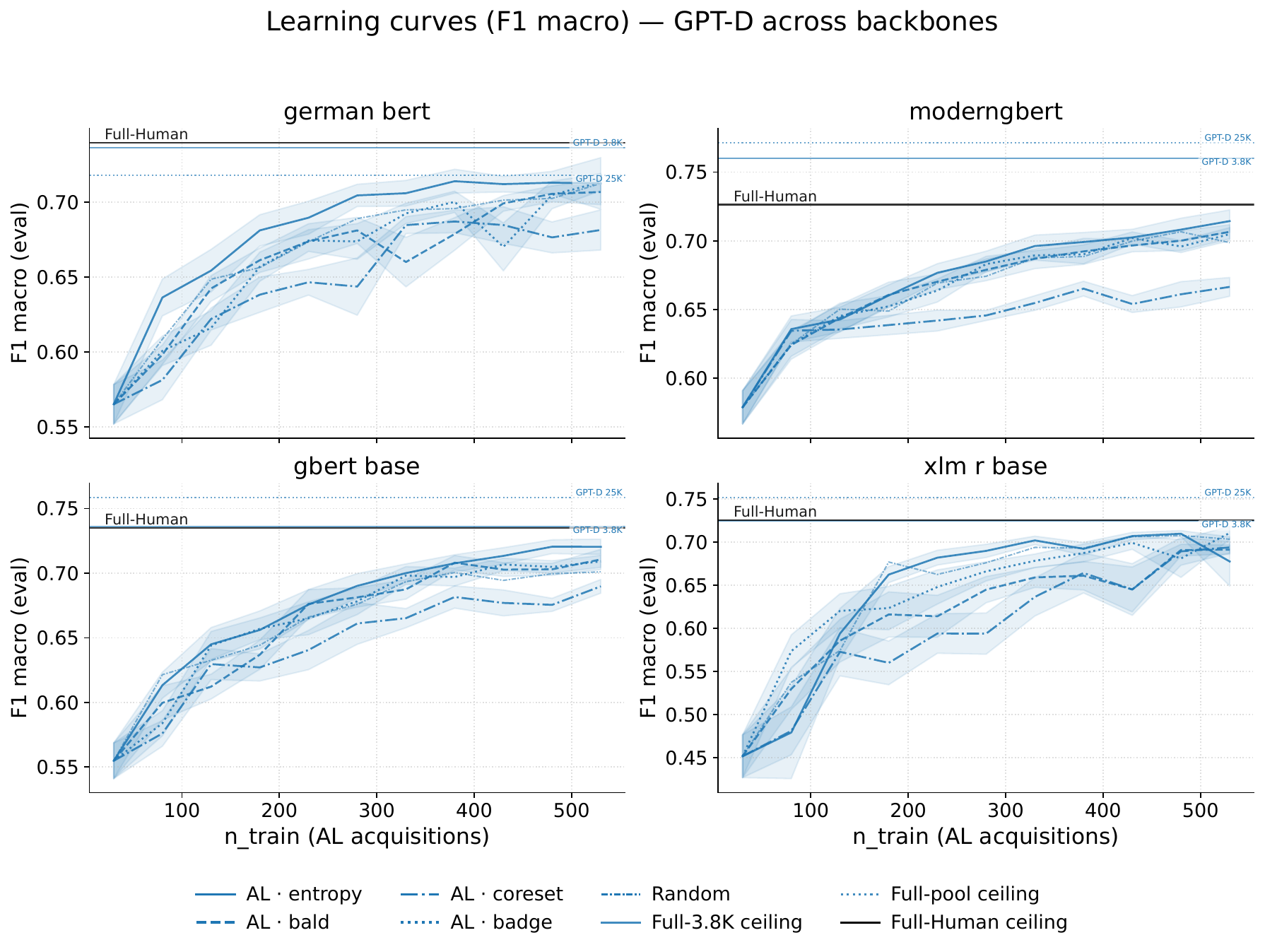}
  \caption{GPT-D}
\end{subfigure}
\hfill
\begin{subfigure}{0.49\textwidth}
  \includegraphics[width=\linewidth]{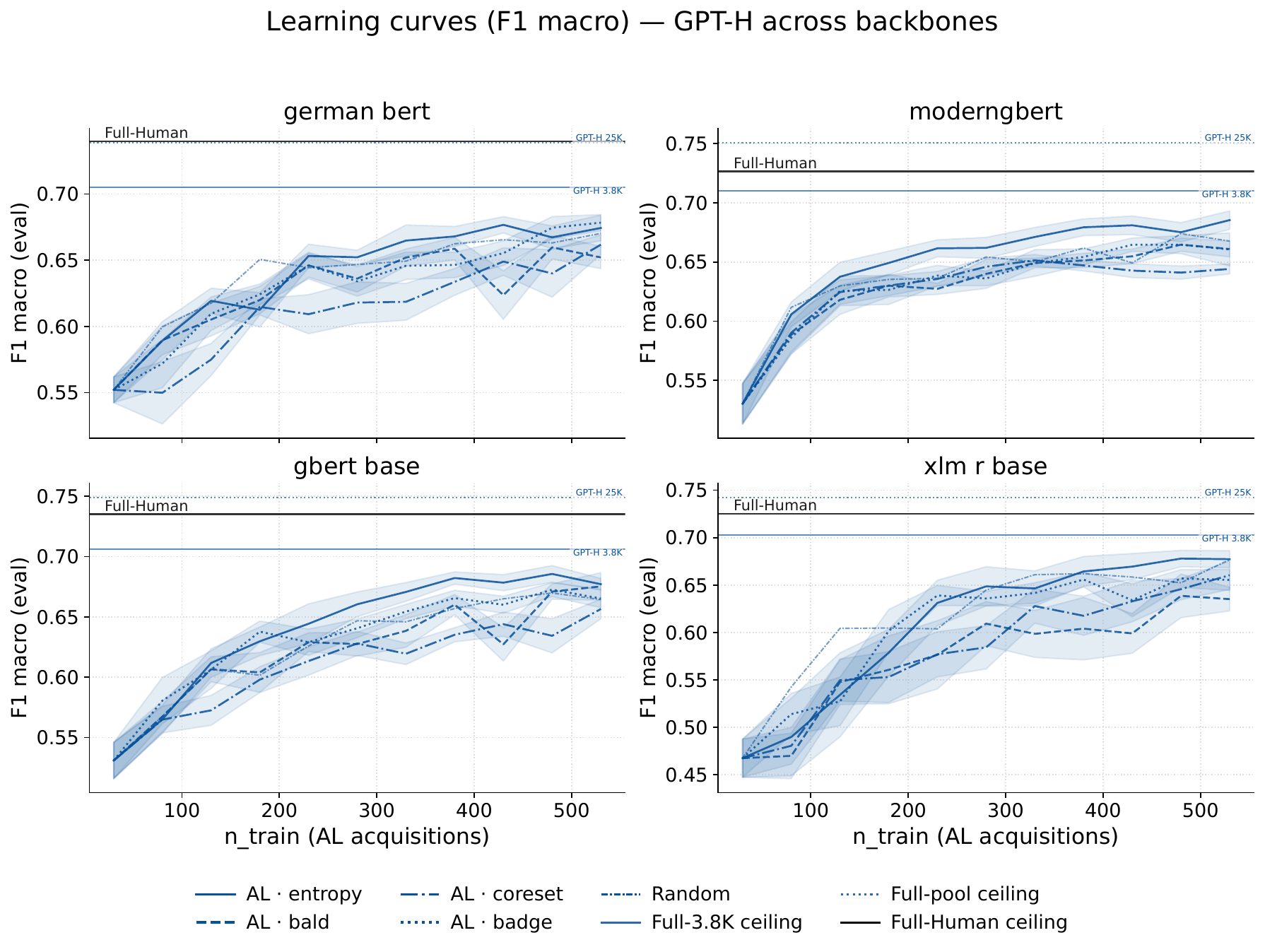}
  \caption{GPT-H}
\end{subfigure}

\begin{subfigure}{0.49\textwidth}
  \includegraphics[width=\linewidth]{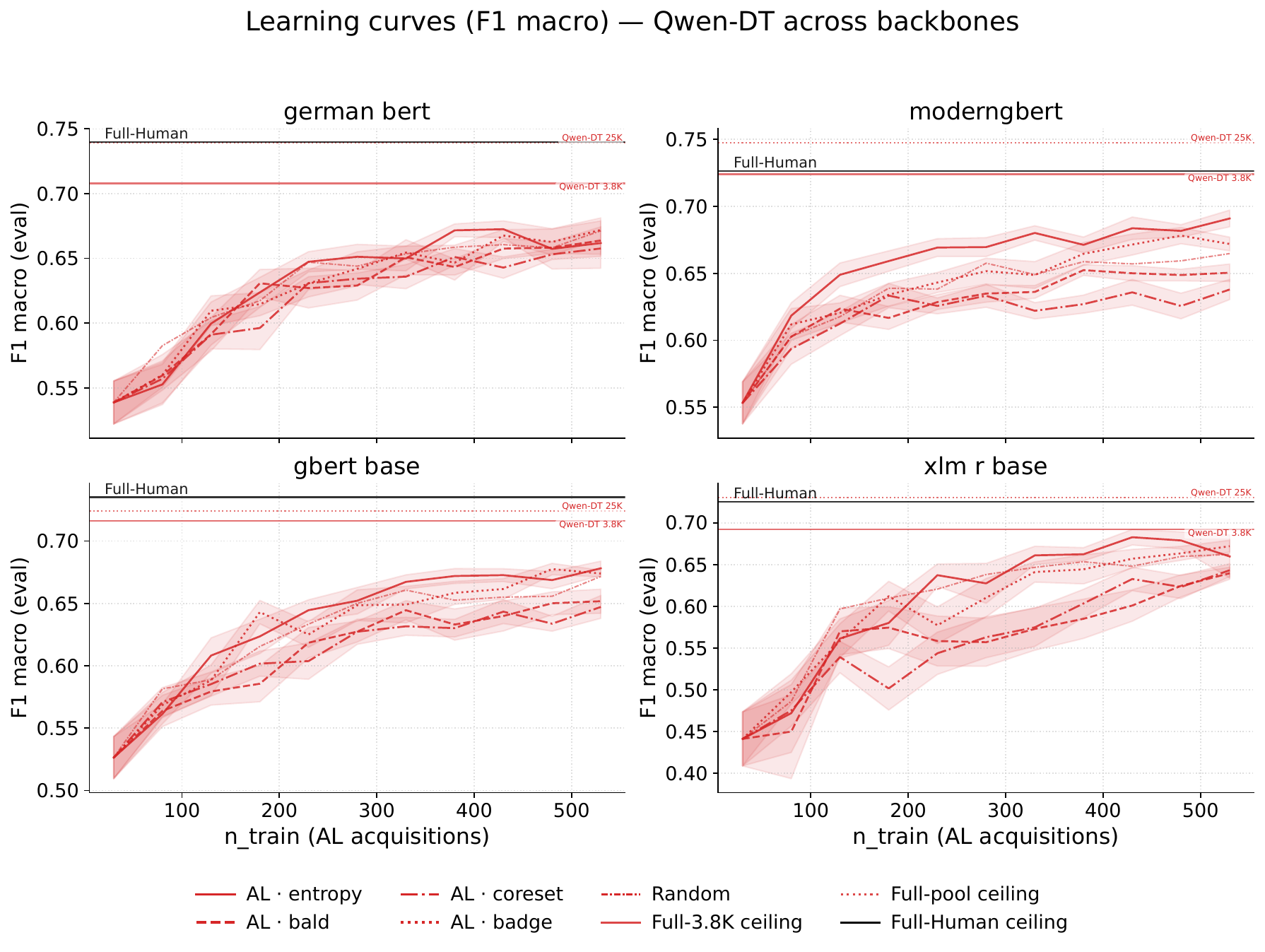}
  \caption{Qwen-DT}
\end{subfigure}
\hfill
\begin{subfigure}{0.49\textwidth}
  \includegraphics[width=\linewidth]{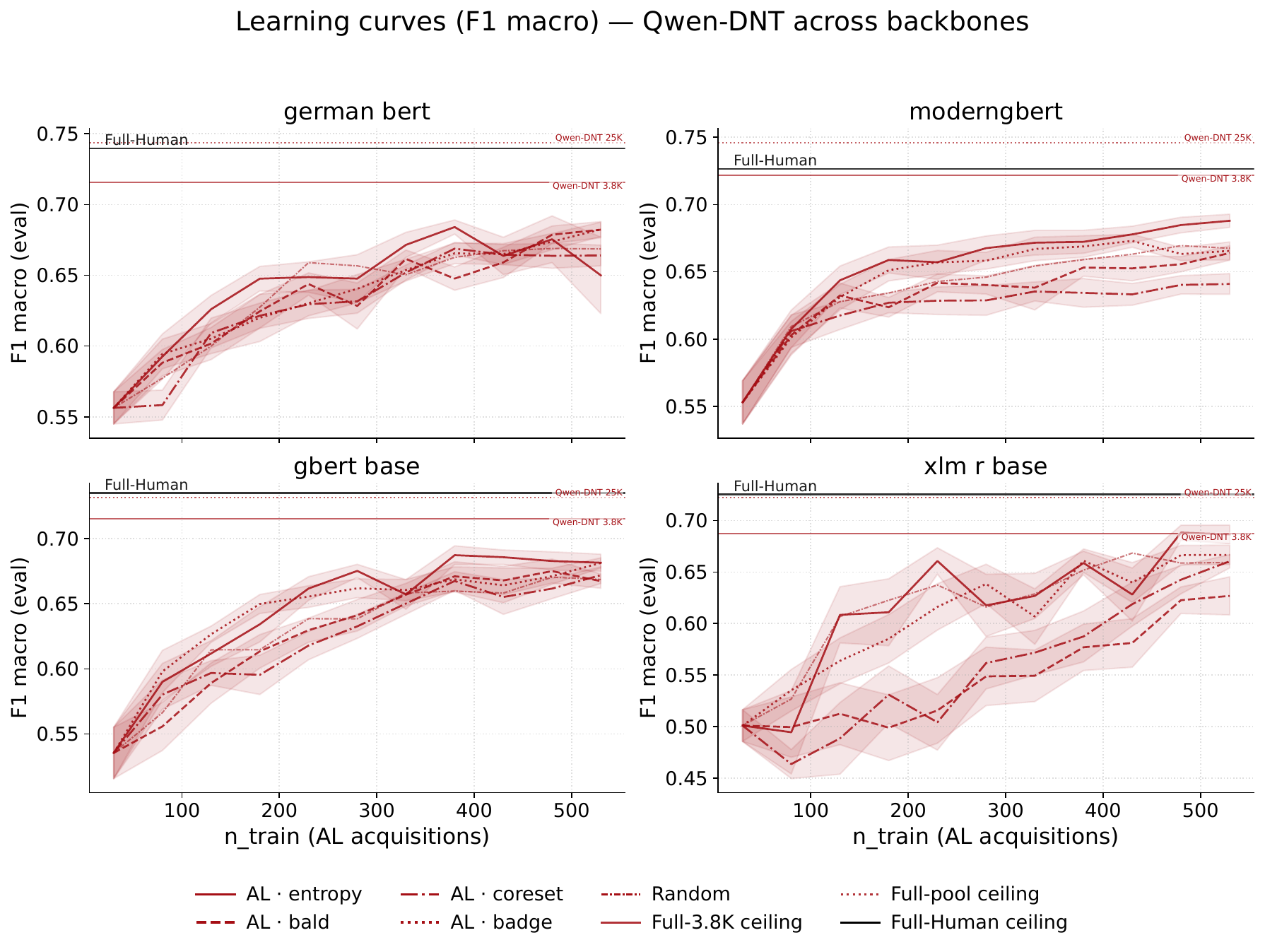}
  \caption{Qwen-DNT}
\end{subfigure}
\caption{F\textsubscript{1}-Macro learning curves per annotation source, across all four encoders. Each panel shows the four active-learning acquisition strategies, the random baseline, the source's own full-pool ceilings at 3.8K and 25K labels, and the \textsc{Full-Human} full-pool ceiling (solid black) as reference.}
\label{fig:curves-f1macro}
\end{figure*}

\begin{figure*}[t]
\centering
\begin{subfigure}{0.49\textwidth}
  \includegraphics[width=\linewidth]{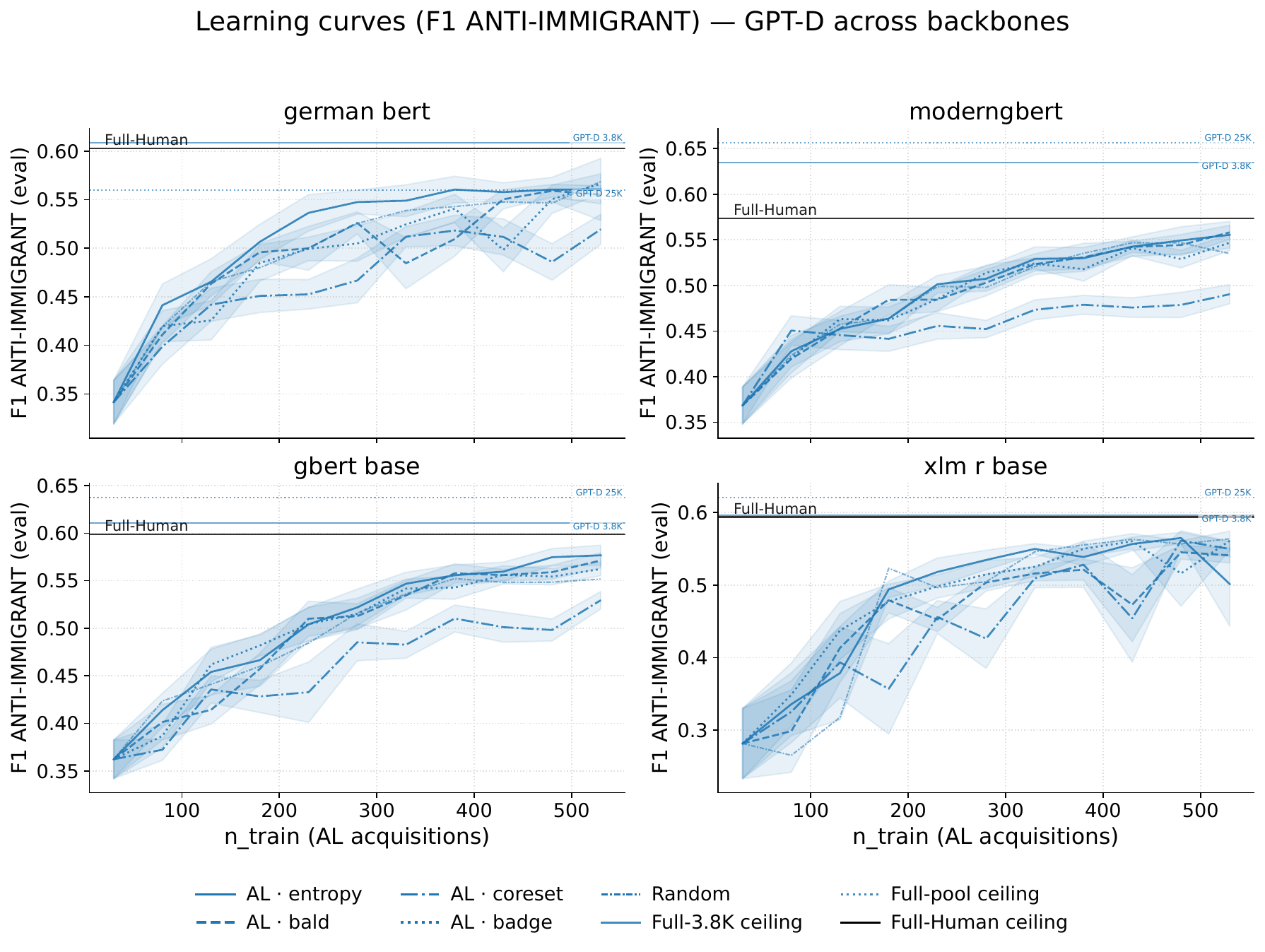}
  \caption{GPT-D}
\end{subfigure}
\hfill
\begin{subfigure}{0.49\textwidth}
  \includegraphics[width=\linewidth]{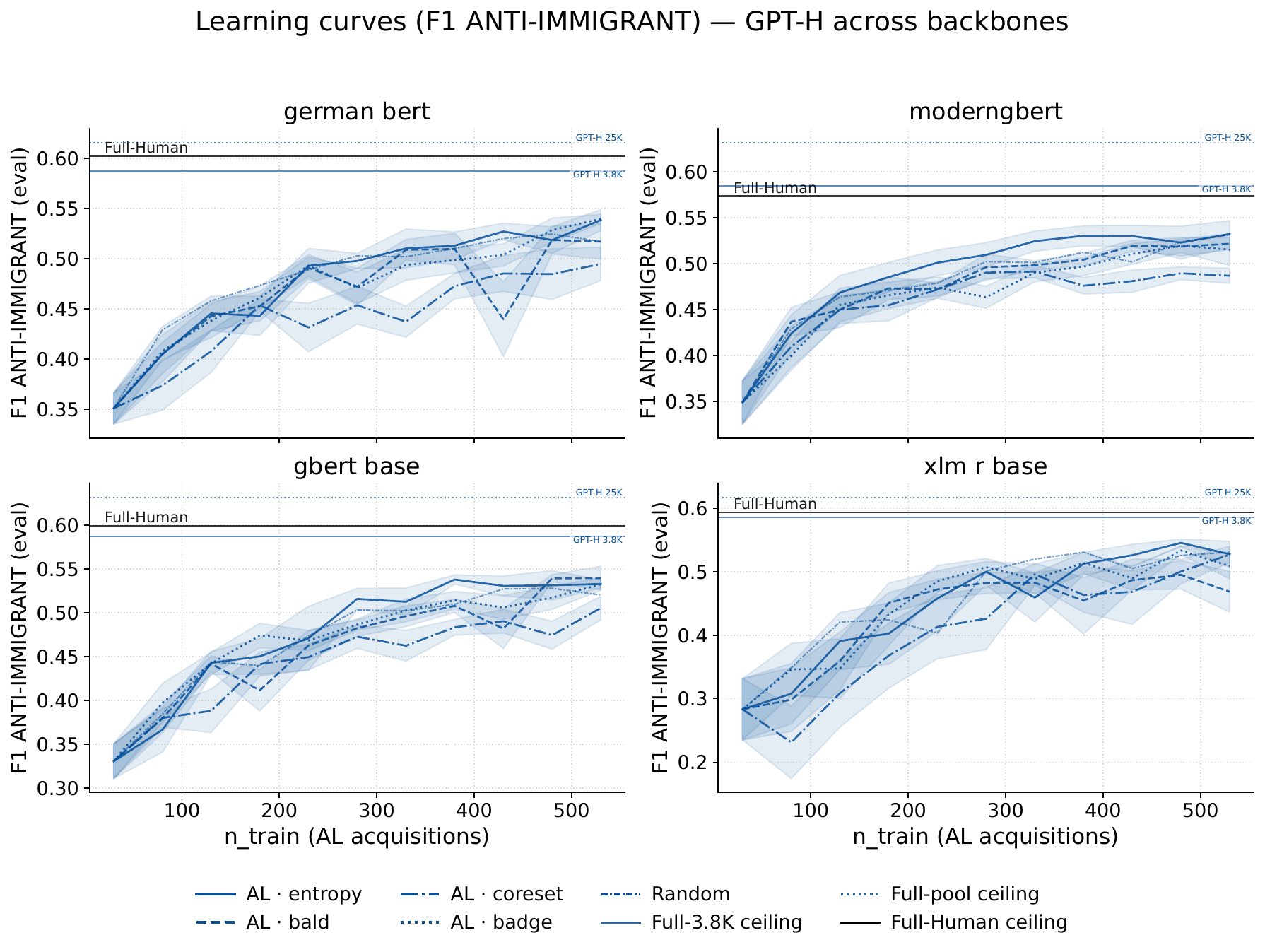}
  \caption{GPT-H}
\end{subfigure}

\begin{subfigure}{0.49\textwidth}
  \includegraphics[width=\linewidth]{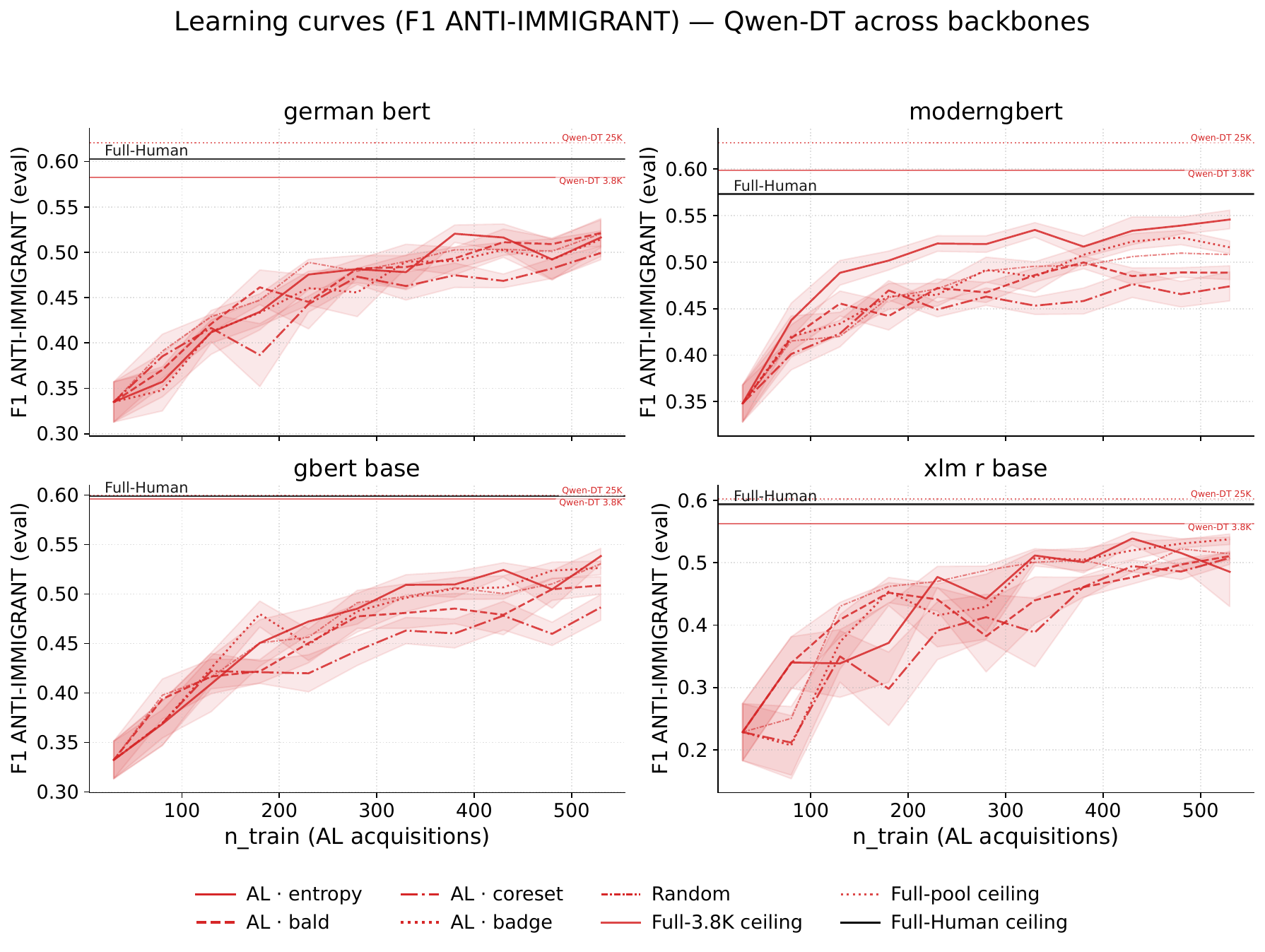}
  \caption{Qwen-DT}
\end{subfigure}
\hfill
\begin{subfigure}{0.49\textwidth}
  \includegraphics[width=\linewidth]{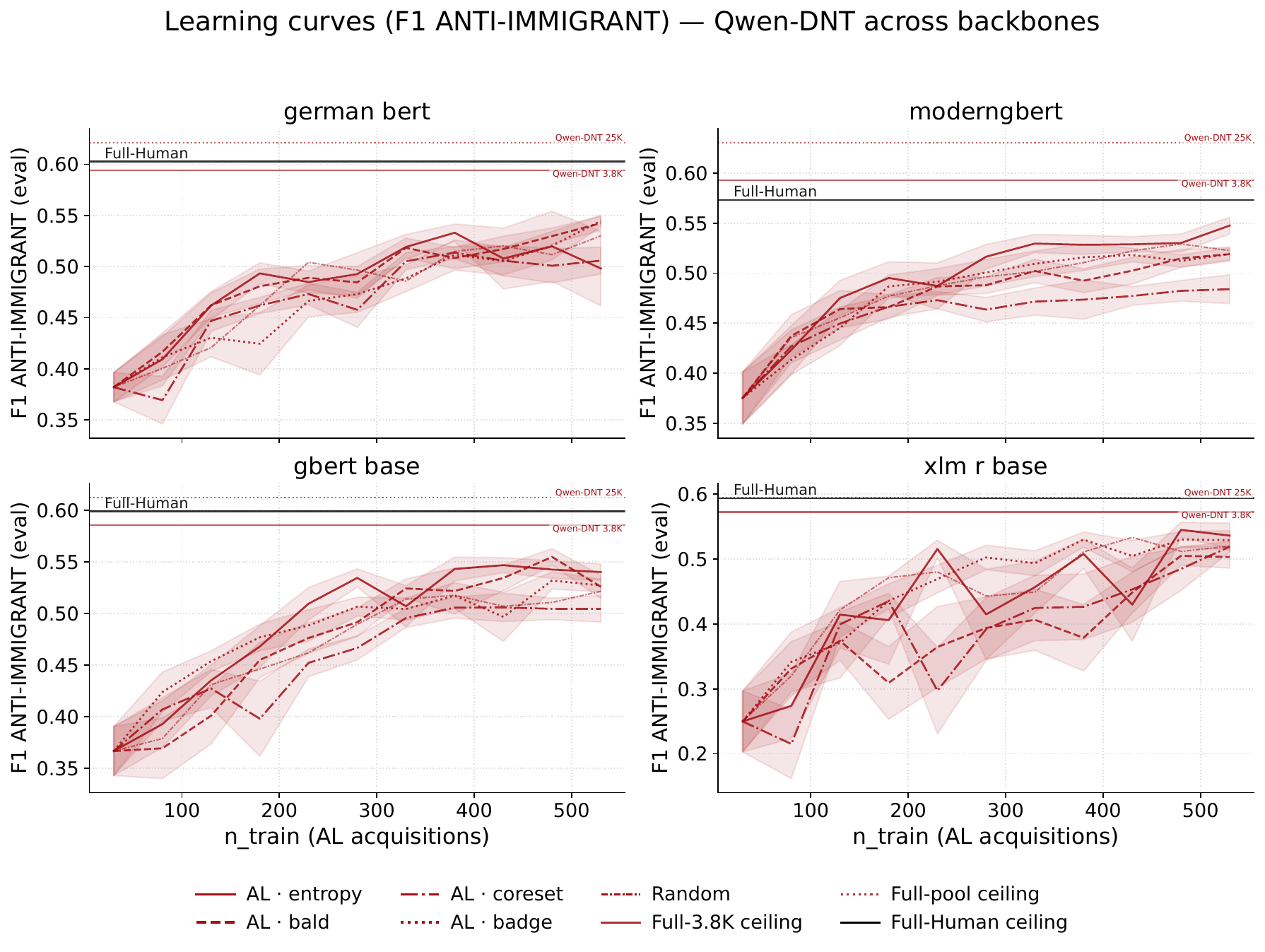}
  \caption{Qwen-DNT}
\end{subfigure}
\caption{F\textsubscript{1} \textsc{Anti-immigrant} learning curves per annotation source, across all four encoders. Each panel shows the four active-learning acquisition strategies, the random baseline, the source's own full-pool ceilings at 3.8K and 25K labels, and the \textsc{Full-Human} full-pool ceiling (solid black) as reference.}
\label{fig:curves-f1anti}
\end{figure*}

\subsection{Full ALC Results}
\label{app:alc-full}

Table~\ref{tab:section4_alc} reports the normalized ALC for all acquisition strategies, LLM annotators, and encoder backbones discussed in \S\ref{sec:rq2}.

\begin{table}[h]
\centering
\small
\begin{tabular}{lcccc}
\toprule
\textbf{Condition} & \textbf{germ.} & \textbf{Mod.} & \textbf{gbert} & \textbf{xlm-r} \\
\midrule
\multicolumn{5}{c}{\textit{Human}} \\
\textsc{AL-Human} Entropy               & .668 & .668 & .659 & .638 \\
\textsc{AL-Human} BALD                  & .658 & .649 & .652 & .616 \\
\textsc{AL-Human} Core-Set              & .648 & .640 & .642 & .611 \\
\textsc{AL-Human} BADGE                 & .663 & .645 & .655 & \textbf{.656} \\
\textsc{Random-Human}                   & .658 & .642 & .643 & .645 \\
\midrule
\multicolumn{5}{c}{\textit{LLM: GPT-D}} \\
\textsc{AL-LLM}, Entropy                & \textbf{.685} & \textbf{.675} & \textbf{.676} & .648 \\
\textsc{AL-LLM}, BALD                   & .664 & \underline{.670} & .664 & .622 \\
\textsc{AL-LLM}, Core-Set               & .649 & .645 & .646 & .601 \\
\textsc{AL-LLM}, BADGE                  & .663 & .669 & \underline{.667} & .646 \\
\textsc{Random-LLM}                     & \underline{.671} & .669 & .666 & \underline{.650} \\
\midrule
\multicolumn{5}{c}{\textit{LLM: Qwen-DT}} \\
\textsc{AL-LLM}, Entropy                & .633 & .660 & .637 & .611 \\
\textsc{AL-LLM}, BALD                   & .625 & .630 & .613 & .563 \\
\textsc{AL-LLM}, Core-Set               & .619 & .620 & .611 & .560 \\
\textsc{AL-LLM}, BADGE                  & .629 & .644 & .632 & .602 \\
\textsc{Random-LLM}                     & .633 & .639 & .629 & .611 \\
\bottomrule
\end{tabular}
\caption{Normalized ALC (area under the F1-Macro learning curve, divided by the budget range), mean across 10 seeds. \textbf{Bold}/\underline{underline} = best/second-best per encoder. Results under GPT-H and Qwen-DNT are in App.~\ref{app:interface-ablation} and App.~\ref{app:reasoning-ablation}.}
\label{tab:section4_alc}
\end{table}

\end{document}